%% file: main.tex
\documentclass[10pt,twocolumn,letterpaper]{article}

\usepackage{cvpr}              

\input{preamble}

\usepackage{graphicx}
\usepackage{subcaption}
\usepackage{amsmath}
\usepackage{amssymb}
\usepackage{booktabs}

\usepackage{url}            
\usepackage{booktabs}       
\usepackage{amsfonts}       
\usepackage{nicefrac}       
\usepackage{microtype}      
\usepackage{xcolor}         

\usepackage{diagbox}
\usepackage[ruled, vlined]{algorithm2e}%
\SetKwInOut{Parameter}{parameter}
\usepackage{wrapfig}
\usepackage{enumitem,amsthm,amsmath,amsfonts,amssymb}
\usepackage{caption}
\usepackage{subcaption}
\usepackage{multirow}
\usepackage{paralist}
\usepackage{makecell}
\usepackage{color}
\usepackage{xcolor}
\usepackage{threeparttable}
\usepackage{amsfonts}
\usepackage{algorithmicx}

\usepackage{dsfont}

\def\sign{\texttt{sign}}

\def\1{\cellcolor{green!30}}
\def\2{\cellcolor{green!30}}
\def\3{\cellcolor{yellow!30}}
\def\4{\cellcolor{yellow!30}}
\def\5{\cellcolor{yellow!30}}
\def\6{\cellcolor{orange!40}}
\def\7{\cellcolor{orange!40}}
\def\8{\cellcolor{orange!40}}
\def\9{\cellcolor{orange!40}}

\definecolor{color_best}{RGB}{22, 138, 173}

\definecolor{highlightcolor}{RGB}{228, 224, 225} 

\usepackage{tikz}

\usepackage{colortbl}
\newcommand{\shu}{\textcolor{red}}

\definecolor{baselinecolor}{gray}{.9}

\definecolor{cvprblue}{rgb}{0.21,0.49,0.74}
\usepackage[pagebackref,breaklinks,colorlinks,citecolor=cvprblue]{hyperref}

\def\paperID{2902} 
\def\confName{CVPR}
\def\confYear{2025}

\title{AI-Face: A Million-Scale Demographically Annotated AI-Generated Face Dataset and Fairness Benchmark\thanks{\textcolor{blue}{This paper has been accepted by CVPR 2025}}}

\author{Li Lin$^{1}$, Santosh$^{1}$, Mingyang Wu$^{1}$, Xin Wang$^{2}$, Shu Hu$^{1}$$^{\dagger}$\\\
$^{1}$Purdue University, West Lafayette, USA {\tt \small\{lin1785, santosh2, wu2415, hu968\}@purdue.edu}\\
$^{2}$University at Albany, State University of New York, New York, USA \tt \small xwang56@albany.edu\\
}
\begin{document}

\maketitle
\let\thefootnote\relax\footnotetext{$^{\dagger}$Corresponding author}

\vspace{-10mm}

\begin{abstract}
AI-generated faces have enriched human life, such as entertainment, education, and art. However, they also pose misuse risks. Therefore, detecting AI-generated faces becomes crucial, yet current detectors show biased performance across different demographic groups. Mitigating biases can be done by designing algorithmic fairness methods, which usually require demographically annotated face datasets for model training. However, \textbf{no existing dataset encompasses both demographic attributes and diverse generative methods simultaneously}, which hinders the development of fair detectors for AI-generated faces. In this work, we introduce the \textbf{AI-Face} dataset, the \textit{first} million-scale demographically annotated AI-generated face image dataset, including real faces, faces from deepfake videos, and faces generated by Generative Adversarial Networks and Diffusion Models. Based on this dataset, we conduct the \textit{first} comprehensive fairness benchmark to assess various AI face detectors and provide valuable insights and findings to promote the future fair design of AI face detectors. Our AI-Face dataset and benchmark code are publicly available at  
{\footnotesize \url{https://github.com/Purdue-M2/AI-Face-FairnessBench}}.
 
\end{abstract}

\vspace{-5mm}
\section{Introduction}\label{sec:introduction}
\vspace{-1mm}
AI-generated faces are created using sophisticated AI technologies that are visually difficult to discern from real ones~\cite{lin2024detecting}. They can be summarized into three categories: deepfake videos \cite{rossler2019faceforensics++} created by typically using Variational Autoencoders (VAEs)~\cite{faceswap_github, fakeapp}, faces generated from Generative Adversarial Networks (GANs)~\cite{brock2018large, karras2019style, karras2020analyzing, karras2021alias}, and Diffusion Models (DMs) \cite{rombach2022high}. These technologies have significantly advanced the realism and controllability of synthetic facial representations. Generated faces can enrich media and increase creativity~\cite{Tojin2023}. However, they also carry significant risks of misuse. For example, during the 2024 United States presidential election, fake face images of Donald Trump surrounded by groups of black people smiling and laughing to encourage African Americans to vote Republican are spreading online~\cite{BBCNews2024}. This could distort public opinion and erode people's trust in media~\cite{saetra2023generative, westerlund2019emergence}, necessitating the detection of AI-generated faces for their ethical use. 
\begin{figure}[t]
    \centering
    \includegraphics[width=0.9\linewidth]{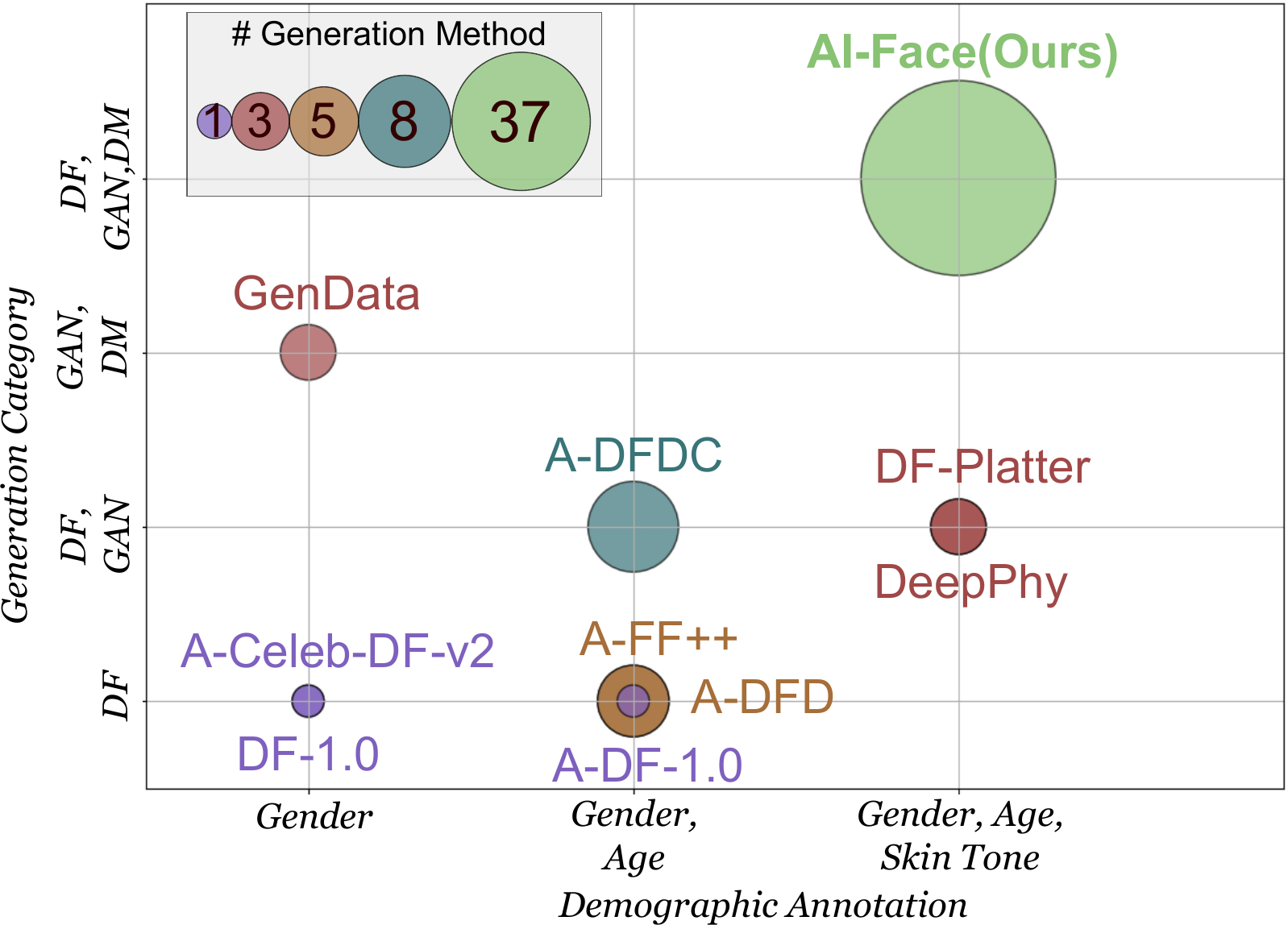}
    \vspace{-2mm}
    \caption{\small \textit{Comparison between AI-Face and other datasets in terms of demographic annotation, generation category, and the number of generation methods. `DF',  `GAN', and `DM' stand for Deepfake Videos, Generative Adversarial Networks, and Diffusion Models.}}
    \label{fig:dataset_compare}
    \vspace{-8mm}
\end{figure}

\begin{table*}[t!]
\centering
\scalebox{0.64}{
\begin{tabular}{c||c|cc|ccc|c|c|ccc}
\Xhline{4\arrayrulewidth}
\rowcolor[HTML]{E4E0E1} 
\cellcolor[HTML]{E4E0E1}                                     & \cellcolor[HTML]{E4E0E1}                       & \multicolumn{2}{c|}{\cellcolor[HTML]{E4E0E1}\textbf{Face Images}} & \multicolumn{3}{c|}{\cellcolor[HTML]{E4E0E1}\textbf{Generation Category}}                  & \cellcolor[HTML]{E4E0E1}                                                                                  & \cellcolor[HTML]{E4E0E1}                                                                                                                                                                                                                                                                                                                                                                                                                                                                                            & \multicolumn{3}{c}{\cellcolor[HTML]{E4E0E1}\textbf{Demographic Annotation}}                \\ \cline{3-7} \cline{10-12} 
\rowcolor[HTML]{E4E0E1} 
\multirow{-2}{*}{\cellcolor[HTML]{E4E0E1}\textbf{Dataset}}            & \multirow{-2}{*}{\cellcolor[HTML]{E4E0E1}\textbf{Year}} & \textbf{\#Real}                      & \textbf{\#Fake}                     & \textbf{Deepfake Videos}           & \textbf{GAN}                       & \textbf{DM}                        & \multirow{-2}{*}{\cellcolor[HTML]{E4E0E1}\begin{tabular}[c]{@{}c@{}}\textbf{\#Generation} \\ \textbf{Methods}\end{tabular}} & \multirow{-2}{*}{\cellcolor[HTML]{E4E0E1}\textbf{Source of Real Images}}                                                                                                                                                                                                                                                                                                                                                                                                                                                     & \textbf{Skin Tone}                    & \textbf{Gender}                 & \textbf{Age}                       \\ \hline \hline

DF-1.0~\cite{jiang2020deeperforensics}    &  2020           & 2.9M            & 14.7M          &  \checkmark           &             &            & 1                                                                               & Self-Recording                                                                                                              &              &   \checkmark             &            \\
DeePhy~\cite{narayan2022deephy}       &2022            & 1K              & 50.4K          &  \checkmark           &  \checkmark           &            & 3                                                                               & YouTube                                                                                                                     &   \checkmark             &    \checkmark          &   \checkmark         \\
DF-Platter~\cite{narayan2023df}     &2023         & 392.3K          & 653.4K         &  \checkmark           &   \checkmark          &            & 3                                                                               & YouTube                                                                                                                     &    \checkmark            &    \checkmark          &    \checkmark        \\
GenData~\cite{teo2023measuring}     &  2023           & -               & 20K            &             &    \checkmark         & \checkmark           & 3                                                                               & CelebA~\cite{liu2015faceattributes}                                                                                                                      &                &   \checkmark           & \\
A-FF++~\cite{xu2024analyzing}  &2024                 & 29.8K           & 149.1K         & \checkmark            &             &            & 5                                                                               & YouTube                                                                                                                     &                &    \checkmark          &   \checkmark         \\
A-DFD~\cite{xu2024analyzing}       &2024             & 10.8K           & 89.6K          &  \checkmark           &             &            & 5                                                                               & Self-Recording                                                                                                              &                &     \checkmark          &    \checkmark        \\
A-DFDC~\cite{xu2024analyzing}     &2024              & 54.5K           & 52.6K          & \checkmark            & \checkmark            &            & 8                                                                               & Self-Recording                                                                                                              &              &      \checkmark        &  \checkmark          \\
A-Celeb-DF-v2~\cite{xu2024analyzing}      &2024      & 26.3K           & 166.5K         &  \checkmark           &             &            & 1                                                                               & Self-Recording                                                                                                              &                &     \checkmark         &            \\
A-DF-1.0~\cite{xu2024analyzing}      &2024           & 870.3K          & 321.5K         &  \checkmark           &             &            & 1                                                                               & Self-Recording                                                                                                              &               &     \checkmark           &  \checkmark          \\
 \hline
AI-Face           &2025          & 400K            & 1.2M           &   \checkmark          &  \checkmark           &    \checkmark        & 37                                                                              & \begin{tabular}[c]{@{}c@{}}FFHQ~\cite{karras2019style}, IMDB-WIKI~\cite{rothe2015dex}, \\ real from FF++~\cite{rossler2019faceforensics++},  DFDC~\cite{dolhansky2020deepfake},\\ DFD~\cite{google2019deepfake},Celeb-DF-v2~\cite{li2020celeb}\end{tabular} &    \checkmark            &   \checkmark        & \checkmark           \\ \Xhline{4\arrayrulewidth}
\end{tabular}
}
\vspace{-2mm}
\caption{\small \textit{{Quantitative comparison of existing datasets with ours on demographically annotated AI-generated faces.} }}
\label{tab:dataset_comparison}
\vspace{-6mm}
\end{table*}

However, one major issue existing in current AI face detectors \cite{pu2022learning, guo2022robust, yan2023ucf, papa2023use} is biased detection (\ie, unfair detection performance among demographic groups~\cite{trinh2021examination, xu2024analyzing, ju2024improving, lin2024preserving}). Mitigating biases can be done by designing algorithmic fairness methods, but they usually require demographically annotated face datasets for model training. For example, works like ~\cite{ju2024improving, lin2024preserving} have made efforts to enhance fairness in the detection based on A-FF++~\cite{xu2024analyzing} and A-DFD~\cite{xu2024analyzing}. However, both datasets are limited to containing only faces from deepfake videos, which could cause the trained models not to be applicable for fairly detecting faces generated by GANs and DMs. 
While some datasets (\eg,  GenData~\cite{teo2023measuring}, DF40 \cite{yan2024df40}) include GAN and DM faces, they either lack demographic annotations or provide only limited demographic attributes.
Most importantly, no existing dataset offers sufficient diversity in generation methods while also providing demographic labels. A comparison of existing datasets is shown in Fig. \ref{fig:dataset_compare}.
\textit{These limitations of existing datasets hamper the development of fair technologies for detecting AI-generated faces.}

Moreover, benchmarking fairness provides a direct method to uncover prevalent and unique fairness issues in recent AI-generated face detection. However, there is a lack of a comprehensive benchmark to estimate the fairness of existing AI face detectors. Existing benchmarks~\cite{li2023continual,deng2024towards, yan2023deepfakebench, le2024sok} primarily assess utility, neglecting systematic fairness evaluation. 
Two studies \cite{trinh2021examination, hazirbas2021towards} do evaluate fairness in detection models, but their examination is based on a few outdated detectors. Furthermore, detectors' fairness reliability (\eg, robustness with test set post-processing, fairness generalization) has not been assessed. 
\textit{The absence of a comprehensive fairness benchmark impedes a thorough understanding of the fairness behaviors of recent AI face detectors and obscures the research path for detector fairness guarantees.}


In this work, we build the \textbf{first} million-scale demographically annotated AI-generated face image dataset: \textbf{AI-Face}. The face images are collected from various public datasets, including the real faces that are usually used to train AI face generators, faces from deepfake videos, and faces generated by GANs and DMs. Each face is demographically annotated 
by our designed measurement method and Contrastive Language-Image Pretraining (CLIP) \cite{radford2021learning}-based lightweight annotator. 
Next, we conduct the \textbf{first} comprehensive fairness benchmark on our dataset to estimate the fairness performance of 12 representative detectors coming from four model types. 
Our benchmark exposes common and unique fairness challenges in recent AI face detectors, providing essential insights that can guide and enhance the future design of fair AI face detectors. Our contributions are as follows:
\begin{compactitem}
    \item We build the first million-scale demographically annotated AI-generated face dataset by leveraging our designed measure and developed lightweight annotator. 
    \item We conduct the first comprehensive fairness benchmark of AI-generated face detectors, providing an extensive fairness assessment of current representative detectors.
    \item Based on our experiments, we summarize the unsolved questions and offer valuable insights within this research field, setting the stage for future investigations.
\end{compactitem}

\section{Background and Motivation}\label{sec:Background and Motivation}
\vspace{-1.5mm}
\textbf{AI-generated Faces and Biased Detection.} AI-generated face images, created by advanced AI technologies, are visually difficult to discern from real ones. They can be summarized into three categories: \textit{1) Deepfake Videos.}  Initiated in 2017~\cite{westerlund2019emergence}, these use face-swapping and face-reenactment techniques with a variational autoencoder to replace a face in a target video with one from a source~\cite{faceswap_github, fakeapp}. Note that our paper focuses solely on images extracted from videos.
\textit{2) GAN-generated Faces.} Post-2017, Generative Adversarial Networks (GANs)~\cite{goodfellow2014generative} like StyleGANs~\cite{karras2019style, karras2020analyzing, karras2021alias} have significantly improved generated face realism.
\textit{3) DM-generated Faces.} Diffusion models (DMs), emerging in 2021, generate detailed faces from textual descriptions and offer greater controllability. Tools like Midjourney~\cite{midjourney} and DALLE2~\cite{ramesh2022hierarchical} facilitate customized face generation. While these AI-generated faces can enhance visual media and creativity~\cite{Tojin2023}, they also pose risks, such as being misused in social media profiles~\cite{osullivan2020, Shannon2022}. Therefore, numerous studies focus on detecting AI-generated faces~\cite{pu2022learning, guo2022robust, yan2023ucf, papa2023use}, but current detectors often show performance disparities among demographic groups~\cite{trinh2021examination, xu2024analyzing, ju2024improving, lin2024preserving}. This bias can lead to unfair targeting or exclusion, undermining trust in detection models. Recent efforts~\cite{ju2024improving, lin2024preserving} aim to enhance fairness in deepfake detection but mainly address deepfake videos, overlooking biases in detecting GAN- and DM-generated faces.

\smallskip
\noindent
\textbf{The Existing Datasets.} Current AI-generated facial datasets with demographic annotations are limited in \textit{size}, \textit{generation categories}, \textit{methods}, and \textit{annotations}, as illustrated in Table~\ref{tab:dataset_comparison}. For instance, A-FF++, A-DFD, A-DFDC, and A-Celeb-DF-v2~\cite{xu2024analyzing} are deepfake video datasets with fewer than one million images. Datasets like DF-1.0~\cite{jiang2020deeperforensics} and DF-Platter~\cite{narayan2023df} lack various demographic annotations. Additionally, existing datasets offer limited generation methods.
These limitations hinder the development of fair AI face detectors, motivating us to build a million-scale demographically annotated AI-Face dataset.

\smallskip
\noindent
\textbf{Benchmark for Detecting AI-generated Faces.} Benchmarks are essential for evaluating AI-generated face detectors under standardized conditions. Existing benchmarks, as shown in Table~\ref{tab:benchmark_comparison}, mainly focus on detectors' utility, often overlooking fairness~\cite{li2023continual, deng2024towards, yan2023deepfakebench, le2024sok,yan2024df40}. Loc et al.~\cite{trinh2021examination} and CCv1~\cite{hazirbas2021towards} examined detector fairness. However, their study did not have an analysis on DM-generated faces and only measured bias between groups in basic scenarios without considering fairness reliability under real-world variations and transformations.
This motivates us to conduct a comprehensive benchmark to evaluate AI face detectors' fairness.

\smallskip
\noindent
\textbf{The Definition of Demographic Categories}.
Demography-related labels are highly salient to measuring bias. Following prior works~\cite{albiero2020does, albiero2020analysis, cook2019demographic, hazirbas2021towards, krishnapriya2020issues, Porgali_2023_CVPR}, we will focus on three key demographic categories: \textit{\textbf{Skin Tone}},    \textit{\textbf{Gender}, and \textbf{Age}}, in this work. 
For skin tone, this vital attribute spans a range from pale to dark. We use the Monk Skin Tone scale~\cite{google_skintone}, specifically designed for computer vision applications.  
For gender, we adopt binary categories (\ie, Male and Female), following practices by many governments~\cite{us_department_of_state_2022, australian_bureau_statistics_2024} and facial recognition research~\cite{cook2019demographic, howard2019effect, raji2019actionable}, based on sex at birth.
For age, using definitions from the United Nations~\cite{united_nations_1982} and Statistics Canada~\cite{statistics_canada_2017_age}, we define five age groups: Child (0-14), Youth (15-24), Adult (25-44), Middle-age Adult (45-64), and Senior (65+). More discussion is in Appendix~\ref{appendix:demographic_definition}.

\section{AI-Face Dataset}
\begin{figure*}[t]
    \centering
    \includegraphics[width=0.9\textwidth]{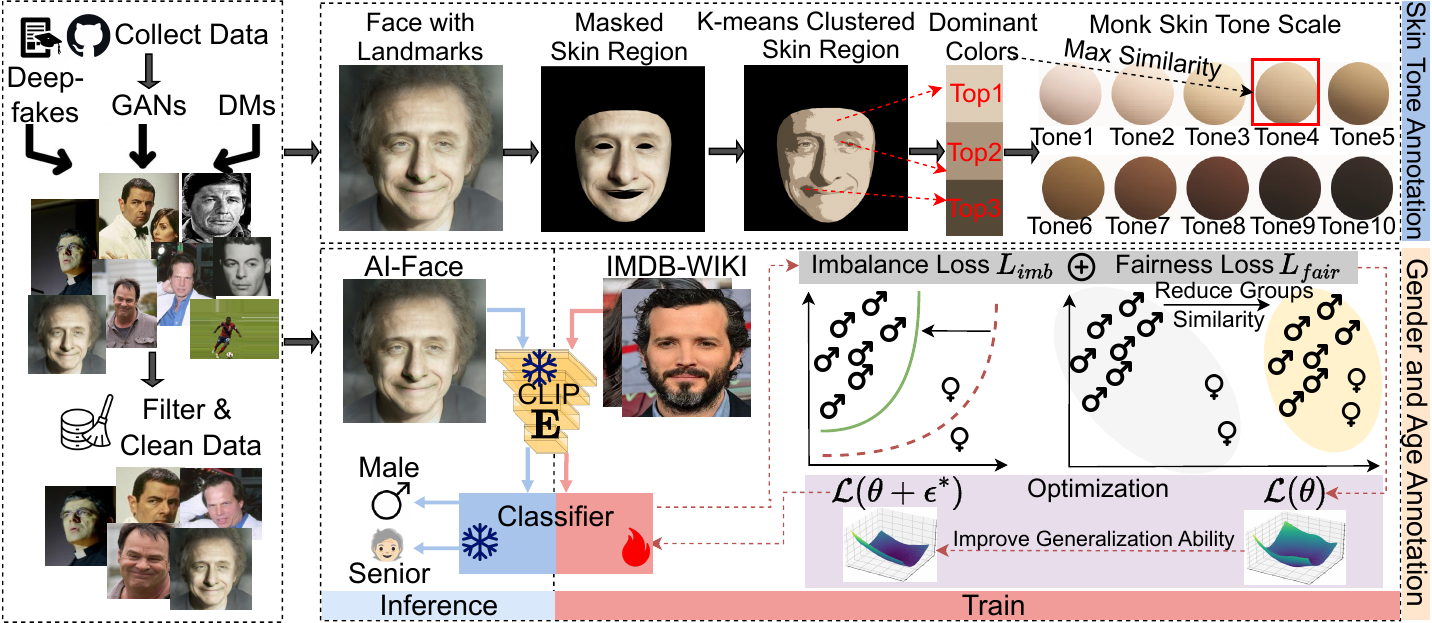}
    \vspace{-2mm}
    \caption{\small \textit{Generation pipeline of our Demographically Annotated AI-Face Dataset. First, we collect and filter face images from Deepfake Videos, GAN-generated faces, and DM-generated faces found in public datasets. Second, we perform skin tone, gender, and age annotation generation. Skin tone is estimated by combining facial landmark detection with color analysis to generate the corresponding annotation. For gender and age, we develop annotators trained on the IMDB-WIKI dataset \cite{rothe2015dex}, then use them to predict attributes for each image.}}
    \label{fig:framework}
    \vspace{-6mm}
\end{figure*}

\vspace{-1mm}
This section outlines the process of building our demographically annotated AI-Face dataset (see Fig.~\ref{fig:framework}), along with its statistics and annotation quality assessment.

\begin{table}[t]
    \centering
    \scalebox{0.61}{
\begin{tabular}{c||c|ccc|ccc}
\Xhline{4\arrayrulewidth}
\rowcolor[HTML]{E4E0E1} 
\cellcolor[HTML]{E4E0E1}                                                                                         & \cellcolor[HTML]{E4E0E1}                                & \multicolumn{3}{c|}{\cellcolor[HTML]{E4E0E1}\textbf{Category}}                                                                                                                                                                 & \multicolumn{3}{c}{\cellcolor[HTML]{E4E0E1}\textbf{Scope of Benchmark}}                                                                         \\ \cline{3-8}  
\rowcolor[HTML]{E4E0E1} 
\cellcolor[HTML]{E4E0E1}                                                                                         & \cellcolor[HTML]{E4E0E1}                                & \cellcolor[HTML]{E4E0E1}                                                                                      & \cellcolor[HTML]{E4E0E1}                               & \cellcolor[HTML]{E4E0E1}                              & \multicolumn{1}{c|}{\cellcolor[HTML]{E4E0E1}}                                   & \multicolumn{2}{c}{\cellcolor[HTML]{E4E0E1}\textbf{Fairness}} \\
\rowcolor[HTML]{E4E0E1} 
\multirow{-3}{*}{\cellcolor[HTML]{E4E0E1}\textbf{\begin{tabular}[c]{@{}c@{}}Existing\\ Benchmarks\end{tabular}}} & \multirow{-3}{*}{\cellcolor[HTML]{E4E0E1}\textbf{Year}} & \multirow{-2}{*}{\cellcolor[HTML]{E4E0E1}\textbf{\begin{tabular}[c]{@{}c@{}}Deepfake \\ Videos\end{tabular}}} & \multirow{-2}{*}{\cellcolor[HTML]{E4E0E1}\textbf{GAN}} & \multirow{-2}{*}{\cellcolor[HTML]{E4E0E1}\textbf{DM}} & \multicolumn{1}{c|}{\multirow{-2}{*}{\cellcolor[HTML]{E4E0E1}\textbf{Utility}}} & \textbf{General}            & \textbf{Reliability}            \\ \hline \hline
Loc et al.~\cite{trinh2021examination}                                                                                                        & 2021                                                    & \checkmark                                                                                                              &                                                        &                                                       & \multicolumn{1}{c|}{\checkmark }                                                           &      \checkmark                        &                                 \\
CCv1~\cite{hazirbas2021towards}                                                                                                             & 2021                                                    &    \checkmark                                                                                                            &     \checkmark                                                    &                                                       & \multicolumn{1}{c|}{\checkmark }                                                           &    \checkmark                          &                                 \\
DeepfakeBench~\cite{yan2023deepfakebench}                                                                                                     & 2023                                                    &    \checkmark                                                                                                            &        \checkmark                                                 &                                                       & \multicolumn{1}{c|}{\checkmark }                                                           &                             &                                 \\
CDDB~\cite{li2023continual}                                                                                                             & 2023                                                    &                                                                                                               &   \checkmark                                                      &                                                       & \multicolumn{1}{c|}{\checkmark }                                                           &                             &                                 \\
Lin et al.~\cite{deng2024towards}                                                                                                       & 2024                                                    &  \checkmark                                                                                                              &          \checkmark                                               &                                                       & \multicolumn{1}{c|}{\checkmark }                                                           &                             &                                 \\
Le et al.~\cite{le2024sok}                                                                                                        & 2024                                                    & \checkmark                                                                                                               &    \checkmark                                                     &                                                       & \multicolumn{1}{c|}{\checkmark }                                                           &                             &                                 \\
DF40~\cite{yan2024df40}                                                                                                              & 2024                                                    &      \checkmark                                                                                                          &      \checkmark                                                   &  \checkmark                                                      & \multicolumn{1}{c|}{\checkmark }                                                           &                             &                                 \\ \hline
Ours                                                                                                             &2025                                                         &        \checkmark                                                                                                        &  \checkmark                                                       &    \checkmark                                                    & \multicolumn{1}{c|}{\checkmark }                                                           & \checkmark                             &        \checkmark                          \\ \Xhline{4\arrayrulewidth}
\end{tabular}
}
\vspace{-2mm}
\caption{\small \textit{{Comparison of existing AI-generated face detection benchmarks and ours. Fairness `General' means fairness evaluation under default/basic settings. Fairness `Reliability' measures fairness consistency across dynamic scenarios (\eg, post-processing).}}}
\label{tab:benchmark_comparison}
\vspace{-6mm}
\end{table}

\subsection{Data Collection}\label{subsec:data_collection}
We build our AI-Face dataset by collecting and integrating public real and AI-generated face images sourced from academic publications, GitHub repositories, and commercial tools. We strictly adhere to the license agreements of all datasets to ensure that they allow inclusion in our datasets and secondary use for training and testing.
More details are in Appendix \shu{\ref{appendix:dataset_detailed_info}}.
Specifically, the fake face images in our dataset originate from \textbf{4 Deepfake Video datasets} (\ie, FF++~\cite{rossler2019faceforensics++}, DFDC~\cite{dolhansky2020deepfake}, DFD~\cite{google2019deepfake}, and Celeb-DF-v2~\cite{li2020celeb}), generated by \textbf{10 GAN} models (\ie, AttGAN~\cite{giudice2021fighting}, MMDGAN~\cite{asnani2023reverse}, StarGAN~\cite{giudice2021fighting}, StyleGANs~\cite{giudice2021fighting, david_beniaguev_2022_SFHQ, lu2023seeing}, MSGGAN~\cite{asnani2023reverse}, ProGAN~\cite{dang2018deep}, STGAN~\cite{asnani2023reverse}, and VQGAN~\cite{esser2021taming}), and \textbf{8 DM} models (\ie, DALLE2~\cite{wang2023dire}, IF~\cite{wang2023dire}, Midjourney~\cite{wang2023dire}, DCFace~\cite{kim2023dcface}, Latent Diffusion~\cite{corvi2023detection}, Palette~\cite{awsafur2023artifact}, Stable Diffusion v1.5~\cite{song2023robustness}, Stable Diffusion Inpainting~\cite{song2023robustness}). This constitutes a total of 1,245,660 fake face images in our dataset.  We include \textbf{6 real} source datasets (\ie, FFHQ~\cite{karras2019style}, IMDB-WIKI~\cite{rothe2015dex}, and real images from FF++~\cite{rossler2019faceforensics++}, DFDC~\cite{dolhansky2020deepfake}, DFD~\cite{google2019deepfake}, and Celeb-DF-v2~\cite{li2020celeb}). All of them are usually used as a training set  for generative models to generate fake face images. This constitutes a total of 400,885 real face images in our dataset. In general,  our dataset contains 28 subsets and 37 generation methods (\ie, 5 in FF++, 5 in DFD, 8 in DFDC, 1 in Celeb-DF-v2, 10 GANs, and 8 DMs).
For all images, we use RetinaFace~\cite{deng2020retinaface} for detecting and cropping faces.

\subsection{Annotation Generation}\label{subsec:annotation_generation}
\subsubsection{Skin Tone Annotation Generation}
Skin tone is typically measured using an intuitive approach \cite{Krishnapriya_2022_WACV, Thong_2023_ICCV}, without requiring a predictive model.
Inspired by \cite{Krishnapriya_2022_WACV}, we developed a method to estimate skin tone using the Monk Skin Tone (MST) Scale~\cite{google_skintone} (including 10-shade scales: Tone 1 to 10) by combining facial landmark detection with color analysis. Specifically, utilizing Mediapipe's FaceMesh~\cite{lugaresi2019mediapipe} for precise facial landmark localization, we isolate skin regions while excluding non-skin areas such as the eyes and mouth. Based on the detected landmarks, we generate a mask to extract skin pixels from the facial area. These pixels are then subjected to K-Means clustering \cite{hartigan1979algorithm} (we use K$=3$ in practice) to identify the dominant skin color within the region of interest. The top-1 largest color cluster is mapped to the closest tone in the MST Scale by calculating the Euclidean distance between the cluster centroid and the MST reference colors in RGB space.

\subsubsection{Gender and Age Annotation Generation}
For generating gender and age annotations, the existing online software (\eg, Face++~\cite{face++}) and open-source tools (\eg, InsightFace~\cite{insightface}) can be used for the prediction. However, they fall short in our task due to two reasons: 
\textit{1)} They are mostly designed for face recognition and trained on datasets of real face images but lack generalization capability for annotating AI-generated face images.
\textit{2)} Their use may introduce bias into our dataset, as they are typically designed and trained without careful consideration of bias and imbalance in the training set. See Appendix ~\ref{appendix:annotator_evaluation} for our experimental study on these tools.
To this end, we have to develop our specific annotators to predict gender and age annotations for each image in our dataset. 

\smallskip
\noindent
\textbf{Problem Definition}. Given a training dataset  
$\mathbb{D}=\{(X_i, A_i)\}_{i=1}^n$ with size $n$, where $X_i$ represents the $i$-th face image and $A_i$ signifies a demographic attribute associated with $X_i$. Here, $A_i\in \mathcal{A}$, where $\mathcal{A}$ represents user-defined groups (\eg, for gender, $\mathcal{A}=\{\text{Female, Male}\}$. For age, $\mathcal{A}= \{\text{Child, Youth, Adult, Middle-age Adult, Senior}\}$). 
Our goal is to design a lightweight, generalizable annotator based on $\mathbb{D}$ that reduces bias while predicting facial demographic attributes for each image in our dataset. In practice, we use IMDB-WIKI \cite{rothe2015dex} as training dataset, which contains images along with profile metadata sourced from IMDb and Wikipedia, \textit{ensuring that the demographic annotations are as accurate as possible}. We trained two annotators with identical architecture and training procedures for gender and age annotations, respectively. 

\smallskip
\noindent
\textbf{Annotator Architecture}. We build a lightweight annotator based on the CLIP \cite{radford2021learning} foundation model by leveraging its strong zero-shot and few-shot learning capabilities. Specifically, our annotator employs a frozen pre-trained CLIP ViT L/14~\cite{openclip2021} as a feature extractor $\mathbf{E}$ followed by a trainable classifier parameterized by $\theta$, which contains 3-layer Multilayer Perceptron (MLP) $\mathbf{M}$ and a classification head $h$.

\smallskip
\noindent
\textbf{Learning Objective}. Aware that neural networks can perform poorly when the training dataset suffers from class-imbalance~\cite{cao2019learning} and CLIP is not free from demographic bias~\cite{agarwal2021evaluating, tanjim2024discovering, wang2024learn}, we introduce an imbalance loss and fairness loss to address these challenges in the annotator training. Specifically, for image $X_i$, its feature $f_i$ is obtained through $f_i \!\!=\!\! \mathbf{M}(\mathbf{E}(X_i))$. Next, two losses are detailed below. 

\underline{Imbalance Loss:} To mitigate the impact of imbalance data, we use Vector Scaling \cite{kini2021label} loss, which is a re-weighting method for training models on the imbalanced data with distribution shifts and can be expressed as 
\vspace{-1mm}
\begin{equation*}
    L_{imb}=\frac{1}{n}\sum_{i=1}^n -u_{A_i} \log \frac{e^{\zeta_{A_i} h(f_i)_{A_i} + \Delta_{A_i}}}{\sum_{A\in \mathcal{A}} e^{\zeta_A h(f_i)_{A} + \Delta_A}},
    \vspace{-1mm}
\end{equation*}
where $u_{A_i}$ is the weighting factor for attribute $A_i$. $h(f_i)_{A_i}$ is the predict logit on $A_i$. $\zeta_{A_i}$ is the multiplicative logit scaling factor, calculated as the inverse of $A_i$’s frequency. $\Delta_{A_i}$ is the additive logit scaling factor, calculated as the log of $A_i$ probabilities. More  details about them are in appendix \ref{appendix:annotator_develop}.

\underline{Fairness Loss:} We introduce a fairness loss to minimize the disparity between the distribution $\mathcal{D}^f$ of $f$ and the conditional distribution $\mathcal{D}^{f_A}$ of $f$ on attribute $A\in\mathcal{A}$. Specifically, we follow \cite{peyre2019computational, cuturi2013sinkhorn} to minimize the summation of the following Sinkhorn distance between these two distributions:
\begin{equation*}   L_{fair}=\sum_{A\in\mathcal{A}}\inf_{\gamma\in\Gamma(\mathcal{D}^f, \mathcal{D}^{f_A})}\big\{\mathbb{E}_{X\sim\gamma}[c(p,q)]+\alpha H(\gamma|\mu \otimes \nu)\big\},
\end{equation*}
where $\Gamma(\mathcal{D}^f, \mathcal{D}^{f_A})$ is the set of joint distributions based on $\mathcal{D}^f$ and $\mathcal{D}^{f_A}$. Let $p$ and $q$ be the points from $\mathcal{D}^f$ and $\mathcal{D}^{f_A}$, respectively. Then, $c(p,q)$ represents the transport cost \cite{cuturi2013sinkhorn}. Let $\mu$ and $\nu$ be the reference measures from the set of measures on $f$.
Then, $H(\gamma|\mu \otimes \nu)$ represents the relative entropy of $\gamma$ with respect to the product measure $\mu \otimes \nu$. $\alpha\geq 0$ is a regularization hyperparameter. In practice, we use the empirical form of $L_{fair}$.

\underline{Total Loss:} Therefore, the final learning objective becomes $\mathcal{L}(\theta)=L_{imb}+\lambda L_{fair}$, where $\lambda$ is a hyperparameter.

\begin{figure*}[t]
    \centering
    \includegraphics[width=1.0\textwidth]{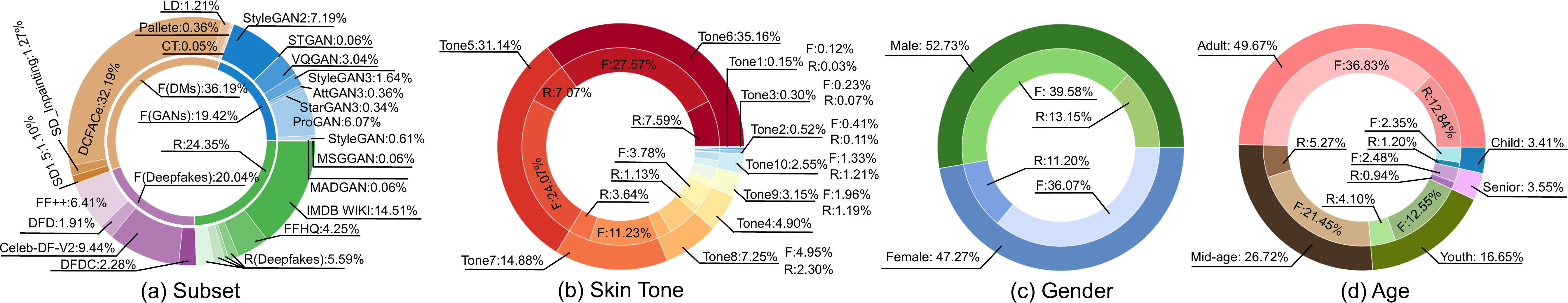}
    \vspace{-6mm}
    \caption{\small \textit{Distribution of face images of the AI-Face dataset. The figure shows the (a) subset distribution and the demographic distribution for (b) skin tone, (c) gender, and (d) gender. The outer rings in (b), (c), and (d) represent the proportion of groups within each attribute category, while the inner rings indicate the distribution of fake (F) and real (R) images within those groups.}}
    \label{fig:dataset_stas}
    \vspace{-6mm}
\end{figure*}

\smallskip
\noindent
\textbf{Train}. Traditional optimization methods like stochastic gradient descent can lead to poor model generalization due to sharp loss landscapes with multiple local and global minima. To address this, we use Sharpness-Aware Minimization (SAM)~\cite{foret2020sharpness} to enhance our annotator's generalization by flattening the loss landscape. Specifically, flattening is attained by determining the optimal $\epsilon^*$ for perturbing model parameters $\theta$ to maximize the loss, formulated as: $\epsilon^*=\arg\max_{\|\epsilon\|_2\leq \beta}{\mathcal{L}}\textbf{(}\theta+\epsilon \textbf{)}
\approx\arg\max_{\|\epsilon\|_2\leq \beta}\epsilon^\top\nabla_\theta \mathcal{L}=\beta\sign(\nabla_\theta \mathcal{L})$, where $\beta$ controls the perturbation magnitude. The approximation is based on the first-order Taylor expansion with assuming $\epsilon$ is small. The final equation is obtained by solving a dual norm problem, where $\sign$ represents a sign function and $\nabla_\theta \mathcal{L}$ being the gradient of $\mathcal{L}$ with respect to $\theta$. As a result, the model parameters are updated by solving: $\min_\theta \mathcal{L}\textbf{(}\theta+\epsilon^*\textbf{)}$. 

\smallskip
\noindent
\textbf{Inference}. We use the trained annotators to predict demographic labels for each image in AI-Face dataset, except for those from IMDB-WIKI, which already contain true labels. 
\vspace{-1mm}

\subsection{Dataset Statistics}
\vspace{-1mm}
Fig.~\ref{fig:dataset_stas} illustrates the subset distribution and demographic attributes of the AI-Face dataset. The dataset contains approximately three times more generated images than real images, with diffusion model-generated images constituting the majority. In terms of demographic attributes, the majorities in skin tone are Tone 5 (31.14\%) and Tone 6 (35.16\%). The lightest skin tones (Tones 1-3) are underrepresented, comprising only 0.97\% of the dataset.
The dataset is relatively balanced across gender. Adult (25-44) (49.67\%) is the predominant representation in age groups. 

\vspace{-1mm}
\subsection{Annotation Quality Assessment}
\vspace{-1mm}
To assess the quality of demographic annotations in our AI-Face dataset, we conducted a user study. Three participants label the demographic attributes for the given images (the details of labeling activities are in appendix~\ref{appendix:human_label}), with the final ground truth determined by majority vote. We then compare our annotations with those in A-FF++, A-DFDC, A-CelebDF-V2, and A-DFD datasets. Specifically, we perform two assessments: \textit{1)} \textit{Strategic comparison}: We select 1,000 images from A-FF++ and A-DFDC that have different annotations from AI-Face. These images likely represent challenging cases. \textit{2)} \textit{Random comparison}: We randomly sampled 1,000 images from A-Celeb-DF-V2 and A-DFD. Due to the limited age classes in these datasets, only gender was evaluated. The results, presented in Table~\ref{tab:Annotation_quality_assessment}, demonstrate the high correctness of the AI-Face annotations and their superior quality compared to the annotations of other datasets. For example, our annotation quality (ACC) surpasses those in A-FF++ by 78.714\%  on gender and 48.000\% on age. 

\vspace{-1mm}
\section{Fairness Benchmark Settings}\label{sec:fairness_benchmark}
\vspace{-1mm}
This section demonstrates the fairness benchmark settings for detection methods and evaluation metrics on AI-Face (80\%/20\%: Train/Test). More settings are in Appendix~\ref{appendix:fairness_implementation_details}.

\smallskip
\noindent
\textbf{Detection Methods}. Our benchmark has implemented 12 detectors. The methodologies cover a spectrum that is specifically tailored to detect AI-generated faces from Deepfake Videos, GANs, and DMs. They can be classified into four types: \textit{\underline{Naive detectors:}} refer to backbone models that can be directly utilized as the detector for binary classification, including CNN-based (\ie, Xception~\cite{chollet2017xception},  EfficientB4~\cite{tan2019efficientnet}) and transformer-based (\ie, ViT-B/16~\cite{dosovitskiy2020image}). \textit{\underline{Frequency-based:}} explore the frequency domain for forgery detection (\ie, F3Net~\cite{qian2020thinking}, SPSL~\cite{liu2021spatial}, SRM~\cite{luo2021generalizing}). \textit{\underline{Spatial-based:}} focus on mining spatial characteristics (\eg, texture) within images for detection (\ie, UCF~\cite{yan2023ucf}, UnivFD~\cite{ojha2023towards}, CORE~\cite{ni2022core}). \textit{\underline{Fairness-enhanced:}} focus on improving fairness in AI-generated face detection by designing specific algorithms (\ie, DAW-FDD~\cite{ju2024improving}, DAG-FDD~\cite{ju2024improving}, PG-FDD~\cite{lin2024preserving}).

\begin{table}[t]
    \centering
    \scalebox{0.6}{
\begin{tabular}{c|c|ccc|ccc}
\Xhline{4\arrayrulewidth}
\rowcolor[HTML]{E4E0E1} 
\cellcolor[HTML]{E4E0E1}                                                                                      & \cellcolor[HTML]{E4E0E1}                                   & \multicolumn{3}{c|}{\cellcolor[HTML]{E4E0E1}\textbf{Gender}} & \multicolumn{3}{c}{\cellcolor[HTML]{E4E0E1}\textbf{Age}} \\ \cline{3-8} 
\rowcolor[HTML]{E4E0E1} 
\multirow{-2}{*}{\cellcolor[HTML]{E4E0E1}\textbf{\begin{tabular}[c]{@{}c@{}}Evaluation \\ Type\end{tabular}}} & \multirow{-2}{*}{\cellcolor[HTML]{E4E0E1}\textbf{Dataset}} & \textbf{ACC}      & \textbf{Precision}   & \textbf{Recall}   & \textbf{ACC}     & \textbf{Precision}  & \textbf{Recall} \\ \hline \hline
                                                                                                              & A-FF++                                                     & 8.143             & 17.583               & 5.966             & 37.700           & 39.459              & 45.381          \\
                                                                                                              & AI-Face                                                    & \textbf{86.857}   & \textbf{74.404}      & \textbf{77.367}   & \textbf{85.700}  & \textbf{74.024}     & \textbf{63.751} \\ \cline{2-8} 
                                                                                                              & A-DFDC                                                     & 21.600              & 28.604               & 23.082            & 33.400           & 38.011              & 40.165          \\
\multirow{-4}{*}{\textbf{Strategic}}                                                                          & AI-Face                                                    & \textbf{91.700}     & \textbf{92.129}      & \textbf{83.448}   & \textbf{77.000}  & \textbf{76.184}     & \textbf{62.646} \\ \hline \hline
                                                                                                              & A-Celeb-DF-V2                                              & 89.628            & 90.626               & 90.494            & \multicolumn{3}{c}{-}                                    \\
                                                                                                              & AI-Face                                                    & \textbf{91.206}   & \textbf{91.474}      & \textbf{91.767}   & \multicolumn{3}{c}{-}                                    \\ \cline{2-8} 
                                                                                                              & A-DFD                                                      & 70.900            & 71.686               & 74.435            & \multicolumn{3}{c}{-}                                    \\
\multirow{-4}{*}{\textbf{Random}}                                                                             & AI-Face                                                    & \textbf{92.300}   & \textbf{91.060}      & \textbf{91.727}   & \multicolumn{3}{c}{-}                                    \\ \Xhline{4\arrayrulewidth}
\end{tabular}
}
\vspace{-2mm}
\caption{\small \textit{Annotation quality assessment results (\%) for A-FF++, A-DFDC, A-Celeb-DF-V2, A-DFD, and our AI-Face. ACC: Accuracy.}}
\vspace{-6mm}
\label{tab:Annotation_quality_assessment}
\end{table}

\newcommand{\coloredTextBox}[4]{
  \colorbox{#3}{\parbox[c][#2][c]{#1}{\centering #4}}
}
\definecolor{deepcolor}{HTML}{9AFF99}
\definecolor{mediumcolor}{HTML}{96FFFB}
\definecolor{lightcolor}{HTML}{FFFC9E}

\begin{table*}[t]
\vspace{-2mm}
\centering
\scalebox{0.64}{
\begin{tabular}{c|c|c|cccccccccccc}
\Xhline{4\arrayrulewidth}
\rowcolor[HTML]{E4E0E1} 
\cellcolor[HTML]{E4E0E1}                                   & \cellcolor[HTML]{E4E0E1}                                     & \cellcolor[HTML]{E4E0E1}                                  & \multicolumn{12}{c}{\cellcolor[HTML]{E4E0E1}\textbf{Model Type}}                                                                                                                                                                                                                                                                                                                                                                                  \\ \cline{4-15} 
\rowcolor[HTML]{E4E0E1} 
\cellcolor[HTML]{E4E0E1}                                   & \cellcolor[HTML]{E4E0E1}                                     & \cellcolor[HTML]{E4E0E1}                                  & \multicolumn{3}{c|}{\cellcolor[HTML]{E4E0E1}\textbf{Naive}}                                                                              & \multicolumn{3}{c|}{\cellcolor[HTML]{E4E0E1}\textbf{Frequency}}                                     & \multicolumn{3}{c|}{\cellcolor[HTML]{E4E0E1}\textbf{Spatial}}                                       & \multicolumn{3}{c}{\cellcolor[HTML]{E4E0E1}\textbf{Fairness-enhanced}}                     \\ \cline{4-15} 
\rowcolor[HTML]{E4E0E1} 
\multirow{-3}{*}{\cellcolor[HTML]{E4E0E1}\textbf{Measure}} & \multirow{-3}{*}{\cellcolor[HTML]{E4E0E1}\textbf{Attribute}} & \multirow{-3}{*}{\cellcolor[HTML]{E4E0E1}\textbf{Metric}} & \begin{tabular}[c]{@{}c@{}}Xception\\ ~\cite{chollet2017xception}\end{tabular}                               & \begin{tabular}[c]{@{}c@{}}EfficientB4\\~\cite{tan2019efficientnet}\end{tabular}                              & \multicolumn{1}{c|}{\cellcolor[HTML]{E4E0E1}\begin{tabular}[c]{@{}c@{}}ViT-B/16\\~\cite{dosovitskiy2020image}\end{tabular}} & \begin{tabular}[c]{@{}c@{}}F3Net\\~\cite{qian2020thinking}\end{tabular}  & \begin{tabular}[c]{@{}c@{}}SPSL\\~\cite{liu2021spatial}\end{tabular}                                    & \multicolumn{1}{c|}{\cellcolor[HTML]{E4E0E1}\begin{tabular}[c]{@{}c@{}}SRM\\~\cite{luo2021generalizing}\end{tabular}} & \begin{tabular}[c]{@{}c@{}}UCF\\~\cite{yan2023ucf}\end{tabular}    & \begin{tabular}[c]{@{}c@{}}UnivFD\\~\cite{ojha2023towards}\end{tabular}                                 & \multicolumn{1}{c|}{\cellcolor[HTML]{E4E0E1}\begin{tabular}[c]{@{}c@{}}CORE\\~\cite{ni2022core}\end{tabular}} & \begin{tabular}[c]{@{}c@{}}DAW-FDD\\~\cite{ju2024improving}\end{tabular} &\begin{tabular}[c]{@{}c@{}}DAG-FDD\\~\cite{ju2024improving}\end{tabular}                                & \begin{tabular}[c]{@{}c@{}}PG-FDD\\~\cite{lin2024preserving}\end{tabular}                                  \\ \hline \hline
                                                           &                                                              & $F_{MEO}$                                                      & 8.836                                  & 8.300                                   & \multicolumn{1}{c|}{6.264}                            & 19.938 & 8.055                                   & \multicolumn{1}{c|}{10.002}                      & 17.325 & \cellcolor[HTML]{E4E0E1}\textbf{2.577} & \multicolumn{1}{c|}{10.779}                       & 14.118  & 6.551                                  & 6.465                                   \\
                                                           &                                                              & $F_{DP}$                                                       & 9.751                                  & \cellcolor[HTML]{E4E0E1}\textbf{6.184}  & \multicolumn{1}{c|}{7.728}                            & 12.876 & 9.379                                   & \multicolumn{1}{c|}{10.897}                      & 12.581 & 8.556                                  & \multicolumn{1}{c|}{10.317}                       & 10.706  & 8.617                                  & 9.746                                   \\
                                                           &                                                              & $F_{OAE}$                                                      & 1.271                                  & 4.377                                   & \multicolumn{1}{c|}{2.168}                            & 2.818  & 1.135                                   & \multicolumn{1}{c|}{0.915}                       & 1.883  & 2.748                                  & \multicolumn{1}{c|}{1.332}                        & 1.667   & 1.388                                  & \cellcolor[HTML]{E4E0E1}\textbf{0.882}  \\
                                                           & \multirow{-4}{*}{Skin Tone}                                  & $F_{EO}$                                                       & 12.132                                 & 11.062                                  & \multicolumn{1}{c|}{8.813}                            & 23.708 & 9.789                                   & \multicolumn{1}{c|}{14.239}                      & 21.92  & \cellcolor[HTML]{E4E0E1}\textbf{5.536} & \multicolumn{1}{c|}{13.069}                       & 16.604  & 7.383                                  & 9.115                                   \\ \cline{2-15} 
                                                           &                                                              & $F_{MEO}$                                                      & 3.975                                  & 5.385                                   & \multicolumn{1}{c|}{5.104}                            & 4.717  & 4.411                                   & \multicolumn{1}{c|}{6.271}                       & 5.074  & 4.503                                  & \multicolumn{1}{c|}{5.795}                        & 5.510   & 5.910                                  & \cellcolor[HTML]{E4E0E1}\textbf{3.190}  \\
                                                           &                                                              & $F_{DP}$                                                       & 1.691                                  & 1.725                                   & \multicolumn{1}{c|}{1.344}                            & 1.864  & 1.827                                   & \multicolumn{1}{c|}{1.957}                       & 1.736  & \cellcolor[HTML]{E4E0E1}\textbf{1.190} & \multicolumn{1}{c|}{2.154}                        & 2.015   & 2.151                                  & 1.252                                   \\
                                                           &                                                              & $F_{OAE}$                                                      & \cellcolor[HTML]{E4E0E1}\textbf{0.975} & 1.487                                   & \multicolumn{1}{c|}{1.803}                            & 1.129  & 1.037                                   & \multicolumn{1}{c|}{1.772}                       & 1.451  & 1.622                                  & \multicolumn{1}{c|}{1.389}                        & 1.325   & 1.420                                  & 1.071                                   \\
                                                           & \multirow{-4}{*}{Gender}                                     & $F_{EO}$                                                       & 4.143                                  & 5.863                                   & \multicolumn{1}{c|}{6.031}                            & 4.870  & 4.534                                   & \multicolumn{1}{c|}{6.78}                        & 5.510  & 5.408                                  & \multicolumn{1}{c|}{5.931}                        & 5.696   & 6.066                                  & \cellcolor[HTML]{E4E0E1}\textbf{3.702}  \\ \cline{2-15} 
                                                           &                                                              & $F_{MEO}$                                                      & 27.883                                 & 6.796                                   & \multicolumn{1}{c|}{14.937}                           & 38.801 & 27.614                                  & \multicolumn{1}{c|}{24.843}                      & 47.500 & \cellcolor[HTML]{E4E0E1}\textbf{5.436} & \multicolumn{1}{c|}{33.882}                       & 45.466  & 15.229                                 & 14.804                                  \\
                                                           &                                                              & $F_{DP}$                                                       & 10.905                                 & 11.849                                  & \multicolumn{1}{c|}{11.839}                           & 14.906 & 11.232                                  & \multicolumn{1}{c|}{11.570}                      & 17.049 & 15.249                                 & \multicolumn{1}{c|}{12.564}                       & 14.106  & \cellcolor[HTML]{E4E0E1}\textbf{9.633} & 10.467                                  \\
                                                           &                                                              & $F_{OAE}$                                                      & 7.265                                  & \cellcolor[HTML]{E4E0E1}\textbf{2.856}  & \multicolumn{1}{c|}{6.838}                            & 10.116 & 7.270                                   & \multicolumn{1}{c|}{6.524}                       & 11.652 & 3.793                                  & \multicolumn{1}{c|}{8.760}                        & 11.878  & 5.533                                  & 5.009                                   \\
                                                           & \multirow{-4}{*}{Age}                                        & $F_{EO}$                                                       & 42.216                                 & \cellcolor[HTML]{E4E0E1}\textbf{10.300} & \multicolumn{1}{c|}{30.795}                           & 55.032 & 40.943                                  & \multicolumn{1}{c|}{38.528}                      & 67.545 & 14.148                                 & \multicolumn{1}{c|}{48.729}                       & 64.384  & 30.182                                 & 29.585                                  \\ \cline{2-15} 
                                                           &                                                              & $F_{MEO}$                                                      & 10.505                                 & 17.586                                  & \multicolumn{1}{c|}{9.384}                            & 21.369 & 10.379                                  & \multicolumn{1}{c|}{15.142}                      & 20.134 & \cellcolor[HTML]{E4E0E1}\textbf{6.119} & \multicolumn{1}{c|}{15.34}                        & 16.565  & 12.178                                 & 9.578                                   \\
                                                           &                                                              & $F_{DP}$                                                       & 14.511                                 & \cellcolor[HTML]{E4E0E1}\textbf{8.607}  & \multicolumn{1}{c|}{11.535}                           & 17.175 & 13.259                                  & \multicolumn{1}{c|}{15.186}                      & 17.03  & 14.026                                 & \multicolumn{1}{c|}{14.301}                       & 14.088  & 11.705                                 & 14.697                                  \\
                                                           &                                                              & $F_{OAE}$                                                      & 2.536                                  & 8.461                                   & \multicolumn{1}{c|}{4.928}                            & 4.870  & \cellcolor[HTML]{E4E0E1}\textbf{2.464}  & \multicolumn{1}{c|}{3.998}                       & 3.536  & 6.287                                  & \multicolumn{1}{c|}{2.775}                        & 3.547   & 4.035                                  & 3.062                                   \\
                                                           & \multirow{-4}{*}{Intersection}                               & $F_{EO}$                                                       & 24.315                                 & 25.114                                  & \multicolumn{1}{c|}{27.443}                           & 47.783 & 21.679                                  & \multicolumn{1}{c|}{30.112}                      & 43.376 & 20.255                                 & \multicolumn{1}{c|}{28.84}                        & 33.122  & 26.295                                 & \cellcolor[HTML]{E4E0E1}\textbf{18.348} \\ \cline{2-15} 
\multirow{-17}{*}{\textbf{Fairness(\%)\textdownarrow}}                       & Individual                                                   & $F_{IND}$                                                      &  10.338                                      &   25.742                                      & \multicolumn{1}{c|}{\cellcolor[HTML]{E4E0E1}\textbf{0.022}}                                 &  1.872      &    2.518                                     & \multicolumn{1}{c|}{7.621}                            & 0.767       &       3.523                                 & \multicolumn{1}{c|}{0.041}                             & 3.772        &0.901                                        &  0.780                                       \\ \hline
                                                           &                                                              & AUC\textuparrow                                                       & 98.583                                 & 98.611                                  & \multicolumn{1}{c|}{98.69}                            & 98.714 & 98.747                                  & \multicolumn{1}{c|}{97.936}                      & 98.082 & 98.192                                 & \multicolumn{1}{c|}{98.579}                       & 97.811  & 98.771                                 & \cellcolor[HTML]{E4E0E1}\textbf{99.172} \\
                                                           &                                                              & ACC\textuparrow                                                       & 96.308                                 & 94.203                                  & \multicolumn{1}{c|}{94.472}                           & 95.719 & \cellcolor[HTML]{E4E0E1}\textbf{96.346} & \multicolumn{1}{c|}{95.092}                      & 95.151 & 93.651                                 & \multicolumn{1}{c|}{96.224}                       & 95.426  & 95.722                                 & 96.174                                  \\
                                                           &                                                              & AP\textuparrow                                                        & 99.350                                 & 99.542                                  & \multicolumn{1}{c|}{99.571}                           & 99.453 & 99.356                                  & \multicolumn{1}{c|}{99.172}                      & 99.273 & 99.400                                 & \multicolumn{1}{c|}{99.360}                       & 99.015  & 99.498                                 & \cellcolor[HTML]{E4E0E1}\textbf{99.694} \\
                                                           &                                                              & EER\textdownarrow                                                       & 5.149                                  & 6.689                                   & \multicolumn{1}{c|}{6.372}                            & 5.256  & \cellcolor[HTML]{E4E0E1}\textbf{4.371}  & \multicolumn{1}{c|}{6.483}                       & 7.708  & 7.633                                  & \multicolumn{1}{c|}{5.145}                        & 7.063   & 5.499                                  & 4.961                                   \\
\multirow{-5}{*}{\textbf{Utility(\%)}}                         & \multirow{-5}{*}{-}                                          & FPR\textdownarrow                                                       & 12.961                                 & 20.066                                  & \multicolumn{1}{c|}{16.426}                           & 14.679 & 13.661                                  & \multicolumn{1}{c|}{15.746}                      & 13.646 & 18.550                                 & \multicolumn{1}{c|}{13.410}                       & 16.670  & 14.844                                 & \cellcolor[HTML]{E4E0E1}\textbf{10.971} \\ \hline
\multicolumn{3}{c|}{Training Time / Epoch}                                                                                                                                                                                      &\cellcolor[HTML]{E4E0E1}\textbf{1h15min}                                        &2h25min                                         & \multicolumn{1}{c|}{2h40min}                                 & 1h18min       & 1h20min                                        & \multicolumn{1}{c|}{3h10min}                            &5h05min        &4h                                        & \multicolumn{1}{c|}{1h16min}                             & 1h25min        &1h17min                                        &7h20min                                         \\\Xhline{4\arrayrulewidth}
\end{tabular}
}
\vspace{-2mm}
\caption{\small \textit{Overall performance comparison of difference methods on the AI-Face dataset. The best performance is shown in \textbf{bold}. }}
\label{tab:intra_domain_overall}
\vspace{-6mm}
\end{table*}

\smallskip
\noindent
\textbf{Evaluation Metrics}.
To provide a comprehensive benchmarking, we consider 5 fairness metrics commonly used in fairness community~\cite{han2024ffb, mehrabi2021survey, wang2022understanding, wang2023aleatoric, dwork2012fairness} and 5 widely used utility metrics. For \textit{\underline{fairness}} metrics, we consider Demographic Parity ($F_{DP}$)~\cite{han2024ffb, mehrabi2021survey}, Max Equalized Odds ($F_{MEO}$)~\cite{wang2023aleatoric}, Equal Odds ($F_{EO}$)~\cite{wang2022understanding}, and Overall Accuracy Equality ($F_{OAE}$)~\cite{wang2023aleatoric} for evaluating group (\eg, gender) and intersectional (\eg, individuals of a specific gender and simultaneously a specific skin tone) fairness. 
In experiments, the intersectional groups are Female-Light (F-L), Female-Medium (F-M), Female-Dark (Dark), Male-Light (M-L), Male-Medium (M-M), and Male-Dark (M-D), where we group 10 categories of skin tones into Light (Tone 1-3), Medium (Tone 4-6), and Dark (Tone 7-10) for simplicity according to \cite{MonkSkinToneScale}.
We also use individual fairness ($F_{IND}$)~\cite{dwork2012fairness, hufairness} (\ie, similar individuals should have similar predicted outcomes) for estimation. 
For \textit{\underline{utility}} metrics, we employ the Area Under the ROC Curve (AUC), Accuracy (ACC), Average Precision (AP), Equal Error Rate (EER), and False Positive Rate (FPR). 

\section{Results and Analysis}
\vspace{-1mm}
In this section, we estimate the existing AI-generated image detectors' fairness performance alongside their utility on our AI-Face Dataset. More results can be found in Appendix~\ref{appendix:more_results}. 

\subsection{General Fairness Comparison}
\textbf{Overall Performance}. 
Table~\ref{tab:intra_domain_overall}  reports the overall performance on our AI-Face test set. Our observations are: \textbf{1)} Fairness-Enhanced Models (specifically PG-FDD~\cite{lin2024preserving}) are the most effective in achieving both high fairness and utility, underscoring the effectiveness of specialized fairness-enhancement techniques in mitigating demographic biases.  \textbf{2)} UnivFD~\cite{ojha2023towards}, based on the CLIP backbone~\cite{openclip2021}, also achieves commendable fairness, suggesting that foundation models equipped with fairness-focused enhancements could be a promising direction for developing fairer detectors. \textbf{3)} Naive detectors, such as EfficientB4~\cite{tan2019efficientnet}, trained on large, diverse datasets (\eg, our AI-Face) can achieve competitive fairness and utility, highlighting the potential of fairness improvements by choosing specific architecture. 
\textbf{4)} 10 out of 12 detectors have an AUC higher than 98\%, demonstrating our AI-Face dataset is significant for training AI-face detectors in resulting high utility. 
\textbf{5)} PG-FDD demonstrates superior performance but has a long training time, which can be explored and addressed in the future.


\begin{figure*}[t]
    \centering
    \includegraphics[width=1\textwidth]{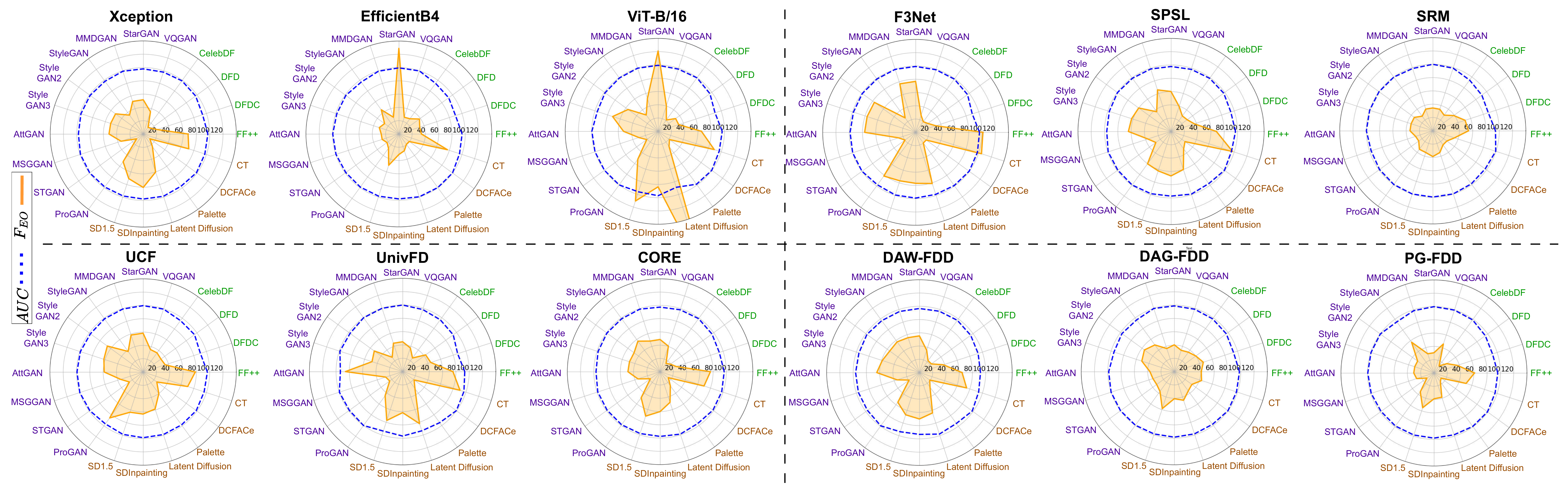}
    \vspace{-8mm}
    \caption{\small \textit{Visualization of the intersectional $F_{EO}$ (\%) and AUC (\%) of detectors on different subsets. The smaller $F_{EO}$ polygon area represents better fairness. The larger AUC area means better utility. }}
    \label{fig:radar_chart}
    \vspace{-4mm}
\end{figure*}

\begin{figure*}[t]
    \centering
    \includegraphics[width=1\textwidth]{Figs/intersectional_fpr.pdf}
    \vspace{-6mm}
    \caption{\small \textit{{FPR(\%) of each intersectional subgroup The dashline represents the lowest FPR on Female-Light (F-L) subgroup. }}}
    \label{fig:intersectional_fpr}
    \vspace{-6mm}
\end{figure*}

\smallskip
\noindent
\textbf{Performance on Different Subsets}. 
{\textbf{1)} Fig.~\ref{fig:radar_chart} demonstrates the intersectional $F_{EO}$ and AUC performance of detectors on each test subset. 
We observe that the fairness performance varies a lot among different generative methods for every detector.  
The largest bias on most detectors comes from detecting face images generated by diffusion models.
\textbf{2)} DAG-FDD~\cite{ju2024improving} and SRM~\cite{luo2021generalizing} demonstrate the most consistent fairness across subsets, indicating a robust handling of bias introduced by different generative methods.
\textbf{3)} Moreover, the stable utility demonstrates our dataset's expansiveness and diversity, enabling effective training to detect AI-generated faces from various generative techniques. 
}



\smallskip
\noindent
\textbf{Performance on Different Subgroups}. We conduct an analysis of all detectors on intersectional subgroups.
\textbf{1)} As shown in Fig.~\ref{fig:intersectional_fpr}, 
facial images with lighter skin tone are more often misclassified as fake,  likely due to the underrepresentation of lighter tones (Tone 1-3) in our dataset (see Fig. \ref{fig:dataset_stas} (b)). This suggests detectors tend to show higher error rates for minority groups.
\textbf{2)} Although gender representation is relatively balanced (see Fig. \ref{fig:dataset_stas} (c)) in our dataset, the detectors consistently exhibit higher false positive rates for female subgroups, indicating a persistent gender-based bias.


\subsection{Fairness Reliability Assessment}
\begin{figure*}[t]
    \centering
    \includegraphics[width=1\textwidth]{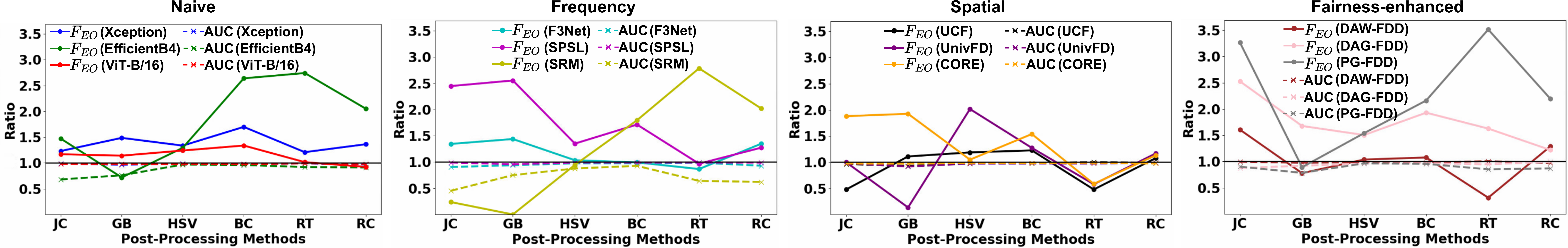}
    \vspace{-7mm}
    \caption{\small \textit{Performance ratio after vs. before post-processing. Points closer to 1.0 (\ie, no post-processing) indicate better robustness. }}
    \label{fig:robustness}
    \vspace{-2mm}
\end{figure*}

\vspace{-1mm}
\textbf{Fairness Robustness Evaluation}.
We apply 6 post-processing methods: Random Crop (RC)~\cite{cocchi2023unveiling}, Rotation (RT)~\cite{yan2023deepfakebench}, Brightness Contrast (BC)~\cite{yan2023deepfakebench}, Hue Saturation Value (HSV)~\cite{yan2023deepfakebench}, Gaussian Blur (GB)~\cite{yan2023deepfakebench}, and JEPG Compression (JC)~\cite{cozzolino2023raising} to the test images. Fig.~\ref{fig:robustness} shows each detector's intersectional $F_{EO}$ and AUC performance changes after using post-processing. Our observations are: \textbf{1)} These impairments tend to wash out forensic traces, so that detectors have evident performance degradation. \textbf{2)} Post-processing does not always cause detectors more bias (\eg, UCF, UnivFD, CORE, DAW-FDD have better fairness after rotation), though they hurt the utility.
\textbf{3)} Fairness-enhanced detectors struggle to maintain fairness when images undergo post-processing. \textbf{4)}
Spatial detectors have better fairness robustness compared with other model types. 

\begin{table*}[t]
\vspace{-1mm}
\centering
\scalebox{0.65}{
\begin{tabular}{c|c|cccccccc}
\Xhline{4\arrayrulewidth}
\rowcolor[HTML]{E4E0E1} 
\cellcolor[HTML]{E4E0E1}                                                                & \cellcolor[HTML]{E4E0E1}                                    & \multicolumn{8}{c}{\cellcolor[HTML]{E4E0E1}\textbf{Dataset}}                                                                                                                                                                                                                                                                                                              \\ \cline{3-10} 
\rowcolor[HTML]{E4E0E1} 
\cellcolor[HTML]{E4E0E1}                                                                & \cellcolor[HTML]{E4E0E1}                                    & \multicolumn{2}{c|}{\cellcolor[HTML]{E4E0E1}\textbf{CCv2~\cite{porgali2023casual}}}                                                                     & \multicolumn{3}{c|}{\cellcolor[HTML]{E4E0E1}\textbf{DF-Platter~\cite{narayan2023df}}}                                                                          & \multicolumn{3}{c}{\cellcolor[HTML]{E4E0E1}\textbf{GenData~\cite{teo2023measuring}}}                                 \\ \cline{3-10} 
\rowcolor[HTML]{E4E0E1} 
\cellcolor[HTML]{E4E0E1}                                                                & \cellcolor[HTML]{E4E0E1}                                    & \multicolumn{1}{c|}{\cellcolor[HTML]{E4E0E1}\textbf{Fairness(\%)\textdownarrow}} & \multicolumn{1}{c|}{\cellcolor[HTML]{E4E0E1}\textbf{Utility(\%)\textuparrow}} & \multicolumn{2}{c|}{\cellcolor[HTML]{E4E0E1}\textbf{Fairness(\%)\textdownarrow}}            & \multicolumn{1}{c|}{\cellcolor[HTML]{E4E0E1}\textbf{Utility(\%)\textuparrow}} & \multicolumn{2}{c|}{\cellcolor[HTML]{E4E0E1}\textbf{Fairness(\%)\textdownarrow}}            & \textbf{Utility(\%)\textuparrow} \\ \cline{3-10} 
\rowcolor[HTML]{E4E0E1} 
\multirow{-4}{*}{\cellcolor[HTML]{E4E0E1}\textbf{Model Type}}                           & \multirow{-4}{*}{\cellcolor[HTML]{E4E0E1}\textbf{Detector}} & \multicolumn{1}{c|}{\cellcolor[HTML]{E4E0E1}\textbf{$F_{OAE}$}}     & \multicolumn{1}{c|}{\cellcolor[HTML]{E4E0E1}\textbf{ACC}}     & \textbf{$F_{OAE}$} & \multicolumn{1}{c|}{\cellcolor[HTML]{E4E0E1}\textbf{$F_{EO}$}} & \multicolumn{1}{c|}{\cellcolor[HTML]{E4E0E1}\textbf{AUC}}     & \textbf{$F_{OAE}$} & \multicolumn{1}{c|}{\cellcolor[HTML]{E4E0E1}\textbf{$F_{EO}$}} & \textbf{AUC}     \\ \hline \hline
                                                                                        & Xception                                                    & \multicolumn{1}{c|}{1.006(\textcolor{red}{+0.031})}                             & \multicolumn{1}{c|}{86.465(-9.843)}                           & 6.836(+5.861) & \multicolumn{1}{c|}{9.789(+5.646)}                        & \multicolumn{1}{c|}{81.273(-17.310)}                          & 2.539(+1.564) & \multicolumn{1}{c|}{13.487(+9.344)}                       & 96.971(-1.612)   \\
                                                                                        & EfficientB4                                                 & \multicolumn{1}{c|}{4.077(+0.259)}                             & \multicolumn{1}{c|}{82.980(-11.223)}                          & 8.786(+7.299) & \multicolumn{1}{c|}{12.370(+6.507)}                       & \multicolumn{1}{c|}{67.694(-30.917)}                          & 3.304(+1.817) & \multicolumn{1}{c|}{\colorbox{highlightcolor}{\textbf{1.995}}(-3.686)}                        & 93.213(-5.398)   \\
\multirow{-3}{*}{\textbf{Naive}}                                                        & ViT-B/16                                                    & \multicolumn{1}{c|}{2.167(+0.364)}                             & \multicolumn{1}{c|}{81.489(-12.983)}                          & \colorbox{highlightcolor}{\textbf{0.015}}(-1.788) & \multicolumn{1}{c|}{12.373(+6.342)}                       & \multicolumn{1}{c|}{76.050(-22.640)}                          & 3.164(+1.361) & \multicolumn{1}{c|}{9.610(+3.579)}                        & 88.253(-10.437)  \\ \hline
                                                                                        & F3Net                                                       & \multicolumn{1}{c|}{5.743(+4.614)}                             & \multicolumn{1}{c|}{87.867(-7.852)}                           & 3.521(+2.392) & \multicolumn{1}{c|}{\colorbox{highlightcolor}{\textbf{6.445}}(+1.575)}                        & \multicolumn{1}{c|}{85.112(\textcolor{red}{-13.602})}                          & 1.188(+0.059) & \multicolumn{1}{c|}{16.306(+11.436)}                      & 91.603(-7.111)   \\
                                                                                        & SPSL                                                        & \multicolumn{1}{c|}{\colorbox{highlightcolor}{\textbf{0.601}}(-0.436)}                            & \multicolumn{1}{c|}{80.006(-16.340)}                          & 5.109(+4.072) & \multicolumn{1}{c|}{7.842(+3.308)}                        & \multicolumn{1}{c|}{82.175(-16.572)}                          & 1.385(+0.348) & \multicolumn{1}{c|}{9.261(+4.272)}                        & \colorbox{highlightcolor}{\textbf{98.838}}(+0.091)    \\
\multirow{-3}{*}{\textbf{Frequency}}                                                    & SRM                                                         & \multicolumn{1}{c|}{7.000(+5.228)}                             & \multicolumn{1}{c|}{79.768(-15.324)}                          & 3.823(+2.051) & \multicolumn{1}{c|}{6.567(\textcolor{red}{-0.213})}                        & \multicolumn{1}{c|}{66.401(-31.535)}                          & 3.281(+1.509) & \multicolumn{1}{c|}{7.907(\textcolor{red}{+1.127})}                        & 90.049(-7.887)   \\ \hline
                                                                                        & UCF                                                         & \multicolumn{1}{c|}{2.169(+0.718)}                             & \multicolumn{1}{c|}{\colorbox{highlightcolor}{\textbf{93.009}}(\textcolor{red}{-2.142})}                           & 8.687(+7.236) & \multicolumn{1}{c|}{17.068(+11.558)}                      & \multicolumn{1}{c|}{80.821(-17.261)}                          & 3.513(+2.062) & \multicolumn{1}{c|}{10.529(+5.019)}                       & 87.778(-10.304)  \\
                                                                                        & UnivFD                                                      & \multicolumn{1}{c|}{7.625(+6.003)}                             & \multicolumn{1}{c|}{67.983(-25.668)}                          & 4.540(+2.918) & \multicolumn{1}{c|}{9.950(+4.542)}                        & \multicolumn{1}{c|}{76.443(-21.749)}                          & 1.645(\textcolor{red}{+0.023}) & \multicolumn{1}{c|}{3.848(-1.560)}                        & 94.418(-3.774)   \\
\multirow{-3}{*}{\textbf{Spatial}}                                                      & CORE                                                        & \multicolumn{1}{c|}{4.410(+3.021)}                             & \multicolumn{1}{c|}{83.328(-12.896)}                          & 7.741(+6.352) & \multicolumn{1}{c|}{17.348(+11.417)}                      & \multicolumn{1}{c|}{77.226(-21.353)}                          & 3.759(+2.370) & \multicolumn{1}{c|}{23.289(+17.358)}                      & 98.408(-0.171)   \\ \hline
                                                                                        & DAW-FDD                                                     & \multicolumn{1}{c|}{4.726(+3.401)}                             & \multicolumn{1}{c|}{84.685(-10.741)}                          & 5.536(+4.211) & \multicolumn{1}{c|}{13.667(+7.791)}                       & \multicolumn{1}{c|}{81.807(-16.004)}                          & 1.443(+0.118) & \multicolumn{1}{c|}{10.228(+4.532)}                       & 97.854(\textcolor{red}{+0.043})    \\
                                                                                        & DAG-FDD                                                     & \multicolumn{1}{c|}{2.364(+0.944)}                             & \multicolumn{1}{c|}{83.918(-11.804)}                          & 3.064(\textcolor{red}{+1.644}) & \multicolumn{1}{c|}{22.203(+16.137)}                      & \multicolumn{1}{c|}{75.206(-23.565)}                          & \colorbox{highlightcolor}{\textbf{0.714}}(-0.706) & \multicolumn{1}{c|}{10.332(+4.266)}                       & 92.108(-6.663)   \\
\multirow{-3}{*}{\textbf{\begin{tabular}[c]{@{}c@{}}Fairness-\\ enhanced\end{tabular}}} & PG-FDD                                                      & \multicolumn{1}{c|}{1.513(+0.442)}                             & \multicolumn{1}{c|}{92.852(-3.322)}                           & 4.565(+3.494) & \multicolumn{1}{c|}{9.717(+6.015)}                        & \multicolumn{1}{c|}{\colorbox{highlightcolor}{\textbf{85.271}}(-13.901)}                          & 3.063(+1.992) & \multicolumn{1}{c|}{9.479(+5.777)}                        & 93.329(-5.843)   \\ \Xhline{4\arrayrulewidth}
\end{tabular}
}
\vspace{-3mm}
\caption{\small \textit{Fairness generalization results based on the gender attribute. 
The smallest performance changes (in parentheses) and the best performance are in \textcolor{red}{red} and in \textbf{bold}, respectively. Only $F_{OAE}$ fairness metric and ACC metric are used in CCv2 due to all samples are real.}}
\label{tab:fairness_generalization}
\vspace{-2mm}
\end{table*}

\begin{figure*}[h!]
    \centering
    \includegraphics[width=1\textwidth]{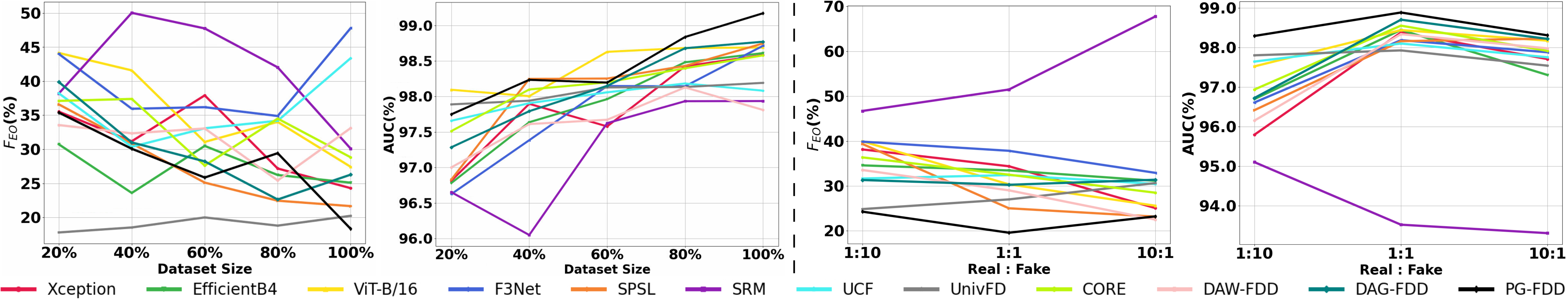}
  \vspace{-7mm}
  \caption{\small \textit{Impact of the training set size (\textbf{Left}) and the ratio of real and fake (\textbf{Right}) on detectors' intersectional $F_{EO}$(\%) and AUC (\%).}}
  \vspace{-6mm}
  \label{fig:datasize_realfake_ratio}
\end{figure*}

\smallskip
\noindent
\textbf{Fairness Generalization Evaluation}. 
{To evaluate detectors' fairness generalization capability, we test them on Casual Conversations v2 (CCv2)~\cite{porgali2023casual}, DF-Platter~\cite{narayan2023df}, and GenData~\cite{teo2023measuring}, none of which are part of AI-Face. Notably, CCv2 is a dataset that contains only real face images with demographic annotations (\eg, gender)  self-reported by the participants. 
Results on gender attribute in  Table~\ref{tab:fairness_generalization} show that: 
\textbf{1)} Even well-designed detectors that focus on improving utility or fairness generalization (\eg, UCF, PG-FDD) struggle to achieve consistently superior performance across different dataset domains. 
This highlights the remaining fairness generalization issue.
\textbf{2)} DAW-FDD and PG-PDD are two fairness-enhanced detectors that require accessing demographic information during training, but their fairness does not encounter a drastic drop when evaluating on CCv2. This reflects the high accuracy of the annotations in our AI-face. 
}

\smallskip
\noindent
\textbf{Effect of Training Set Size}.
We randomly sample 20\%, 40\%, 60\%, and 80\% of each training subset from AI-Face to assess the impact of training size on performance. Key observations from  Fig.~\ref{fig:datasize_realfake_ratio} (Left):
\textbf{1)} Among all detectors, UnivFD demonstrates the most stable fairness and utility performance as the training dataset size changes, likely due to its fixed CLIP backbone.
\textbf{2)} Increasing the training dataset size generally improves model utility, but this pattern does not extend to fairness metrics. In fact, certain detectors such as F3Net and UCF exhibit worsening fairness as the training size reaches its maximum. This suggests that more training data does not necessarily lead to fairer detectors. 

\smallskip
\noindent
\textbf{Effect of the Ratio of Real and Fake}. To examine how training real-to-fake sample ratios affect detector performance, we set the ratios at 1:10, 1:1, and 10:1 while keeping the total sample count constant. Experimental results in Fig.~\ref{fig:datasize_realfake_ratio} (Right) show: 
\textbf{1)} Most detectors' fairness improves as real sample representation increases. Probably because increasing real and reducing fake samples helps detectors reduce overfitting to artifacts specific to fake samples. This makes it easier for detectors to distinguish real from fake, even for underrepresented groups, thereby enhancing fairness.
\textbf{2)} Most detectors achieve the highest AUC with balanced data. 

\subsection{Discussion} \label{subsec:Discussion}
\vspace{-2mm}
According to the above experiments, we summarize the unsolved fairness problems in recent detectors: 
\textbf{1)} Detectors' fairness is unstable when detecting face images generated by different generative methods, indicating a future direction for enhancing fairness stability since new generative models continue to emerge. 
\textbf{2)} Even though fairness-enhanced detectors exhibit small overall fairness metrics, they still show biased detection towards minority groups. Future studies should be more cautious when designing fair detectors to ensure balanced performance across all demographic groups. 
\textbf{3)} There is currently no reliable detector, as all detectors experience severe large performance degradation under image post-processing and cross-domain evaluation.
Future studies should aim to develop a unified framework that ensures fairness, robustness, and generalization, as these three characteristics are essential for creating a reliable detector. Moreover, integrating foundation models (\eg, CLIP) into detector design may help mitigate bias.

\section{Conclusion}\label{sec:discussion}
\vspace{-1mm}
This work presents the \textit{first} demographically annotated million-scale AI-Face dataset, serving as a pivotal foundation for addressing the urgent need for developing fair AI face detectors. Based on this dataset, we conduct the \textit{first} comprehensive fairness benchmark, shedding light on the fairness performance and challenges of current representative AI face detectors.
Our findings can inspire and guide researchers in refining current models and exploring new methods to mitigate bias. \textbf{Limitation and Future Work:} One limitation is that our dataset’s annotations are algorithmically generated, so they may lack 100\% accuracy. This challenge is difficult to resolve, as demographic attributes for most AI-generated faces are often too ambiguous to predict and do not map to real-world individuals. We plan to enhance annotation quality through human labeling in the future. We also plan to extend our fairness benchmark to evaluate large language models like LLaMA2~\cite{touvron2023llama} and GPT4~\cite{achiam2023gpt} for detecting AI faces. \textbf{Social Impact:}
Malicious users could misuse AI-generated face images from our dataset to create fake social media profiles and spread misinformation. To mitigate this risk, only users who submit a signed end-user license agreement will be granted access to our dataset.
\vspace{-2mm}

\section*{Ethics Statement}
\vspace{-2mm}
\textbf{Our dataset collection and annotation generation are approved by Purdue's Institutional Review Board.}
The dataset is only for research purposes. 
All data included in this work are sourced from publicly available datasets, and we strictly comply with each dataset’s license agreement to ensure lawful inclusion and permissible secondary use for training and testing.  All collected data and their associated licenses are mentioned in the Datasheet of AI-Face in Appendix~\ref{appendix:ai_face_datasheet}. Our annotation processes prioritize ethical considerations: \textit{1)} 76\% images we annotated are generated facial images, ensuring no potential for harm to any individual. \textit{2)} For real images, we only provide annotations for content either licensed by the original copyright holders or explicitly stated as freely shareable for research purposes.

\section*{Acknowledgments}
\vspace{-2mm}
This work is supported by the U.S. National Science Foundation (NSF) under grant IIS-2434967 and the National Artificial Intelligence Research Resource (NAIRR) Pilot and TACC Lonestar6.
The views, opinions and/or findings expressed are those of the author and should not be interpreted as representing the official views or policies of NSF and NAIRR Pilot.

{
    \small
    \bibliographystyle{ieeetr}
    \bibliography{main}
}

\newpage
\input{Appendix}

\end{document}

%% file: preamble.tex
\usepackage[dvipsnames]{xcolor}


%% file: Appendix.tex
\onecolumn
\appendix
\numberwithin{equation}{section}
\numberwithin{theorem}{section}
\numberwithin{figure}{section}
\numberwithin{table}{section}
\renewcommand{\thesection}{{\Alph{section}}}
\renewcommand{\thesubsection}{\Alph{section}.\arabic{subsection}}
\renewcommand{\thesubsubsection}{\Roman{section}.\arabic{subsection}.\arabic{subsubsection}}

\def\p{\mathbf{p}}
\def\v{\mathbf{v}}
\def\u{\mathbf{u}}

\begin{center}
\textbf{\Large Appendix}
\end{center}

\section{The Definition of Demographic Categories}\label{appendix:demographic_definition}

\textbf{Skin Tone:} Skin tone is an important attribute of human appearance, with significant variation from pale to dark. Recently, AI systems, especially computer vision models, have become controversial over concerns about the potential bias of performance varying based on skin tone~\cite{buolamwini2018gender, krishnapriya2020issues, lu2019experimental}. Additionally, existing research has pointed out that skin tone annotations can be potentially less biased than building a racial classifier~\cite{khan2021one}. And the ethnicity attribute is subjective and can conceptually cause confusion in many aspects; for example, there may be no difference in facial appearance of African-American and African people, although, they may be referred to with two distinct racial categories. We, therefore, have opted to annotate the apparent skin tone of each face image. The Monk Skin Tone Scale~\cite{google_skintone} is developed specifically for the computer vision use case. We intentionally use the Monk Skin Tone scale over the Fitzpatrick skin type~\cite{sachdeva2009fitzpatrick}, which is developed as means for determining one's likelihood of getting sunburn and lacks variance in darker skin tones~\cite{howard2021reliability, okoji2021equity}. Additionally, Fitzpatrick skin type has been shown to be unreliable for image annotation~\cite{groh2022towards}.

\smallskip
\noindent
\textbf{Gender:} Many governments~\cite{us_department_of_state_2022, australian_bureau_statistics_2024} have adopted binary gender (\ie, Man/Male (M) and Woman/Female(F), defined as sex at birth, as a common choice for legal and institutional systems and official documents. Most facial recognition research~\cite{cook2019demographic, howard2019effect, raji2019actionable} also considers binary genders in their analyses. Our AI-Face dataset adopts binary gender as gender attributes.

\smallskip
\noindent
\textbf{Age:} Follow United Nations~\cite{united_nations_1982} and  Statistics Canada~\cite{statistics_canada_2017_age},  we have five distinct perceived age groups- Child (0-14), Youth (15-24), Adults (25-44), Middle-age Adults (45-64), and Seniors (65+). 

\begin{figure*}[t]
    \centering
    \includegraphics[width=0.6\textwidth]{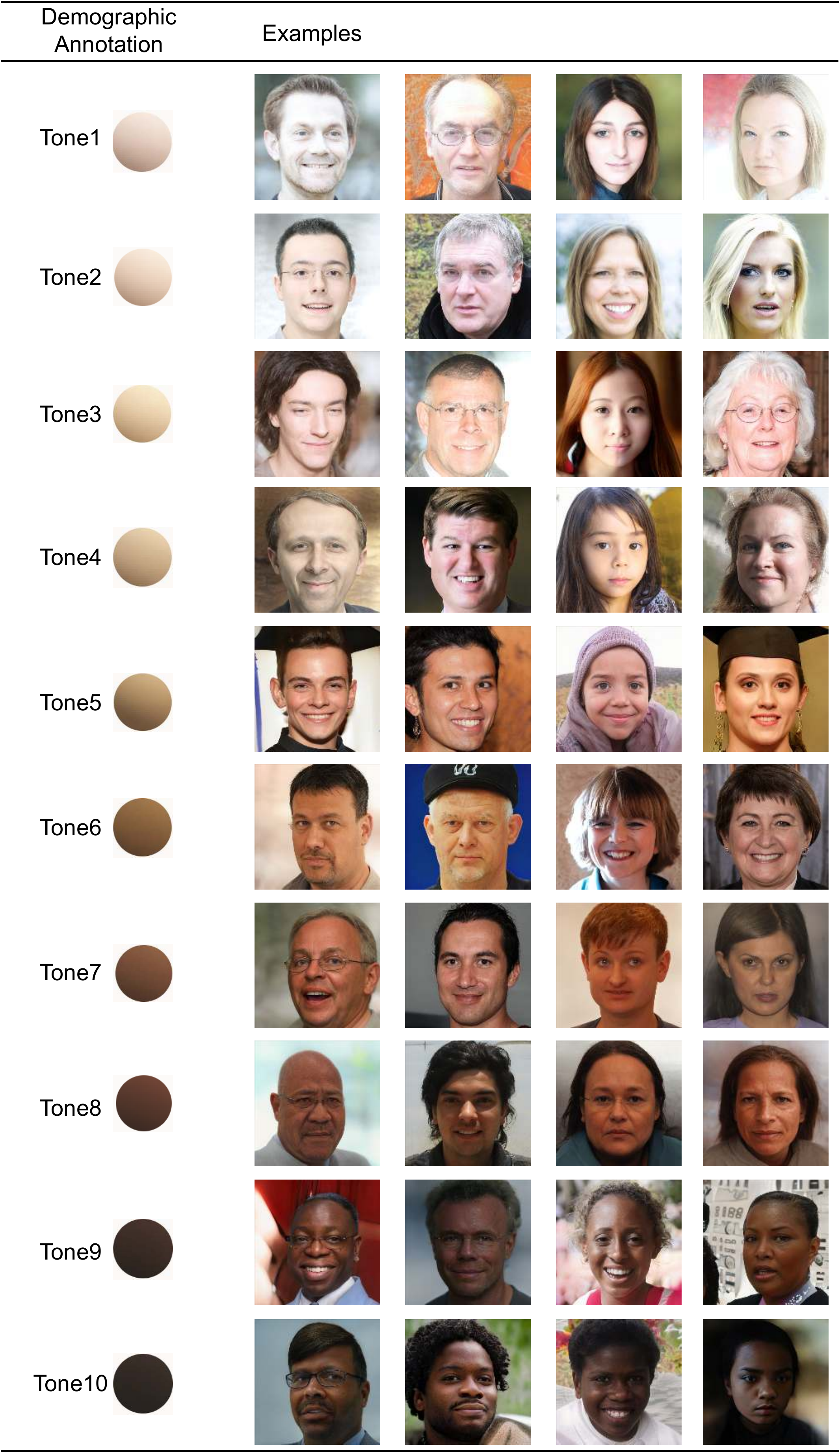}
    \vspace{-2mm}
    \caption{\textit{ Demographic annotation definition and examples of \textbf{skin tone} attribute.}}
    \label{fig:skintone_example}
\end{figure*}

\begin{figure*}[t]
    \centering
    \includegraphics[width=0.6\textwidth]{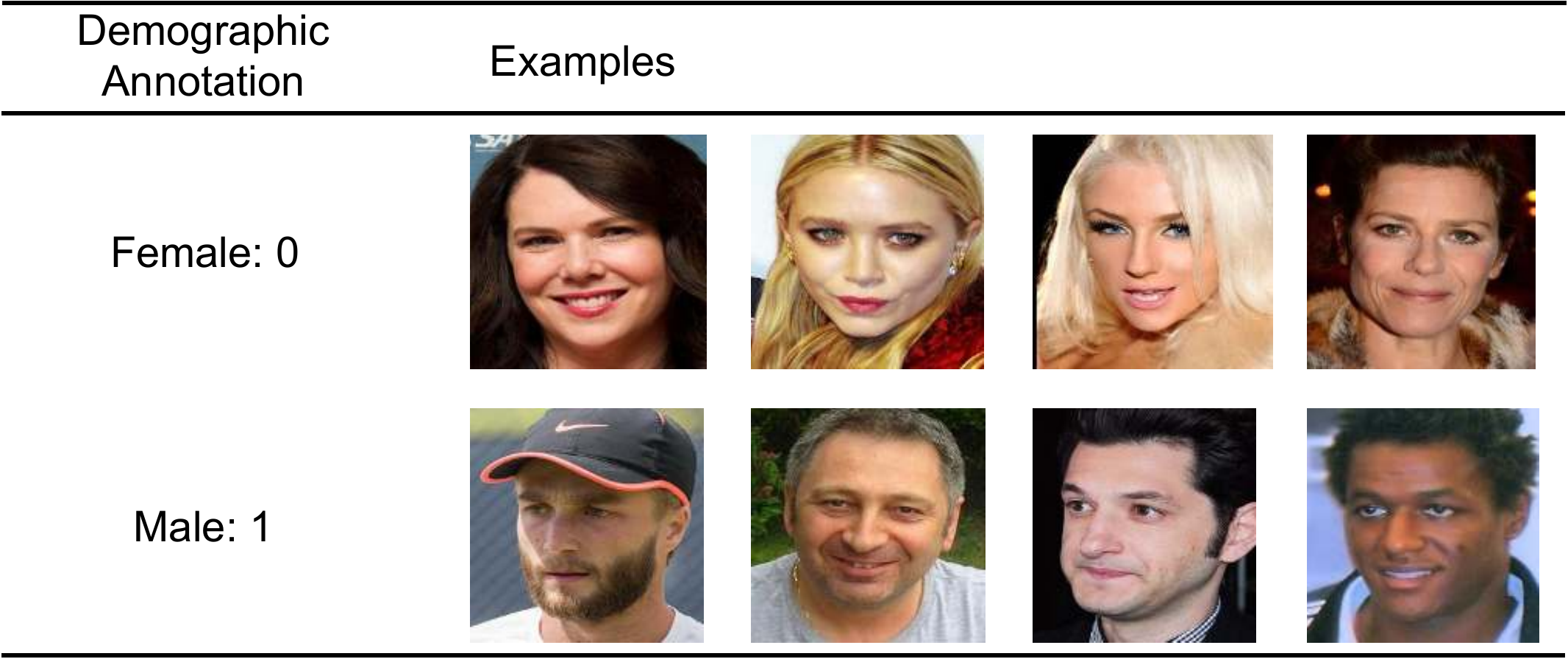}
    \vspace{-2mm}
    \caption{\textit{ Demographic annotation definition and examples of \textbf{gender} attribute.}}
    \label{fig:gender_example}
\end{figure*}

\begin{figure*}[t]
    \centering
    \includegraphics[width=0.6\textwidth]{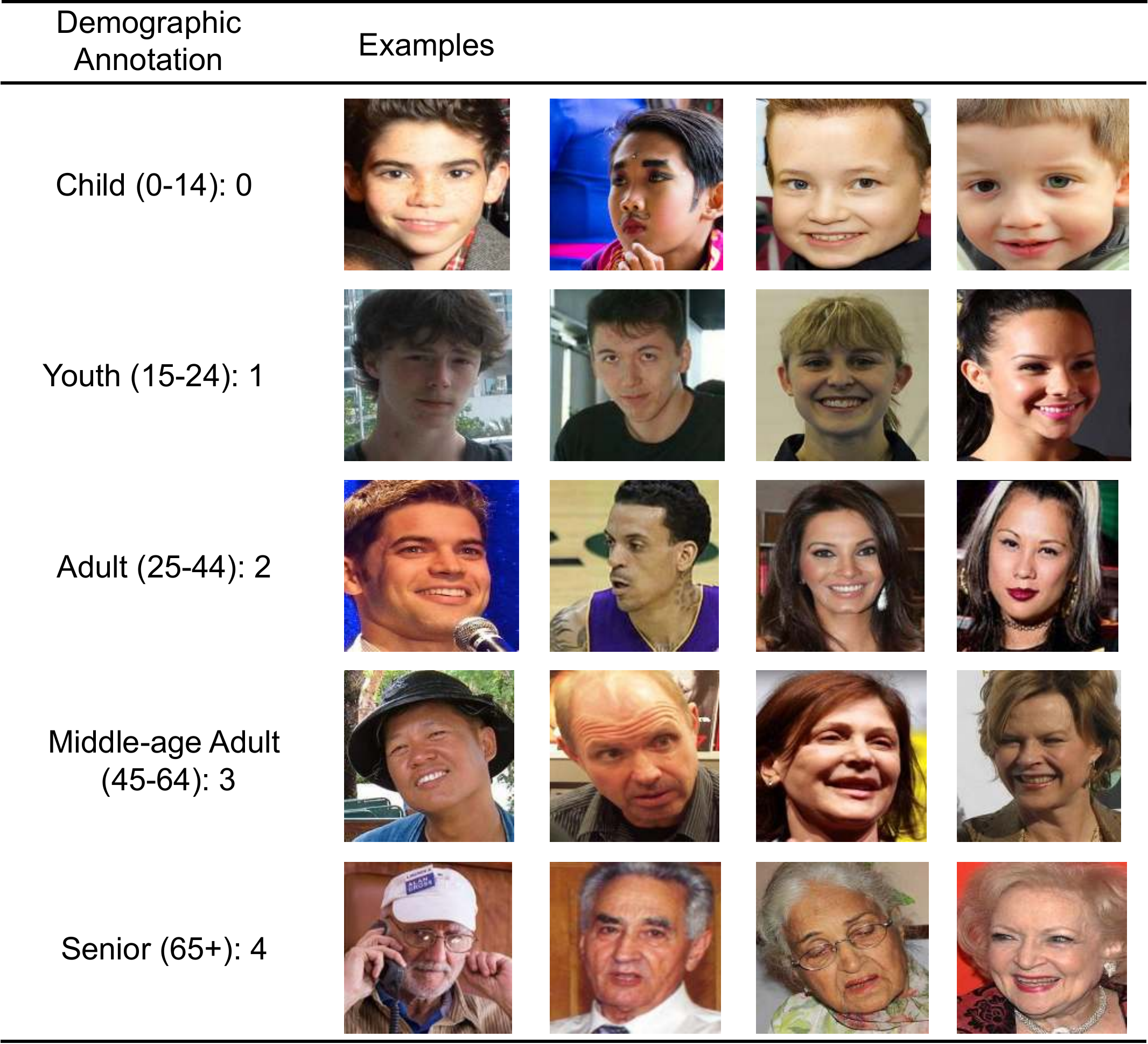}
    \vspace{-2mm}
    \caption{\textit{ Demographic annotation definition and examples of \textbf{age} attribute.}}
    \label{fig:age_example}
\end{figure*}

\smallskip
\noindent
The demographic attribute and its corresponding example images are shown from Fig.~\ref{fig:skintone_example} to Fig.~\ref{fig:age_example}.

\section{The Details of Demographically Annotated AI-Face Dataset} \label{appendix:details_dataset_overview}
\begin{figure*}[t]
    \centering
    \includegraphics[width=1\textwidth]{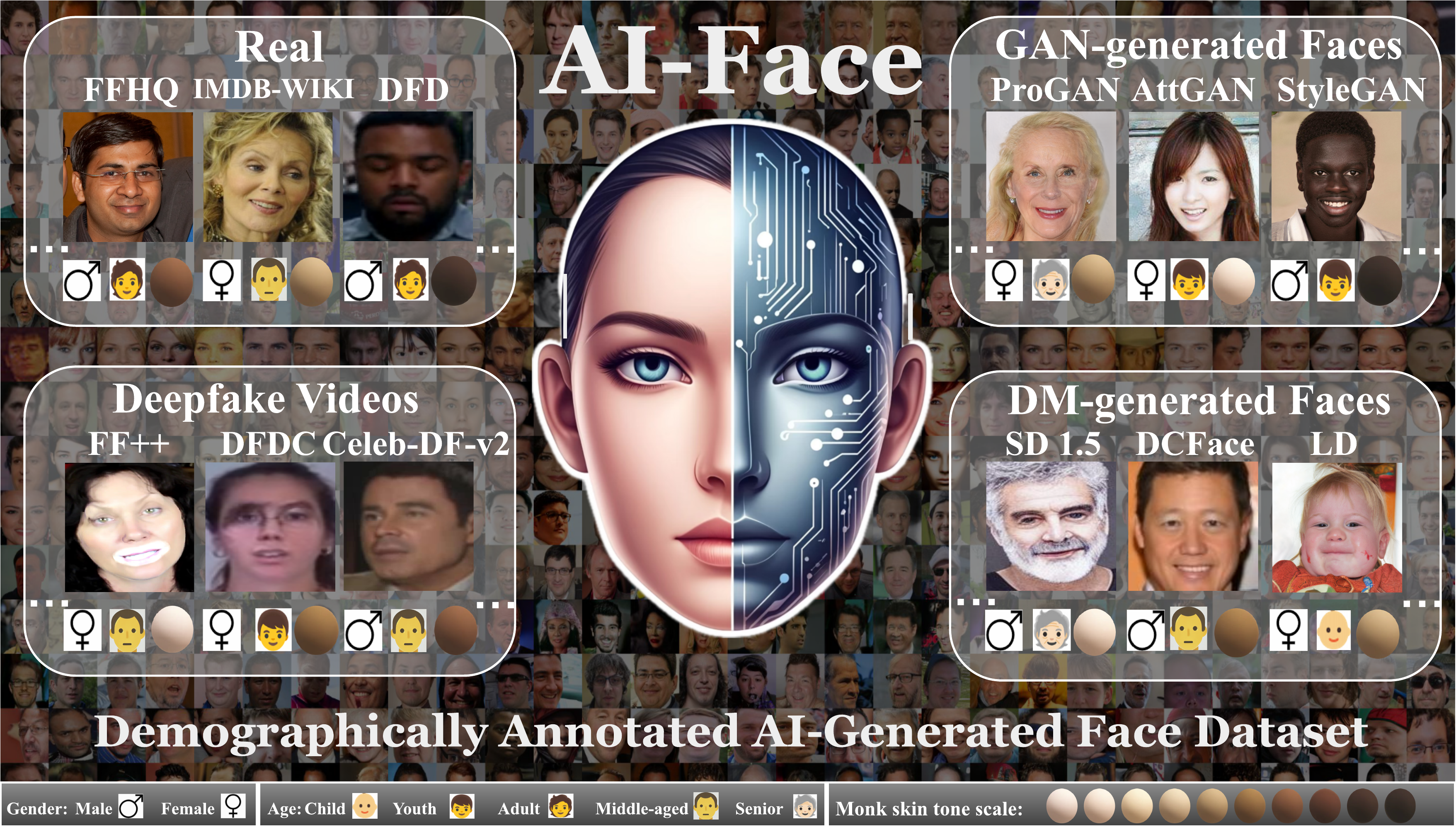}
    \vspace{-5mm}
    \caption{\small \textit{Overview of AI-Face dataset. Each face has three demographic annotations.}}
    \label{fig:dataset_overview}
    \vspace{-2mm}
\end{figure*}
\subsection{Detailed Information of Datasets} \label{appendix:dataset_detailed_info}
We build our AI-Face dataset by collecting and integrating public real and AI-generated face images sourced from academic publications, GitHub repositories, and commercial tools. We strictly adhere to the license agreements of all datasets to ensure that they allow inclusion in our datasets and secondary use for training and testing. Table~\ref{tab:dataset_info_detail} shows the detailed information of each dataset we used in our AI-Face, including the number of samples, the link for downloading the dataset, the accessibility, and their licenses.

\begin{table*}[t]
\centering
\scalebox{0.58}{
\begin{tabular}{c|c|c|c|c}
\hline
Dataset          & \#Samples & Link                                                                                               & Access                                                                                                                                                                                                                                                                                             & License                               \\ \hline
\begin{tabular}[c]{@{}c@{}}FF++\\ ~\cite{rossler2019faceforensics++}\end{tabular}             & 127K      & {\footnotesize\url {https://github.com/ondyari/FaceForensics/tree/master/dataset}}                                 & \begin{tabular}[c]{@{}c@{}}freely shared for a research purpose,\\  submit aggreement\end{tabular}                                                                                                                                                                                                 & Non Commercial                        \\ \hline
\begin{tabular}[c]{@{}c@{}}DFDC\\~\cite{dolhansky2020deepfake}\end{tabular}               & 75K       & {\footnotesize\url {https://www.kaggle.com/c/deepfake-detection-challenge/data}}                                   &                                                                                                                                                                                                                                                                                                    & Unknown                               \\ \cline{1-3} \cline{5-5} 
\begin{tabular}[c]{@{}c@{}}DFD\\~\cite{google2019deepfake}\end{tabular}              & 40K       & {\footnotesize\url {https://research.google/blog/contributing-data-to-deepfake-detection-research/}}               & \multirow{-2}{*}{\begin{tabular}[c]{@{}c@{}}the rights have been cleared for real videos, \\ submit aggreement\end{tabular}}                                                                                                                                                                       & Non Commercial                        \\ \hline
\begin{tabular}[c]{@{}c@{}}Celeb-DF-v2\\~\cite{li2020celeb}\end{tabular}      & 179K      & {\footnotesize\url {https://cse.buffalo.edu/$\sim$siweilyu/celeb-deepfakeforensics.html}}                          & \begin{tabular}[c]{@{}c@{}}freely shared for a research purpose,\\  submit aggreement\end{tabular}                                                                                                                                                                                                 & Non Commercial                        \\ \hline
\begin{tabular}[c]{@{}c@{}}AttGAN\\~\cite{giudice2021fighting}\end{tabular}           & 6K        &                                                                                                    &                                                                                                                                                                                                                                                                                                    &                                       \\ \cline{1-2}
\begin{tabular}[c]{@{}c@{}}StarGAN\\~\cite{giudice2021fighting}\end{tabular}          & 5.6K      &                                                                                                    &                                                                                                                                                                                                                                                                                                    &                                       \\ \cline{1-2}
\begin{tabular}[c]{@{}c@{}}StyleGAN\\~\cite{giudice2021fighting}\end{tabular}         & 10K       & \multirow{-3}{*}{{\footnotesize\url {https://iplab.dmi.unict.it/mfs/Deepfakes/PaperGANDCT-2021/}}}                 & \multirow{-3}{*}{\begin{tabular}[c]{@{}c@{}}Online dataset, download directly, \\ no license or agreement to sign\end{tabular}}                                                                                                                                                                    & \multirow{-3}{*}{Unknown}             \\ \hline
\begin{tabular}[c]{@{}c@{}}StyleGAN2\\~\cite{david_beniaguev_2022_SFHQ}\end{tabular}         & 118K      & {\url {https://github.com/SelfishGene/SFHQ-dataset}}                                                  & \begin{tabular}[c]{@{}c@{}}Since all images in this dataset are synthetically\\  generated there are no privacy issues or \\ license issues surrounding these images.\end{tabular}                                                                                                                 & MIT License                           \\ \hline
\begin{tabular}[c]{@{}c@{}}StyleGAN3\\~\cite{lu2023seeing}\end{tabular}        & 26.7K     & {\footnotesize\url {https://huggingface.co/datasets/InfImagine/FakeImageDataset}}                                  & \begin{tabular}[c]{@{}c@{}}This dataset are fully open for academic research \\ and can be used for commercial purposes\\  with official written permission.\end{tabular}                                                                                                                          & {\color[HTML]{333333} Apache-2.0}     \\ \hline
\begin{tabular}[c]{@{}c@{}}MMDGAN\\~\cite{asnani2023reverse}\end{tabular}            & 1K        &                                                                                                    &                                                                                                                                                                                                                                                                                                    &                                       \\ \cline{1-2}
\begin{tabular}[c]{@{}c@{}}MSGGAN\\~\cite{asnani2023reverse}\end{tabular}            & 1K        &                                                                                                    &                                                                                                                                                                                                                                                                                                    &                                       \\ \cline{1-2}
\begin{tabular}[c]{@{}c@{}}STGAN\\~\cite{asnani2023reverse}\end{tabular}            & 1K        &  \multirow{-3}{*}{{\footnotesize\url {https://github.com/vishal3477/Reverse\_Engineering\_GMs/blob/main/dataset/}}}                                                                                                  &  \multirow{-3}{*}{\begin{tabular}[c]{@{}c@{}}The dataset can be used for research purposes\\  only and can be used for commercial purposes\\  with official written permission.\end{tabular}}                                                                                                                                                                                                                                                                                                                                                  &      \multirow{-3}{*}{Non Commercial}                                  \\ \hline
\begin{tabular}[c]{@{}c@{}}ProGAN\\~\cite{dang2018deep}\end{tabular}           & 100K      & \footnotesize\url{https://drive.google.com/drive/folders/1jU-hzyvDZNn_M3ucuvs9xxtJNc9bPLGJ} &     \begin{tabular}[c]{@{}c@{}}Online dataset, download directly, \\ no license or agreement to sign\end{tabular}                                                   &   Unknown   \\ \hline
\begin{tabular}[c]{@{}c@{}}VQGAN\\~\cite{esser2021taming}\end{tabular}            & 50K       & {\footnotesize\url {https://github.com/awsaf49/artifact}}                                                          & \begin{tabular}[c]{@{}c@{}}This dataset comes from ArtiFact dataset,\\  which dataset takes leverage of data\\  from multiple methods thus different parts \\ of the dataset come with different licenses.\end{tabular}                                                                            & MIT License                           \\ \hline
\begin{tabular}[c]{@{}c@{}}DALLE2\\~\cite{wang2023dire}\end{tabular}           & 204       &                                                                                                    &                                                                                                                                                                                                                                                                                                    &                                       \\ \cline{1-2}
\begin{tabular}[c]{@{}c@{}}IF\\~\cite{wang2023dire}\end{tabular}               & 505       &                                                                                                    &                                                                                                                                                                                                                                                                                                    &                                       \\ \cline{1-2}
\begin{tabular}[c]{@{}c@{}}Midjourney\\~\cite{wang2023dire}\end{tabular}       & 100       & \multirow{-3}{*}{{\footnotesize\url {https://github.com/ZhendongWang6/DIRE}}}                                      & \multirow{-3}{*}{freely shared for a research purpose}                                                                                                                                                                                                                                             &                                       \\ \cline{1-4}
\begin{tabular}[c]{@{}c@{}}DCFACe\\~\cite{kim2023dcface}\end{tabular}           & 529K      & {\footnotesize\url {https://github.com/mk-minchul/dcface}}                                                         & freely shared for a research purpose                                                                                                                                                                                                                                                               & \multirow{-4}{*}{Unknown}             \\ \hline
\begin{tabular}[c]{@{}c@{}}Latent Diffusion\\~\cite{corvi2023detection}\end{tabular} & 20K       & {\footnotesize\url {https://github.com/grip-unina/DMimageDetection}}                                               & \begin{tabular}[c]{@{}c@{}}Copyright 2024 Image Processing Research Group of \\ University Federico II of Naples ('GRIP-UNINA').\\  All rights reserved.Licensed under the \\ Apache License, Version 2.0 (the "License")\end{tabular}                                                             & Apache-2.0                            \\ \hline
\begin{tabular}[c]{@{}c@{}}Palette\\~\cite{awsafur2023artifact}\end{tabular}          & 6K        & {\footnotesize\url {https://github.com/awsaf49/artifact/?tab=readme-ov-file\#data-generation}}                     & \begin{tabular}[c]{@{}c@{}}This dataset comes from ArtiFact dataset, \\ which dataset takes leverage of data from\\  multiple methods thus different parts\\  of the dataset come with different licenses.\end{tabular}                                                                            & MIT License                           \\ \hline
\begin{tabular}[c]{@{}c@{}}SD v1.5\\~\cite{song2023robustness}\end{tabular}          & 18K       &                                                                                                    &                                                                                                                                                                                                                                                                                                    &                                       \\ \cline{1-2}
\begin{tabular}[c]{@{}c@{}}SD Inpainting\\~\cite{song2023robustness}\end{tabular}    & 20.9K     & \multirow{-2}{*}{{\url {https://huggingface.co/datasets/OpenRL/DeepFakeFace}}}                        & \multirow{-2}{*}{freely shared for a research purpose}                                                                                                                                                                                                                                             & \multirow{-2}{*}{Apache-2.0}          \\ \hline
\begin{tabular}[c]{@{}c@{}}FFHQ\\~\cite{karras2019style}\end{tabular}              & 70K       & {\footnotesize\url {https://github.com/NVlabs/ffhq-dataset}}                                                       & \begin{tabular}[c]{@{}c@{}}You can use, redistribute, and adapt it for \\ non-commercial purposes, as long as you\\  (a) give appropriate credit by citing our paper, \\ (b) indicate any changes that you've made,and \\ (c) distribute any derivative works under the same license.\end{tabular} &\begin{tabular}[c]{@{}c@{}} Creative Commons \\ BY-NC-SA 4.0 license\end{tabular}  \\ \hline
\begin{tabular}[c]{@{}c@{}}IMDB-WIKI\\~\cite{rothe2015dex}\end{tabular}        & 239K      & {\url {https://data.vision.ee.ethz.ch/cvl/rrothe/imdb-wiki/}}                                         & \begin{tabular}[c]{@{}c@{}}This dataset is made available for academic \\ research purpose only. All the images are \\ collected from the Internet, and the \\ copyright belongs to the original owners.\end{tabular}                                                                              & Non Commericial                       \\ \hline
\end{tabular}
}
\caption{\small \textit{A list of datasets used in AI-Face, including the number of samples, links, access details, and licenses.}} 
\label{tab:dataset_info_detail}
\end{table*}

\subsection{Artifacts of Deepfake Forgeries in Frequency} \label{appendix:artifacts_frequency}
\begin{figure*}[t]
    \centering
    \includegraphics[width=1\textwidth]{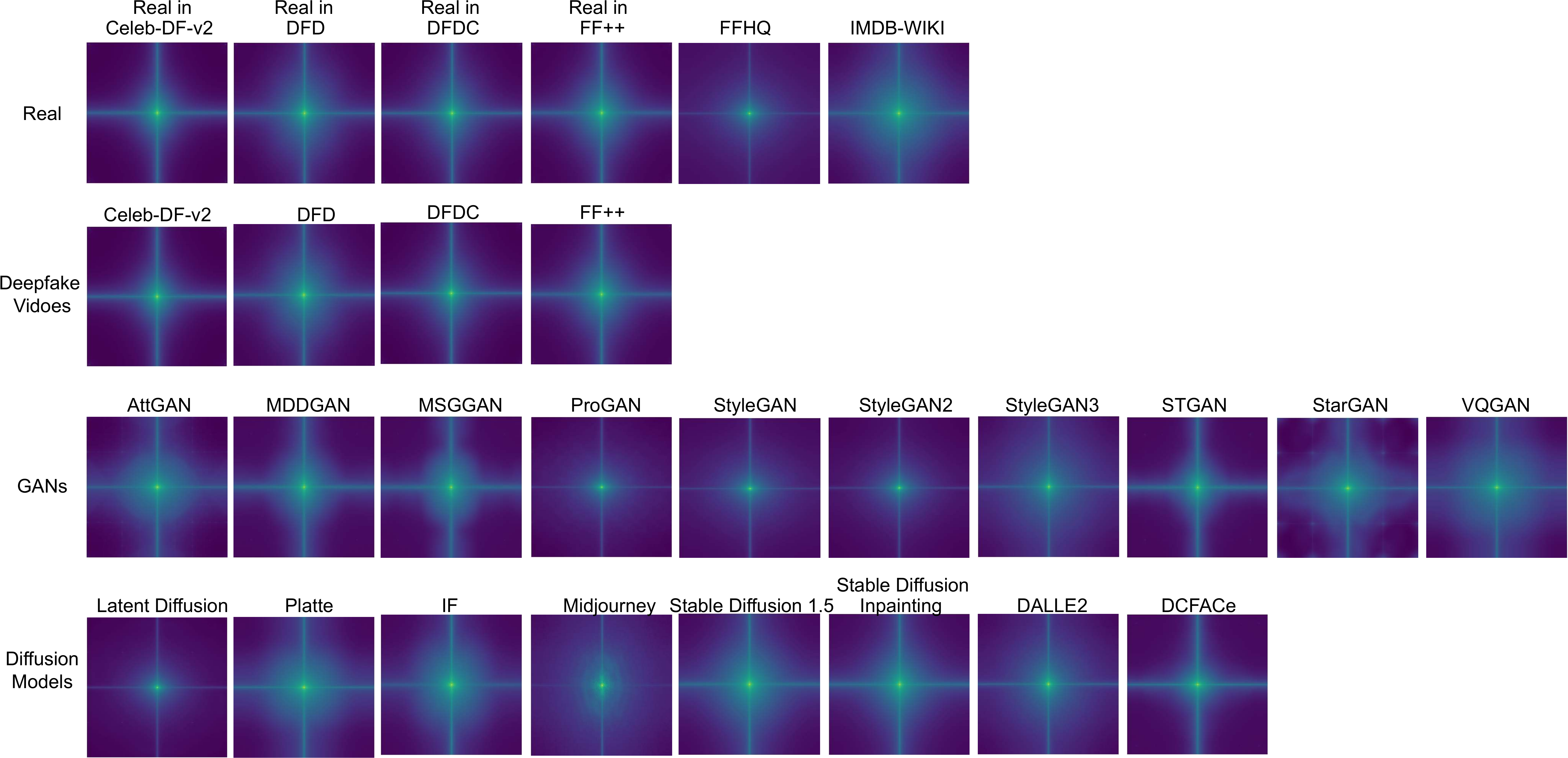}
    \vspace{-2mm}
    \caption{\textit{Frequency analysis on various sources. The mean FFT spectrum computation involves averaging over 2,000 images. DALLE2, IF, and Midjourney take average over 200, 500, and 100, respectively, due to their small number of samples.}}
    \label{fig:fft}
\end{figure*}

Leveraging frequency domain information plays a pivotal role in detecting AI-generated images. Frequency-based methods analyze the frequency components of an image, capturing information that may not be readily apparent in the spatial domain. In Fig.~\ref{fig:fft}, we present the mean Fast Fourier Transform (FFT) spectrum of images sampled from various sources in our AI-Face dataset. The results indicate that generative models often concentrate their output energy in the low-frequency range, represented by the central area of the FFT spectrum, resulting in overly smooth images. Notably, some models, such as StarGAN and Midjourney, exhibit distinct frequency artifacts, suggesting that they continue to struggle with eliminating generative patterns in the frequency domain. These artifacts serve as critical cues for distinguishing synthetic images from real ones. While most prior work has focused on applying frequency information to enhance the utility performance of detectors, exploring how frequency features can be leveraged to improve the fairness of detectors presents a promising direction for future research.

\subsection{Experimental Study of Existing Face Attribute Prediction Tools}\label{appendix:annotator_evaluation}
We compare current state-of-the-art face attribute prediction tools Face++~\cite{face++} and InsightFace~\cite{insightface} with our annotator.  
We perform  \textit{intra-domain} (train and test on IMDB-WIKI) and \textit{cross-domain} (train on IMDB-WIKI, test on four AI-generated face datasets) evaluations. FF++, DFDC, DFD, and Celeb-DF-v2 are selected for cross-domain evaluation because they contain AI-generated faces, which match our objective and are not used to train Face++ and InsightFace. Additionally, they have demographic attribute annotations from~\cite{xu2024analyzing}, which can be used as ground truth for annotator evaluation. Since those annotations provided by ~\cite{xu2024analyzing} have limited age annotations, our evaluation of these four datasets is confined to gender.  The intra-domain results are shown in Table~\ref{tab:annotator evaluation IMDB} and the results of cross-domain are in Table~\ref{tab:annotator evaluation}. Those results demonstrate our annotator's superiority in demographic attribute prediction and generalization capability against Face++ and InsightFace. For example, under intra-domain evaluation (Table~\ref{tab:annotator evaluation IMDB}), its precision surpasses the second-best method, InsightFace, by 3.47\% on Female and 24.81\% on Senior. In cross-domain evaluation (Table~\ref{tab:annotator evaluation IMDB}), our annotator maintains high accuracy on all datasets, reflecting good generalization. For instance, on the DFDC dataset, the precision our annotator outperforms Face++ by a margin of up to 1.07\% and InsightFace by 3.32\% on Female.

\begin{table*}[t]
\centering
\scalebox{0.48}{
\begin{tabular}{
>{\columncolor[HTML]{FFFFFF}}c |
>{\columncolor[HTML]{FFFFFF}}c 
>{\columncolor[HTML]{FFFFFF}}c 
>{\columncolor[HTML]{FFFFFF}}c |
>{\columncolor[HTML]{FFFFFF}}c 
>{\columncolor[HTML]{FFFFFF}}c 
>{\columncolor[HTML]{FFFFFF}}c |
>{\columncolor[HTML]{FFFFFF}}c 
>{\columncolor[HTML]{FFFFFF}}c 
>{\columncolor[HTML]{FFFFFF}}c |
>{\columncolor[HTML]{FFFFFF}}c 
>{\columncolor[HTML]{FFFFFF}}c 
>{\columncolor[HTML]{FFFFFF}}c |
>{\columncolor[HTML]{FFFFFF}}c 
>{\columncolor[HTML]{FFFFFF}}c 
>{\columncolor[HTML]{FFFFFF}}c |
>{\columncolor[HTML]{FFFFFF}}c 
>{\columncolor[HTML]{FFFFFF}}c 
>{\columncolor[HTML]{FFFFFF}}c |
>{\columncolor[HTML]{FFFFFF}}c 
>{\columncolor[HTML]{FFFFFF}}c 
>{\columncolor[HTML]{FFFFFF}}c }
\hline
\cellcolor[HTML]{FFFFFF}                              & \multicolumn{3}{c|}{\cellcolor[HTML]{FFFFFF}Female}     & \multicolumn{3}{c|}{\cellcolor[HTML]{FFFFFF}Male}                                       & \multicolumn{3}{c|}{\cellcolor[HTML]{FFFFFF}Child}        & \multicolumn{3}{c|}{\cellcolor[HTML]{FFFFFF}Young}        & \multicolumn{3}{c|}{\cellcolor[HTML]{FFFFFF}Adult}        & \multicolumn{3}{c|}{\cellcolor[HTML]{FFFFFF}Mid}          & \multicolumn{3}{c}{\cellcolor[HTML]{FFFFFF}Senior}        \\ \cline{2-22} 
\multirow{-2}{*}{\cellcolor[HTML]{FFFFFF}Method}      & Precision         & Recall            & F1              & Precision         & Recall            & \multicolumn{1}{l|}{\cellcolor[HTML]{FFFFFF}F1} & Precision         & Recall            & F1                & Precision         & Recall            & F1                & Precision         & Recall            & F1                & Precision         & Recall            & F1                & Precision         & Recall            & F1                \\ \hline
\cellcolor[HTML]{FFFFFF}                              & 0.9161            & 0.9142            & 0.9151          & 0.9143            & 0.9163            & 0.9153                                          & 0.8579            & 0.0220            & 0.0429            & 0.3942            & 0.2940            & 0.3368            & 0.3215            & 0.6700            & 0.4345            & 0.5423            & 0.6460            & 0.5894            & 0.8078            & 0.7700            & 0.7884            \\
\multirow{-2}{*}{\cellcolor[HTML]{FFFFFF}Face ++}     & (0.0064)          & (0.0031)          & 0.0045          & (0.0033)          & (0.0067)          & 0.0048                                          & (0.0962)          & (0.0062)          & (0.0119)          & (0.0222)          & (0.0161)          & (0.0186)          & (0.0054)          & (0.0192)          & (0.0087)          & (0.0160)          & (0.0210)          & (0.0147)          & (0.0143)          & (0.0165)          & (0.0127)          \\ \hline
\cellcolor[HTML]{FFFFFF}                              & 0.9648            & 0.9405            & 0.9525          & 0.9420            & 0.9656            & 0.9537                                          & \textbf{1.0000}   & 0.0020            & 0.0040            & 0.3970            & 0.0693            & 0.1180            & 0.2599            & 0.6180            & 0.3659            & 0.4105            & 0.5733            & 0.4783            & 0.7224            & 0.7560            & 0.7386            \\
\multirow{-2}{*}{\cellcolor[HTML]{FFFFFF}Insightface} & (0.0055)          & (0.0058)          & 0.0017          & (0.0050)          & (0.0057)          & 0.0017                                          & \textbf{(0.0000)} & (0.0016)          & (0.0032)          & (0.0337)          & (0.0098)          & (0.0155)          & (0.0027)          & (0.0124)          & (0.0048)          & (0.0067)          & (0.0196)          & (0.0094)          & (0.0124)          & (0.0238)          & (0.0143)          \\ \hline
\cellcolor[HTML]{FFFFFF}                              & \textbf{0.9995}   & \textbf{0.9992}   & \textbf{0.9993} & \textbf{0.9992}   & \textbf{0.9995}   & \textbf{0.9993}                                 & 0.9787            & \textbf{0.9780}   & \textbf{0.9783}   & \textbf{0.9560}   & \textbf{0.9393}   & \textbf{0.9476}   & \textbf{0.9265}   & \textbf{0.9547}   & \textbf{0.9402}   & \textbf{0.9498}   & \textbf{0.9320}   & \textbf{0.9408}   & \textbf{0.9642}   & \textbf{0.9700}   & \textbf{0.9671}   \\
\multirow{-2}{*}{\cellcolor[HTML]{FFFFFF}Ours}        & \textbf{(0.0006)} & \textbf{(0.0006)} & \textbf{0.0006} & \textbf{(0.0006)} & \textbf{(0.0006)} & \textbf{0.0006}                                 & (0.0027)          & \textbf{(0.0062)} & \textbf{(0.0039)} & \textbf{(0.0078)} & \textbf{(0.0080)} & \textbf{(0.0042)} & \textbf{(0.0122)} & \textbf{(0.0107)} & \textbf{(0.0058)} & \textbf{(0.0123)} & \textbf{(0.0123)} & \textbf{(0.0124)} & \textbf{(0.0085)} & \textbf{(0.0087)} & \textbf{(0.0082)} \\ \hline
\end{tabular}
}
\caption{\textit{Detailed comparison of our annotator against Face++~\cite{face++} and InsightFace~\cite{insightface} on IMDB-WIKI ~\cite{rothe2015dex} dataset. Prediction mean and standard deviation (in parentheses) of each method across 5 random samplings are reported. The best results are shown in Bold.}}
\label{tab:annotator evaluation IMDB}
\end{table*}

\begin{table}[t]
\centering
\scalebox{0.8}{
\begin{tabular}{
>{\columncolor[HTML]{FFFFFF}}c |
>{\columncolor[HTML]{FFFFFF}}c |
>{\columncolor[HTML]{FFFFFF}}c 
>{\columncolor[HTML]{FFFFFF}}c 
>{\columncolor[HTML]{FFFFFF}}c |
>{\columncolor[HTML]{FFFFFF}}c 
>{\columncolor[HTML]{FFFFFF}}c 
>{\columncolor[HTML]{FFFFFF}}c }
\hline
\cellcolor[HTML]{FFFFFF}                          & \cellcolor[HTML]{FFFFFF}                              & \multicolumn{3}{c|}{\cellcolor[HTML]{FFFFFF}Female}       & \multicolumn{3}{c}{\cellcolor[HTML]{FFFFFF}Male}          \\ \cline{3-8} 
\multirow{-2}{*}{\cellcolor[HTML]{FFFFFF}Dataset} & \multirow{-2}{*}{\cellcolor[HTML]{FFFFFF}Method}      & precision         & recall            & F1                & precision         & recall            & F1                \\ \hline
\cellcolor[HTML]{FFFFFF}                          & \cellcolor[HTML]{FFFFFF}                              & \textbf{0.9816}   & 0.9795            & 0.9805            & 0.9795            & \textbf{0.9816}   & 0.9805            \\
\cellcolor[HTML]{FFFFFF}                          & \multirow{-2}{*}{\cellcolor[HTML]{FFFFFF}Face ++}     & \textbf{(0.3021)} & (0.1360)          & (0.1459)          & (0.1312)          & \textbf{(0.3084)} & (0.1508)          \\ \cline{2-8} 
\cellcolor[HTML]{FFFFFF}                          & \cellcolor[HTML]{FFFFFF}                              & 0.9700            & 0.9664            & 0.9682            & 0.9666            & 0.9713            & 0.9683            \\
\cellcolor[HTML]{FFFFFF}                          & \multirow{-2}{*}{\cellcolor[HTML]{FFFFFF}Insightface} & (0.4697)          & (0.6046)          & (0.3867)          & (0.5794)          & (0.4815)          & (0.3802)          \\ \cline{2-8} 
\cellcolor[HTML]{FFFFFF}                          & \cellcolor[HTML]{FFFFFF}                              & 0.9799            & \textbf{0.9992}   & \textbf{0.9894}   & \textbf{0.9992}   & 0.9795            & \textbf{0.9892}   \\
\multirow{-6}{*}{\cellcolor[HTML]{FFFFFF}FF++}    & \multirow{-2}{*}{\cellcolor[HTML]{FFFFFF}Ours}        & (0.0022)          & \textbf{(0.0006)} & \textbf{(0.0009)} & \textbf{(0.0007)} & (0.0023)          & \textbf{(0.0009)} \\ \hline
\cellcolor[HTML]{FFFFFF}                          & \cellcolor[HTML]{FFFFFF}                              & 0.9412            & 0.8992            & 0.9197            & 0.9035            & 0.9437            & \textbf{0.9231}   \\
\cellcolor[HTML]{FFFFFF}                          & \multirow{-2}{*}{\cellcolor[HTML]{FFFFFF}Face ++}     & (0.9771)          & (1.0095)          & (0.5639)          & (0.8353)          & (1.0246)          & \textbf{(0.5353)} \\ \cline{2-8} 
\cellcolor[HTML]{FFFFFF}                          & \cellcolor[HTML]{FFFFFF}                              & 0.9187            & 0.7869            & 0.8475            & 0.8139            & 0.9301            & 0.8680            \\
\cellcolor[HTML]{FFFFFF}                          & \multirow{-2}{*}{\cellcolor[HTML]{FFFFFF}Insightface} & (0.9855)          & (1.7976)          & (0.7444)          & (1.1401)          & (1.0587)          & (0.3842)          \\ \cline{2-8} 
\cellcolor[HTML]{FFFFFF}                          & \cellcolor[HTML]{FFFFFF}                              & \textbf{0.9519}   & \textbf{0.9741}   & \textbf{0.9629}   & \textbf{0.9735}   & \textbf{0.9507}   & 0.0114            \\
\multirow{-6}{*}{\cellcolor[HTML]{FFFFFF}DFDC}    & \multirow{-2}{*}{\cellcolor[HTML]{FFFFFF}Ours}        & \textbf{(0.0106)} & \textbf{(0.0014)} & \textbf{(0.0059)} & \textbf{(0.0015)} & \textbf{(0.9619)} & (0.0064)          \\ \hline
\cellcolor[HTML]{FFFFFF}                          & \cellcolor[HTML]{FFFFFF}                              & \textbf{0.9501}   & 0.8228            & 0.8818            & 0.8440            & \textbf{0.9568}   & 0.8968            \\
\cellcolor[HTML]{FFFFFF}                          & \multirow{-2}{*}{\cellcolor[HTML]{FFFFFF}Face ++}     & \textbf{(0.3773)} & (1.8758)          & (1.0113)          & (1.3784)          & \textbf{(0.3907)} & (0.6856)          \\ \cline{2-8} 
\cellcolor[HTML]{FFFFFF}                          & \cellcolor[HTML]{FFFFFF}                              & 0.9441            & 0.7557            & 0.8394            & 0.7964            & 0.9552            & 0.8686            \\
\cellcolor[HTML]{FFFFFF}                          & \multirow{-2}{*}{\cellcolor[HTML]{FFFFFF}Insightface} & (0.7765)          & (0.9555)          & (0.7937)          & (0.6834)          & (0.6344)          & (0.5961)          \\ \cline{2-8} 
\cellcolor[HTML]{FFFFFF}                          & \cellcolor[HTML]{FFFFFF}                              & 0.9378            & \textbf{0.9365}   & \textbf{0.9366}   & \textbf{0.9366}   & 0.9379            & \textbf{0.9372}   \\
\multirow{-6}{*}{\cellcolor[HTML]{FFFFFF}DFD}     & \multirow{-2}{*}{\cellcolor[HTML]{FFFFFF}Ours}        & (0.0045)          & \textbf{(0.0053)} & \textbf{(0.0033)} & \textbf{(0.0049)} & (0.0048)          & \textbf{(0.0033)} \\ \hline
\cellcolor[HTML]{FFFFFF}                          & \cellcolor[HTML]{FFFFFF}                              & 0.9989            & 0.9648            & 0.9815            & 0.9660            & 0.9989            & 0.9822            \\
\cellcolor[HTML]{FFFFFF}                          & \multirow{-2}{*}{\cellcolor[HTML]{FFFFFF}Face ++}     & (0.0553)          & (0.4182)          & (0.2361)          & (0.3918)          & (0.0533)          & (0.2215)          \\ \cline{2-8} 
\cellcolor[HTML]{FFFFFF}                          & \cellcolor[HTML]{FFFFFF}                              & 0.9984            & 0.9811            & 0.9896            & 0.9814            & 0.9984            & 0.9898            \\
\cellcolor[HTML]{FFFFFF}                          & \multirow{-2}{*}{\cellcolor[HTML]{FFFFFF}Insightface} & (0.0541)          & (0.3518)          & (0.1801)          & (0.3396)          & (0.0534)          & (0.1737)          \\ \cline{2-8} 
\cellcolor[HTML]{FFFFFF}                          & \cellcolor[HTML]{FFFFFF}                              & \textbf{1.0000}   & \textbf{0.9997}   & \textbf{0.9999}   & \textbf{0.9997}   & \textbf{1.0000}   & \textbf{0.9999}   \\
\multirow{-6}{*}{\cellcolor[HTML]{FFFFFF}Celeb-DF-v2} & \multirow{-2}{*}{\cellcolor[HTML]{FFFFFF}Ours}        & \textbf{(0.0000)} & \textbf{(0.0005)} & \textbf{(0.0003)} & \textbf{(0.0005)} & \textbf{(0.0000)} & \textbf{(0.0003)} \\ \hline
\end{tabular}
}
\caption{\small \textit{Detailed comparison of our annotator against Face++~\cite{face++} and InsightFace~\cite{insightface} on FF++~\cite{rossler2019faceforensics++}, DFDC~\cite{dolhansky2020deepfake}, DFD~\cite{google2019deepfake}, and Celeb-DF-v2~\cite{li2020celeb} datasets. Prediction mean and standard deviation (in parentheses) of each method across 5 random samplings. The best results are shown in Bold.}}

\label{tab:annotator evaluation}
\end{table}

\subsection{Annotator Implementation Detail}\label{appendix:annotator_develop}
Our annotators are implemented by PyTorch and trained with a single NVIDIA RTX A6000 GPU. For training, we fix the batch size 64, epochs 32, and use Adam optimizer with an initial learning rate $\beta=1e-3$. Additionally, we employ a Cosine Annealing Learning Rate Scheduler to modulate the learning rate adaptively across the training duration.
In terms of the imbalance loss, $u_{A_i}$ is the weighting factor for attribute $A_i$. $h(f_i)_{A_i}$ is the predict logit on $A_i$. $\zeta_{A_i}$ is the multiplicative logit scaling factor, $\zeta_{A_i} = \left( \frac{N_{A_i}}{N_{\text{max}}} \right)^{\kappa}$, $N_{\text{max}}$ is the number of samples in the most frequent class, $\kappa$ is the hyperparameter controlling the sensitivity of scaling, it is set as 0.2 here. $\Delta_{A_i}$ is the additive logit scaling factor, calculated as the log of $A_i$ probabilities $\Delta_{A_i}=\rho \cdot \log \left( \frac{N_{A_i}}{N_{\text{total}}} \right)$. The regularization hyperparameter $\alpha$ in fairness loss is $1\text{e-}4$. The hyperparameter $\gamma$ in SAM optimization is set as 0.05. 


\begin{figure*}[t]
    \centering
    \includegraphics[width=1\textwidth]{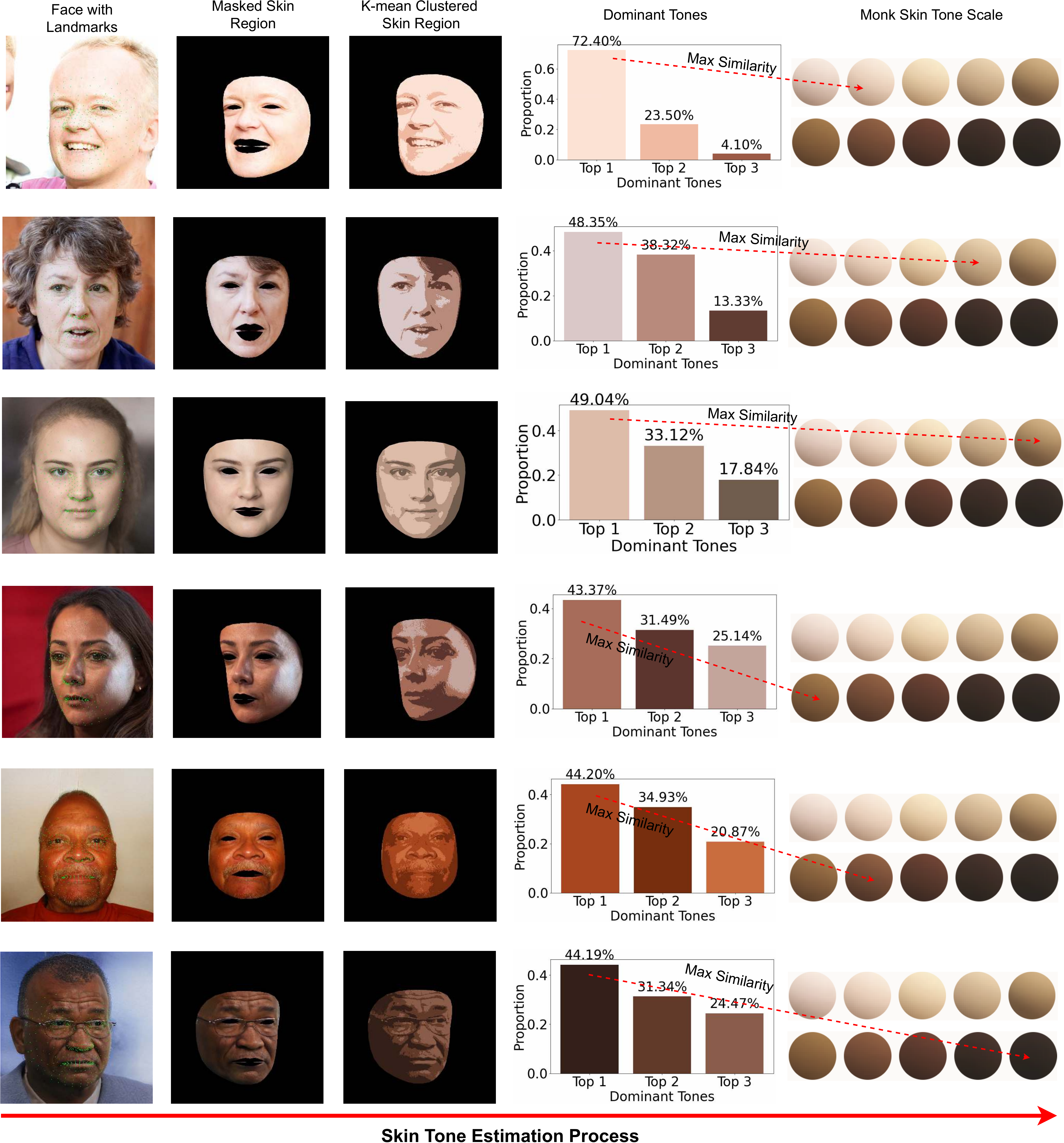}
    \vspace{-4mm}
    \caption{\textit{Visualization of the skin tone estimation process.}}
    \label{fig:skintone_process}
\end{figure*}


\subsection{Details of Human Labeling Activities in Annotation Quality Assessment}\label{appendix:human_label}
The annotation process for assessing the quality of AI-generated face image annotations followed a structured and ethically grounded methodology. Prior to labeling, all human annotators signed an Annotator Agreement outlining the project objectives, confidentiality requirements, and detailed labeling guidelines for gender and age classification. This agreement emphasized impartiality, respect, and adherence to professional conduct throughout the annotation activities. Human annotators then underwent tutorial training using real example images to familiarize themselves with demographic attributes and labeling criteria, focusing on identifying gender-specific features, such as facial structure and presence of facial hair, and age-related indicators, including wrinkles and skin elasticity.

\smallskip
\noindent
Following the agreement and training, human annotators independently labeled the images based on the established criteria, categorizing gender and age into predefined classes, and recorded their annotations in a CSV file. A structured conflict resolution approach ensured accuracy and consistency in annotations. Labels agreed upon by a majority of annotators were finalized directly, while unanimous disagreements were resolved through collaborative discussions guided by the annotation guidelines. This process ensured that all annotations were objective, reliable, and aligned with ethical standards set forth in the signed agreement.

\subsection{Visualization of Skin Tone Annotation Generation}\label{appendix:vis_skintone}
The visualization shown in Fig.~\ref{fig:skintone_process} illustrates the skin tone estimation process using the Monk Skin Tone (MST) Scale. Each row represents a sample image, showing the progression from the original face with facial landmarks to the masked skin region that excludes non-skin areas like eyes and lips. Subsequently, the K-Means clustered skin region highlights the dominant skin tones extracted from the facial area. On the right, bar plots display the proportions of the top three dominant tones within the clustered region, with the top tone (largest cluster) mapped to the closest MST Scale shade. This mapping is achieved by calculating the maximum similarity, as indicated by the Euclidean distance in RGB space between the cluster centroid and MST reference colors. This process visually demonstrates how the methodology isolates, clusters, and estimates skin tones for accurate skin tone annotation generation.

\section{Fairness Benchmark Settings}
\subsection{Implementation Detail}\label{appendix:fairness_implementation_details}
For fairness benchmark, all experiments are based on the PyTorch with a single NVIDIA RTX A6000 GPU. During training, we utilize SGD optimizer with a learning rate of 0.0005, with momentum of 0.9 and weight decay of 0.005. The batch size is set to 128 for most detectors. However, for the SRM~\cite{luo2021generalizing}, UCF~\cite{yan2023ucf}, and PG-FDD~\cite{lin2024preserving}, the batch size is adjusted to 32 due to GPU memory. For hyperparameters defined in these detectors, we use the default values set in their original papers. All detectors are initialized with their official pre-trained weights, and trained for 10 epochs.

\subsection{Details of Detection Methods}\label{appendix:detection_methods}
We summarized the backbone architecture, GitHub repository link, and publication venue of the detectors implemented in our fairness benchmark in Table~\ref{tab:implemented_detectors}. A brief introduction to each detector is provided below:
\begin{table*}[t]
\centering
\scalebox{0.61}{
\begin{tabular}{c|c|c|c|c}
\hline
Model Type                         & Detector     & Backbone     & GitHub Link                                                                                   & VENUE      \\ \hline
\multirow{3}{*}{Naive}             & Xception~\cite{chollet2017xception}     & Xception     & {\url {https://github.com/ondyari/FaceForensics/blob/master}} & ICCV-2019  \\
                                   & Efficient-B4~\cite{tan2019efficientnet} & EfficientNet & {\url {https://github.com/lukemelas/EfficientNet-PyTorch}}   & ICML-2019  \\
                                   & ViT-B/16~\cite{dosovitskiy2020image}     & Transformer  & {\url {https://github.com/lucidrains/vit-pytorch}}                                               & ICLR-2021  \\ \hline
\multirow{3}{*}{Spatial}           & UCF~\cite{yan2023ucf}          & Xception     & {\url {https://github.com/SCLBD/DeepfakeBench/tree/main}}                                        & ICCV-2023  \\
                                   & UnivFD~\cite{ojha2023towards}       & CLIP VIT     & {\url {https://github.com/Yuheng-Li/UniversalFakeDetect}}                                        & CVPR-2023  \\
                                   & CORE~\cite{ni2022core}         & Xception     & {\url {https://github.com/niyunsheng/CORE}}                                                      & CVPRW-2022 \\ \hline
\multirow{3}{*}{Frequency}         & F3Net~\cite{qian2020thinking}        & Xception     & {\url {https://github.com/yyk-wew/F3Net}}                                                        & ECCV-2020  \\
                                   & SRM~\cite{luo2021generalizing}          & Xception     & {\url {https://github.com/SCLBD/DeepfakeBench/tree/main}}                                        & CVPR-2021  \\
                                   & SPSL~\cite{liu2021spatial}         & Xception     & {\url {https://github.com/SCLBD/DeepfakeBench/tree/main}}                                        & CVPR-2021  \\ \hline
\multirow{3}{*}{\begin{tabular}[c]{@{}c@{}}Fairness-\\ enhanced\end{tabular}} & DAW-FDD~\cite{ju2024improving}      & Xception     & Unpublished code, reproduced by us                                                            & WACV-2024  \\
                                   & DAG-FDD~\cite{ju2024improving}      & Xception     & Unpublished code, reproduced by us                                                            & WACV-2024  \\
                                   & PG-FDD~\cite{lin2024preserving}       & Xception     & {\url {https://github.com/Purdue-M2/Fairness-Generalization}}                                    & CVPR-2024  \\ \hline
\end{tabular}
}
\vspace{-2mm}
\caption{\textit{Summary of the implemented detectors in our fairness benchmark.}}
\label{tab:implemented_detectors}
\end{table*}

\smallskip
\noindent
\textbf{Xception}~\cite{chollet2017xception}: is a deep convolutional neural network (CNN) architecture that relies on depthwise separable convolutions. This approach significantly reduces the number of parameters and computational cost while maintaining high performance. Xception serves as a classic backbone in deepfake detectors.

\noindent
\textbf{EfficientB4}~\cite{tan2019efficientnet}: is part of the EfficientNet family~\cite{tan2019efficientnet}, which utilizes a novel model scaling method that uniformly scales all dimensions of depth, width, and resolution using a compound coefficient. EfficientNet also serves as a classic backbone in deepfake detectors.

\noindent
\textbf{ViT-B/16}~\cite{dosovitskiy2020image}: is a model that applies the transformer architecture, the 'B' denotes the base model size, and '16' indicates the patch size. ViT-B/16 splits images into 16 patches, linearly embeds each patch, adds positional embeddings, and feeds the resulting sequence of vectors into a standard transformer encoder. 

\noindent
\textbf{F3Net}~\cite{qian2020thinking}: utilizes a cross-attention two-stream network to effectively identify frequency-aware clues by integrating two branches: FAD and LFS. The FAD (Frequency-aware Decomposition) module divides the input image into various frequency bands using learnable partitions, representing the image with frequency-aware components to detect forgery patterns through this decomposition. Meanwhile, the LFS (Localized Frequency Statistics) module captures local frequency statistics to highlight statistical differences between authentic and counterfeit faces. 

\noindent
\textbf{SPSL}~\cite{liu2021spatial}: integrates spatial image data with the phase spectrum to detect up-sampling artifacts in face forgeries, enhancing the model's generalization ability for face forgery detection. The paper provides a theoretical analysis of the effectiveness of using the phase spectrum. Additionally, it highlights that local texture information is more important than high-level semantic information for accurately detecting face forgeries.

\noindent
\textbf{SRM}~\cite{luo2021generalizing}: extracts high-frequency noise features and combines two different representations from the RGB and frequency domains to enhance the model's generalization ability for face forgery detection.

\noindent
\textbf{UCF}~\cite{yan2023ucf}: presents a multi-task disentanglement framework designed to tackle two key challenges in deepfake detection: overfitting to irrelevant features and overfitting to method-specific textures. By identifying and leveraging common features, this framework aims to improve the model's generalization ability.

\noindent
\textbf{UnivFD}~\cite{ojha2023towards}: uses the frozen CLIP ViT-L/14~\cite{openclip2021} as feature extractor and trains the last linear layer to classify fake and real images.

\noindent
\textbf{CORE}~\cite{ni2022core}: explicitly enforces the consistency of different representations. It first captures various representations through different augmentations and then regularizes the cosine distance between these representations to enhance their consistency.

\noindent
\textbf{DAW-FDD}~\cite{ju2024improving}: a demographic-aware Fair Deepfake Detection (DAW-FDD) method leverages demographic information and employs an existing fairness risk measure~\cite{williamson2019fairness}. At a high level,
DAW-FDD aims to ensure that the losses achieved by different user-specified groups of interest (\eg, different races or genders) are similar to each other (so that the AI face detector is not more accurate on one group vs another) and, moreover, that the losses across all groups are low.  Specifically, DAW-FDD uses a CVaR~\cite{levy2020large, rockafellar2000optimization} loss function across groups (to address imbalance in demographic groups) and, per group, DAW-FDD uses another CVaR loss function (to address imbalance in real vs AI-generated training examples).

\noindent
\textbf{DAG-FDD}~\cite{ju2024improving}: a demographic-agnostic Fair Deepfake Detection (DAG-FDD) method, which is based on the distributionally robust optimization (DRO)~\cite{hashimoto2018fairness, duchi2021learning}. To use DAG-FDD, the user does not have to specify which attributes to treat as sensitive such as race and gender, only need to specify a probability threshold for a minority group without explicitly identifying all possible groups.

\noindent
\textbf{PG-FDD}~\cite{lin2024preserving}: PG-FDD (Preserving Generalization Fair Deepfake Detection) employs disentanglement learning to extract demographic and domain-agnostic forgery features, promoting fair learning across a flattened loss landscape. Its framework combines disentanglement learning, fairness learning, and optimization modules. The disentanglement module introduces a loss to expose demographic and domain-agnostic features that enhance fairness generalization. The fairness learning module combines these features to promote fair learning, guided by generalization principles. The optimization module flattens the loss landscape, helping the model escape suboptimal solutions and strengthen fairness generalization.

\subsection{Fairness Metrics} \label{appendix:fairness_metrics}
We assume a test set comprising indices \{1, \ldots, $n$\}. $Y_j$ and $\hat{Y}_j$ respectively represent the true and predicted labels of the sample $X_j$. Their values are binary, where 0 means real and 1 means fake. For all fairness metrics, a lower value means better performance. The formulations of fairness metrics are as follows,
\begin{align*}
    &F_{EO} := \sum_{\mathcal{J}_j\in\mathcal{J}} \sum_{q=0}^{1} \left| \frac{\sum_{j=1}^{n} \mathbb{I}_{\left[\hat{Y}_j=1, D_j=\mathcal{J}_j, Y_j=q \right]}}{\sum_{j=1}^{n} \mathbb{I}_{\left[ D_j=\mathcal{J}_j, Y_j=q\right]}} - \frac{\sum_{j=1}^{n} \mathbb{I}_{\left[\hat{Y}_j=1, Y_j=q \right]}}{\sum_{j=1}^{n} \mathbb{I}_{\left[Y_j=q\right]}} \right|,\\
    &F_{O\!A\!E} := \max_{\mathcal{J}_j\in\mathcal{J}} \left\{ \frac{\sum_{j=1}^{n} \mathbb{I}_{[\hat{Y}_j = Y_j, D_j = \mathcal{J}_j]}}{\sum_{j=1}^{n} \mathbb{I}_{[D_j = \mathcal{J}_j]}} \right. \quad \left. - \min_{{\mathcal{J}_j}' \in \mathcal{J}} \frac{\sum_{j=1}^{n} \mathbb{I}_{[\hat{Y}_j = Y_j, D_j = {\mathcal{J}_j}']}}{\sum_{j=1}^{n} \mathbb{I}_{[D_j = {\mathcal{J}_j}']}} \right\},\\
    &F_{DP} := \max_{q \in \{0,1\}} \left\{ \max_{J_j \in \mathcal{J}} \frac{\sum_{j=1}^{n} \mathbb{I}_{[\hat{Y}_j=q, D_j=J_j]}}{\sum_{j=1}^{n} \mathbb{I}_{[D_j=J_j]}} \right.\quad \left. - \min_{J_j' \in \mathcal{J}} \frac{\sum_{j=1}^{n} \mathbb{I}_{[\hat{Y}_j=q, D_j=J_j']}}{\sum_{j=1}^{n} \mathbb{I}_{[D_j=J_j']}} \right\}, \\
    &F_{M\!E\!O} := \max_{q, q' \in \{0,1\}} \left\{ \max_{J_j \in \mathcal{J}} \frac{\sum_{j=1}^{n} \mathbb{I}_{[\hat{Y}_j=q, Y_j=q', D_j=J_j]}}{\sum_{j=1}^{n} \mathbb{I}_{[D_j=J_j, Y_j=q]}} \right. \quad \left. - \min_{J_j' \in \mathcal{J}} \frac{\sum_{j=1}^{n} \mathbb{I}_{[\hat{Y}_j=q, Y_j=q', D_j=J_j']}}{\sum_{j=1}^{n} \mathbb{I}_{[D_j=J_j', Y_j=q]}} \right\}, \\
    &F_{IND} := \sum_{j=1}^{n-1} \sum_{l=j+1}^{n} [\left| f(X_j) - f(X_l) \right| - \delta \|X_j - X_l\|_2]_+,
\end{align*}
where $D$ is the demographic variable, $\mathcal{J}$ is the set of subgroups with each subgroup $\mathcal{J}_j\in\mathcal{J}$.
$M$ is the set of detection models and $F$ is the set of fairness metrics.
$F_{EO}$ measures the disparity in TPR and FPR between each subgroup and the overall population. $F_{OAE}$ measures the maximum ACC gap across all demographic groups. $F_{DP}$ measures the maximum difference in prediction rates across all demographic groups. And $F_{MEO}$ captures the largest disparity in prediction outcomes (either positive or negative) when comparing different demographic groups. $\delta$ in $F_{IND}$ is a predefined scale factor (0.08 in our experiments). $[\cdot]_+$ is the hinge function, $\|\cdot\|_2$ is the $\ell_2$ norm. $f(X_j)$ represents the predicted logits of the model for input sample $X_j$. $F_{IND}$ points that a model should be fair across individuals if similar individuals have similar predicted outcomes.

\section{More Fairness Benchmark Results and Analysis}\label{appendix:more_results}

\subsection{Detailed Results of Overall Performance Comparison}
Detailed test results of each subgroup of each detector on AI-Face are presented in this section. Table~\ref{tab:detailed_overall_results} provides comprehensive metrics of each subgroup on AI-Face. These
results and findings align with the results reported in Table.~\ref{tab:intra_domain_overall} submitted main manuscript.

\begin{table*}[t]
\centering
\scalebox{0.675}{
\begin{tabular}{c|c|c|cc|ccc|ccccc|cccccc}
\hline
\multirow{2}{*}{\textbf{Model Type}}                                                    & \multirow{2}{*}{\textbf{Method}}                                               & \multirow{2}{*}{\textbf{Metric}} & \multicolumn{2}{c|}{\textbf{Gender}} & \multicolumn{3}{c|}{\textbf{Skin Tone}} & \multicolumn{5}{c|}{\textbf{Age}}                                                    & \multicolumn{6}{c}{\textbf{Intersection}}                                               \\ \cline{4-19} 
                                                                                        &                                                                                &                                  & \textbf{M}        & \textbf{F}       & \textbf{L}  & \textbf{M}  & \textbf{D} & \textbf{Child} & \textbf{Young} & \textbf{Adult} & \textbf{Middle} & \textbf{Senior} & \textbf{M-L} & \textbf{M-M} & \textbf{M-D} & \textbf{F-L} & \textbf{F-M} & \textbf{F-D} \\ \hline
\multirow{12}{*}{\textbf{Naive}}                                                        & \multirow{4}{*}{\begin{tabular}[c]{@{}c@{}}Xception\\ ~\cite{chollet2017xception}\end{tabular}}     & AUC                              & 98.90             & 98.20            & 97.69       & 98.44       & 98.88      & 95.95          & 97.86          & 99.10          & 98.66           & 96.54           & 97.88        & 98.70        & 99.22        & 97.53        & 98.17        & 98.19        \\
                                                                                        &                                                                                & FPR                              & 11.12             & 15.10            & 18.37       & 14.72       & 9.54       & 36.83          & 18.26          & 8.95           & 12.78           & 20.42           & 17.70        & 12.82        & 8.42         & 18.93        & 16.58        & 11.32        \\
                                                                                        &                                                                                & TPR                              & 99.38             & 99.22            & 98.94       & 99.62       & 98.43      & 99.55          & 99.38          & 99.21          & 99.49           & 98.51           & 98.55        & 99.64        & 98.87        & 99.16        & 99.59        & 97.58        \\
                                                                                        &                                                                                & ACC                              & 96.77             & 95.80            & 95.16       & 96.43       & 96.03      & 89.83          & 94.98          & 97.10          & 97.06           & 92.06           & 94.26        & 96.79        & 96.78        & 95.70        & 96.10        & 94.67        \\ \cline{2-19} 
                                                                                        & \multirow{4}{*}{\begin{tabular}[c]{@{}c@{}}EfficientB4\\ ~\cite{tan2019efficientnet}\end{tabular}} & AUC                              & 98.86             & 98.31            & 99.23       & 98.94       & 97.59      & 99.63          & 98.61          & 98.44          & 98.82           & 98.69           & 99.01        & 99.10        & 98.36        & 99.39        & 98.78        & 95.94        \\
                                                                                        &                                                                                & FPR                              & 17.57             & 22.96            & 21.32       & 17.17       & 25.47      & 16.13          & 22.92          & 19.79          & 19.41           & 19.07           & 20.50        & 15.44        & 20.74        & 22.00        & 18.87        & 33.02        \\
                                                                                        &                                                                                & TPR                              & 99.04             & 98.56            & 99.49       & 99.02       & 98.19      & 99.56          & 98.89          & 98.55          & 99.14           & 98.66           & 99.22        & 99.25        & 98.60        & 99.64        & 98.81        & 97.40        \\
                                                                                        &                                                                                & ACC                              & 94.91             & 93.42            & 94.95       & 95.43       & 91.05      & 95.37          & 93.45          & 93.81          & 95.48           & 92.62           & 94.01        & 95.90        & 93.04        & 95.51        & 94.99        & 87.44        \\ \cline{2-19} 
                                                                                        & \multirow{4}{*}{\begin{tabular}[c]{@{}c@{}}ViT-B/16\\ ~\cite{dosovitskiy2020image}\end{tabular}}          & AUC                              & 99.02             & 98.26            & 97.50       & 98.77       & 98.49      & 95.55          & 98.23          & 98.98          & 98.80           & 97.28           & 97.21        & 99.03        & 99.03        & 97.80        & 98.50        & 97.21        \\
                                                                                        &                                                                                & FPR                              & 14.06             & 19.17            & 21.74       & 16.86       & 15.48      & 28.28          & 21.09          & 13.34          & 17.22           & 20.92           & 21.43        & 14.93        & 12.61        & 22.00        & 18.76        & 20.05        \\
                                                                                        &                                                                                & TPR                              & 98.43             & 97.51            & 96.75       & 98.23       & 97.36      & 94.58          & 97.88          & 98.14          & 98.40           & 96.16           & 96.77        & 98.56        & 98.21        & 96.74        & 97.93        & 95.72        \\
                                                                                        &                                                                                & ACC                              & 95.33             & 93.52            & 92.72       & 94.88       & 93.49      & 88.48          & 93.16          & 95.17          & 95.31           & 90.34           & 91.96        & 95.48        & 95.10        & 93.16        & 94.33        & 90.55        \\ \hline
\multirow{12}{*}{\textbf{Frequency}}                                                    & \multirow{4}{*}{\begin{tabular}[c]{@{}c@{}}F3Net\\ ~\cite{qian2020thinking}\end{tabular}}        & AUC                              & 99.09             & 98.24            & 98.51       & 98.68       & 98.79      & 96.05          & 98.17          & 99.15          & 98.97           & 97.40           & 98.66        & 99.00        & 99.27        & 98.30        & 98.36        & 97.80        \\
                                                                                        &                                                                                & FPR                              & 12.49             & 17.21            & 30.72       & 16.54       & 10.78      & 49.32          & 21.21          & 10.51          & 12.52           & 20.00           & 30.75        & 14.33        & 9.38         & 30.69        & 18.72        & 13.02        \\
                                                                                        &                                                                                & TPR                              & 99.15             & 99.00            & 99.69       & 99.41       & 98.11      & 99.77          & 99.40          & 99.00          & 99.06           & 97.96           & 99.55        & 99.34        & 98.76        & 99.76        & 99.48        & 96.86        \\
                                                                                        &                                                                                & ACC                              & 96.25             & 95.12            & 93.05       & 95.87       & 95.43      & 86.66          & 94.27          & 96.54          & 96.77           & 91.84           & 91.55        & 96.22        & 96.42        & 93.95        & 95.55        & 93.63        \\ \cline{2-19} 
                                                                                        & \multirow{4}{*}{\begin{tabular}[c]{@{}c@{}}SPSL\\ ~\cite{liu2021spatial}\end{tabular}}         & AUC                              & 98.88             & 98.58            & 98.99       & 98.70       & 98.81      & 97.41          & 98.05          & 99.17          & 98.75           & 97.32           & 99.05        & 98.78        & 99.02        & 98.92        & 98.61        & 98.36        \\
                                                                                        &                                                                                & FPR                              & 11.62             & 16.03            & 19.50       & 14.77       & 11.44      & 37.20          & 20.20          & 9.58           & 13.12           & 19.07           & 18.94        & 12.88        & 9.57         & 19.95        & 16.62        & 14.43        \\
                                                                                        &                                                                                & TPR                              & 99.64             & 99.52            & 99.53       & 99.73       & 99.15      & 99.78          & 99.58          & 99.53          & 99.69           & 99.12           & 99.44        & 99.77        & 99.37        & 99.58        & 99.70        & 98.73        \\
                                                                                        &                                                                                & ACC                              & 96.84             & 95.80            & 95.38       & 96.51       & 95.96      & 89.90          & 94.65          & 97.17          & 97.16           & 92.92           & 94.59        & 96.88        & 96.80        & 95.85        & 96.17        & 94.42        \\ \cline{2-19} 
                                                                                        & \multirow{4}{*}{\begin{tabular}[c]{@{}c@{}}SRM\\ ~\cite{luo2021generalizing}\end{tabular}}          & AUC                              & 98.45             & 97.40            & 98.71       & 97.94       & 97.95      & 97.24          & 97.27          & 98.42          & 98.13           & 97.78           & 99.24        & 98.46        & 98.52        & 98.36        & 97.48        & 96.82        \\
                                                                                        &                                                                                & FPR                              & 12.84             & 19.11            & 22.30       & 17.50       & 12.30      & 36.76          & 22.42          & 11.92          & 15.33           & 19.64           & 18.94        & 14.70        & 9.92         & 25.06        & 20.25        & 16.10        \\
                                                                                        &                                                                                & TPR                              & 98.84             & 98.33            & 99.41       & 99.02       & 97.35      & 99.23          & 98.91          & 98.29          & 98.97           & 97.52           & 99.89        & 99.20        & 98.06        & 99.16        & 98.85        & 95.99        \\
                                                                                        &                                                                                & ACC                              & 95.93             & 94.16            & 94.67       & 95.35       & 94.44      & 89.62          & 93.59          & 95.65          & 96.14           & 91.67           & 94.91        & 96.03        & 95.76        & 94.53        & 94.72        & 92.03        \\ \hline
\multirow{12}{*}{\textbf{Spatial}}                                                      & \multirow{4}{*}{\begin{tabular}[c]{@{}c@{}}UCF\\~\cite{yan2023ucf}\end{tabular}}          & AUC                              & 98.62             & 97.45            & 97.20       & 97.92       & 98.49      & 95.59          & 97.26          & 98.74          & 98.67           & 97.04           & 97.67        & 98.44        & 99.00        & 96.88        & 97.41        & 97.47        \\
                                                                                        &                                                                                & FPR                              & 11.30             & 16.37            & 26.23       & 16.01       & 8.90       & 55.93          & 21.10          & 8.43           & 11.70           & 20.40           & 24.85        & 13.82        & 7.23         & 27.37        & 18.16        & 11.57        \\
                                                                                        &                                                                                & TPR                              & 98.20             & 97.77            & 98.75       & 98.37       & 96.89      & 99.53          & 98.50          & 97.58          & 98.35           & 96.91           & 99.00        & 98.47        & 97.63        & 98.61        & 98.28        & 95.45        \\
                                                                                        &                                                                                & ACC                              & 95.84             & 94.39            & 93.30       & 95.18       & 95.14      & 84.71          & 93.61          & 96.03          & 96.36           & 91.01           & 92.70        & 95.66        & 96.24        & 93.65        & 94.73        & 93.15        \\ \cline{3-18}
                                                                                        & \multirow{4}{*}{\begin{tabular}[c]{@{}c@{}}UnivFD\\ ~\cite{ojha2023towards}\end{tabular}}       & AUC                              & 98.55             & 97.76            & 98.38       & 98.47       & 97.40      & 98.13          & 98.07          & 98.33          & 98.11           & 96.94           & 98.24        & 98.76        & 98.13        & 98.46        & 98.19        & 95.84        \\
                                                                                        &                                                                                & FPR                              & 16.46             & 20.97            & 20.06       & 19.04       & 17.60      & 19.90          & 19.91          & 16.70          & 22.13           & 17.04           & 17.70        & 16.84        & 15.88        & 22.00        & 21.20        & 20.36        \\
                                                                                        &                                                                                & TPR                              & 98.02             & 97.12            & 97.57       & 98.26       & 95.69      & 97.43          & 97.40          & 97.39          & 98.41           & 94.49           & 97.10        & 98.54        & 96.99        & 97.83        & 98.02        & 93.17        \\
                                                                                        &                                                                                & ACC                              & 94.42             & 92.80            & 93.73       & 94.43       & 91.68      & 92.80          & 93.09          & 93.74          & 94.35           & 90.56           & 93.19        & 95.03        & 93.29        & 94.04        & 93.87        & 88.74        \\ \cline{2-19} 
                                                                                        & \multirow{4}{*}{\begin{tabular}[c]{@{}c@{}}CORE\\ ~\cite{ni2022core}\end{tabular}}         & AUC                              & 99.04             & 98.01            & 97.80       & 98.47       & 98.79      & 95.88          & 97.82          & 99.09          & 98.77           & 97.11           & 98.12        & 98.91        & 99.26        & 97.57        & 98.02        & 97.84        \\
                                                                                        &                                                                                & FPR                              & 10.73             & 16.52            & 21.46       & 14.76       & 10.68      & 43.07          & 19.42          & 9.19           & 12.34           & 20.10           & 18.63        & 12.14        & 8.45         & 23.79        & 17.34        & 14.25        \\
                                                                                        &                                                                                & TPR                              & 99.40             & 99.27            & 99.61       & 99.51       & 98.84      & 99.83          & 99.27          & 99.27          & 99.46           & 99.00           & 99.78        & 99.53        & 99.13        & 99.52        & 99.49        & 98.28        \\
                                                                                        &                                                                                & ACC                              & 96.88             & 95.49            & 95.01       & 96.34       & 95.97      & 88.37          & 94.61          & 97.08          & 97.13           & 92.49           & 94.91        & 96.87        & 96.95        & 95.07        & 95.85        & 94.17        \\ \hline
\multirow{12}{*}{\textbf{\begin{tabular}[c]{@{}c@{}}Fairness-\\ enhanced\end{tabular}}} & \multirow{4}{*}{\begin{tabular}[c]{@{}c@{}}DAW-FDD\\~\cite{ju2024improving}\end{tabular}}          & AUC                              & 98.36             & 97.15            & 96.84       & 97.45       & 98.51      & 94.97          & 96.41          & 98.62          & 98.00           & 94.95           & 96.83        & 98.05        & 98.85        & 96.77        & 96.87        & 97.91        \\
                                                                                        &                                                                                & FPR                              & 14.12             & 19.63            & 26.93       & 18.59       & 12.81      & 56.69          & 23.64          & 11.23          & 15.40           & 26.17           & 26.40        & 16.16        & 10.80        & 27.37        & 20.98        & 16.03        \\
                                                                                        &                                                                                & TPR                              & 99.42             & 99.24            & 99.73       & 99.38       & 99.20      & 99.77          & 99.18          & 99.28          & 99.51           & 98.98           & 99.78        & 99.47        & 99.32        & 99.70        & 99.29        & 98.97        \\
                                                                                        &                                                                                & ACC                              & 96.05             & 94.73            & 93.91       & 95.39       & 95.58      & 84.69          & 93.49          & 96.56          & 96.56           & 90.41           & 92.86        & 95.90        & 96.41        & 94.53        & 94.91        & 94.06        \\ \cline{2-19} 
                                                                                        & \multirow{4}{*}{\begin{tabular}[c]{@{}c@{}}DAG-FDD\\ ~\cite{ju2024improving}\end{tabular}}          & AUC                              & 99.05             & 98.44            & 97.56       & 98.79       & 98.73      & 96.55          & 98.14          & 99.10          & 98.91           & 97.59           & 98.11        & 99.00        & 99.16        & 97.10        & 98.57        & 97.83        \\
                                                                                        &                                                                                & FPR                              & 12.11             & 18.02            & 20.76       & 15.10       & 14.21      & 26.95          & 20.01          & 11.72          & 14.61           & 22.40           & 18.32        & 13.04        & 10.58        & 22.76        & 17.14        & 20.00        \\
                                                                                        &                                                                                & TPR                              & 99.21             & 99.05            & 98.98       & 99.25       & 98.83      & 97.64          & 98.92          & 99.18          & 99.37           & 99.13           & 98.77        & 99.33        & 98.98        & 99.10        & 99.17        & 98.54        \\
                                                                                        &                                                                                & ACC                              & 96.39             & 94.97            & 94.67       & 96.06       & 94.90      & 91.07          & 94.20          & 96.36          & 96.61           & 91.79           & 94.26        & 96.51        & 96.23        & 94.92        & 95.65        & 92.47        \\ \cline{2-19} 
                                                                                        & \multirow{4}{*}{\begin{tabular}[c]{@{}c@{}}PG-FDD\\ ~\cite{lin2024preserving}\end{tabular}}      & AUC                              & 99.36             & 98.94            & 98.94       & 99.18       & 99.13      & 98.83          & 98.89          & 99.35          & 99.27           & 97.85           & 98.97        & 99.35        & 99.39        & 98.89        & 99.02        & 98.51        \\
                                                                                        &                                                                                & FPR                              & 9.49              & 12.68            & 15.29       & 12.06       & 8.82       & 22.77          & 14.31          & 7.97           & 11.64           & 19.74           & 13.35        & 10.90        & 7.30         & 16.88        & 13.20        & 11.26        \\
                                                                                        &                                                                                & TPR                              & 98.73             & 98.21            & 98.63       & 98.85       & 97.43      & 99.11          & 98.44          & 98.28          & 98.87           & 97.74           & 98.66        & 99.02        & 98.12        & 98.61        & 98.69        & 96.10        \\
                                                                                        &                                                                                & ACC                              & 96.68             & 95.61            & 95.59       & 96.43       & 95.55      & 93.27          & 95.26          & 96.66          & 96.79           & 91.78           & 95.49        & 96.75        & 96.56        & 95.65        & 96.12        & 93.69        \\ \hline
\end{tabular}
}
\caption{\small \textit{Detailed test results of each subgroup of each detector on the AI-Face. In the Skin Tone groups, `L' represents Light (Tone 1-3), `M' is Medium (4-6), `D' is Dark (Tone 7-10).}}
\label{tab:detailed_overall_results}
\end{table*}

\begin{table*}[t]
\centering
\scalebox{0.675}{
\begin{tabular}{c|c|c|cccccccccccc}
\hline
\multirow{3}{*}{Measure}       & \multirow{3}{*}{Attribute}    & \multirow{3}{*}{Metric} & \multicolumn{12}{c}{Model Type}                                                                                                                                                                                                                                                                                                                                                                                                                                                                                                                                                                                                                                                                                                                                          \\ \cline{4-15} 
                               &                               &                         & \multicolumn{3}{c|}{Native}                                                                                                                                                                           & \multicolumn{3}{c|}{Frequency}                                                                                                                                                             & \multicolumn{3}{c|}{Spatial}                                                                                                                                                                & \multicolumn{3}{c}{Fairness-enhanced}                                                                                                                                 \\ \cline{4-15} 
                               &                               &                         & \begin{tabular}[c]{@{}c@{}}Xception\\ ~\cite{chollet2017xception}\end{tabular}                               & \begin{tabular}[c]{@{}c@{}}EfficientB4\\~\cite{tan2019efficientnet}\end{tabular}                              & \multicolumn{1}{c|}{\begin{tabular}[c]{@{}c@{}}ViT-B/16\\~\cite{dosovitskiy2020image}\end{tabular}} & \begin{tabular}[c]{@{}c@{}}F3Net\\~\cite{qian2020thinking}\end{tabular}  & \begin{tabular}[c]{@{}c@{}}SPSL\\~\cite{liu2021spatial}\end{tabular}                                    & \multicolumn{1}{c|}{\begin{tabular}[c]{@{}c@{}}SRM\\~\cite{luo2021generalizing}\end{tabular}} & \begin{tabular}[c]{@{}c@{}}UCF\\~\cite{yan2023ucf}\end{tabular}    & \begin{tabular}[c]{@{}c@{}}UnivFD\\~\cite{ojha2023towards}\end{tabular}                                 & \multicolumn{1}{c|}{\begin{tabular}[c]{@{}c@{}}CORE\\~\cite{ni2022core}\end{tabular}} & \begin{tabular}[c]{@{}c@{}}DAW-FDD\\~\cite{ju2024improving}\end{tabular} &\begin{tabular}[c]{@{}c@{}}DAG-FDD\\~\cite{ju2024improving}\end{tabular}                                & \begin{tabular}[c]{@{}c@{}}PG-FDD\\~\cite{lin2024preserving}\end{tabular}                                  \\ \hline \hline
\multirow{16}{*}{Fairness(\%)} & \multirow{4}{*}{Skin Tone}    & $F_{MEO}$                    & 10.901                                                    & 4.384                                                         & \multicolumn{1}{c|}{17.219}                                               & 14.583                                                 & 9.620                                                 & \multicolumn{1}{c|}{15.508}                                               & 14.978                                               & 2.441                                                   & \multicolumn{1}{c|}{13.135}                                                & 12.519                                               & 12.597                                               & 13.965                                                  \\
                               &                               & $F_{DP}$                     & 11.274                                                    & 9.191                                                         & \multicolumn{1}{c|}{10.713}                                               & 12.739                                                 & 11.711                                                & \multicolumn{1}{c|}{11.282}                                               & 11.549                                               & 8.117                                                   & \multicolumn{1}{c|}{11.968}                                                & 11.179                                               & 12.026                                               & 11.768                                                  \\
                               &                               & $F_{OAE}$                    & 2.434                                                     & 3.609                                                         & \multicolumn{1}{c|}{2.276}                                                & 2.232                                                  & 2.814                                                 & \multicolumn{1}{c|}{1.780}                                                & 1.950                                                & 2.940                                                   & \multicolumn{1}{c|}{1.658}                                                 & 0.878                                                & 1.753                                                & 1.439                                                   \\
                               &                               & $F_{EO}$                     & 0.160                                                     & 0.093                                                         & \multicolumn{1}{c|}{0.205}                                                & 0.209                                                  & 0.156                                                 & \multicolumn{1}{c|}{0.191}                                                & 0.186                                                & 0.034                                                   & \multicolumn{1}{c|}{0.176}                                                 & 0.165                                                & 0.176                                                & 0.176                                                   \\ \cline{2-15} 
                               & \multirow{4}{*}{Gender}       & $F_{MEO}$                    & 5.475                                                     & 5.458                                                         & \multicolumn{1}{c|}{8.003}                                                & 5.749                                                  & 5.754                                                 & \multicolumn{1}{c|}{5.848}                                                & 5.575                                                & 3.244                                                   & \multicolumn{1}{c|}{4.367}                                                 & 5.186                                                & 5.808                                                & 4.086                                                   \\
                               &                               & $F_{DP}$                     & 1.205                                                     & 1.412                                                         & \multicolumn{1}{c|}{2.416}                                                & 1.340                                                  & 1.445                                                 & \multicolumn{1}{c|}{1.959}                                                & 1.810                                                & 0.781                                                   & \multicolumn{1}{c|}{0.980}                                                 & 1.715                                                & 1.470                                                & 1.545                                                   \\
                               &                               & $F_{OAE}$                    & 2.043                                                     & 1.800                                                         & \multicolumn{1}{c|}{1.896}                                                & 2.054                                                  & 1.969                                                 & \multicolumn{1}{c|}{1.471}                                                & 1.569                                                & 1.413                                                   & \multicolumn{1}{c|}{1.848}                                                 & 1.430                                                & 2.012                                                & 1.133                                                   \\
                               &                               & $F_{EO}$                     & 0.066                                                     & 0.063                                                         & \multicolumn{1}{c|}{0.083}                                                & 0.068                                                  & 0.067                                                 & \multicolumn{1}{c|}{0.062}                                                & 0.060                                                & 0.043                                                   & \multicolumn{1}{c|}{0.056}                                                 & 0.056                                                & 0.068                                                & 0.044                                                   \\ \cline{2-15} 
                               & \multirow{4}{*}{Age}          & $F_{MEO}$                    & 28.244                                                    & 7.460                                                         & \multicolumn{1}{c|}{38.521}                                               & 27.860                                                 & 24.768                                                & \multicolumn{1}{c|}{40.542}                                               & 44.342                                               & 8.584                                                   & \multicolumn{1}{c|}{34.156}                                                & 36.450                                               & 35.031                                               & 36.197                                                  \\
                               &                               & $F_{DP}$                     & 11.228                                                    & 12.245                                                        & \multicolumn{1}{c|}{12.140}                                               & 11.395                                                 & 11.466                                                & \multicolumn{1}{c|}{14.564}                                               & 15.856                                               & 16.134                                                  & \multicolumn{1}{c|}{13.525}                                                & 12.256                                               & 13.478                                               & 12.082                                                  \\
                               &                               & $F_{OAE}$                    & 7.138                                                     & 5.234                                                         & \multicolumn{1}{c|}{10.940}                                               & 6.933                                                  & 6.053                                                 & \multicolumn{1}{c|}{11.126}                                               & 11.481                                               & 4.171                                                   & \multicolumn{1}{c|}{8.294}                                                 & 9.192                                                & 8.636                                                & 8.934                                                   \\
                               &                               & $F_{EO}$                     & 0.460                                                     & 0.175                                                         & \multicolumn{1}{c|}{0.560}                                                & 0.460                                                  & 0.410                                                 & \multicolumn{1}{c|}{0.550}                                                & 0.610                                                & 0.191                                                   & \multicolumn{1}{c|}{0.508}                                                 & 0.537                                                & 0.524                                                & 0.539                                                   \\ \cline{2-15} 
                               & \multirow{4}{*}{Intersection} & $F_{MEO}$                    & 15.752                                                    & 10.644                                                        & \multicolumn{1}{c|}{24.460}                                               & 18.455                                                 & 15.157                                                & \multicolumn{1}{c|}{18.381}                                               & 17.397                                               & 5.300                                                   & \multicolumn{1}{c|}{16.257}                                                & 14.806                                               & 15.219                                               & 16.517                                                  \\
                               &                               & $F_{DP}$                     & 16.943                                                    & 14.565                                                        & \multicolumn{1}{c|}{13.773}                                               & 18.071                                                 & 17.490                                                & \multicolumn{1}{c|}{14.943}                                               & 15.612                                               & 12.967                                                  & \multicolumn{1}{c|}{17.063}                                                & 14.802                                               & 16.786                                               & 15.513                                                  \\
                               &                               & $F_{OAE}$                    & 6.805                                                     & 8.029                                                         & \multicolumn{1}{c|}{5.025}                                                & 6.658                                                  & 7.200                                                 & \multicolumn{1}{c|}{3.614}                                                & 3.532                                                & 6.226                                                   & \multicolumn{1}{c|}{5.079}                                                 & 3.314                                                & 5.757                                                & 2.989                                                   \\
                               &                               & $F_{EO}$                     & 0.355                                                     & 0.307                                                         & \multicolumn{1}{c|}{0.441}                                                & 0.440                                                  & 0.366                                                 & \multicolumn{1}{c|}{0.382}                                                & 0.382                                                & 0.178                                                   & \multicolumn{1}{c|}{0.371}                                                 & 0.336                                                & 0.399                                                & 0.354                                                   \\ \hline
\multirow{5}{*}{Utility(\%)}   & \multirow{5}{*}{-}            & AUC                     & 0.968                                                     & 0.968                                                         & \multicolumn{1}{c|}{0.981}                                                & 0.966                                                  & 0.968                                                 & \multicolumn{1}{c|}{0.967}                                                & 0.977                                                & 0.979                                                   & \multicolumn{1}{c|}{0.975}                                                 & 0.970                                                & 0.973                                                & 0.978                                                   \\
                               &                               & ACC                     & 0.931                                                     & 0.922                                                         & \multicolumn{1}{c|}{0.924}                                                & 0.930                                                  & 0.929                                                 & \multicolumn{1}{c|}{0.946}                                                & 0.951                                                & 0.933                                                   & \multicolumn{1}{c|}{0.947}                                                 & 0.941                                                & 0.941                                                & 0.952                                                   \\
                               &                               & AP                      & 0.987                                                     & 0.988                                                         & \multicolumn{1}{c|}{0.994}                                                & 0.986                                                  & 0.988                                                 & \multicolumn{1}{c|}{0.985}                                                & 0.991                                                & 0.993                                                   & \multicolumn{1}{c|}{0.989}                                                 & 0.987                                                & 0.989                                                & 0.991                                                   \\
                               &                               & EER                     & 0.093                                                     & 0.101                                                         & \multicolumn{1}{c|}{0.082}                                                & 0.096                                                  & 0.095                                                 & \multicolumn{1}{c|}{0.076}                                                & 0.074                                                & 0.083                                                   & \multicolumn{1}{c|}{0.076}                                                 & 0.085                                                & 0.084                                                & 0.072                                                   \\
                               &                               & FPR                     & 0.205                                                     & 0.219                                                         & \multicolumn{1}{c|}{0.290}                                                & 0.199                                                  & 0.190                                                 & \multicolumn{1}{c|}{0.205}                                                & 0.163                                                & 0.188                                                   & \multicolumn{1}{c|}{0.144}                                                 & 0.186                                                & 0.168                                                & 0.151                                                   \\ \hline
\end{tabular}
}
\caption{\small \textit{Detailed fairness and utility evaluation results on 20\% training subset.}}
\label{tab:20_trainset}
\end{table*}

\begin{table*}[t]
\centering
\scalebox{0.675}{
\begin{tabular}{c|c|c|cccccccccccc}
\hline
\multirow{3}{*}{Measure}       & \multirow{3}{*}{Attribute}    & \multirow{3}{*}{Metric} & \multicolumn{12}{c}{Model Type}                                                            \\ \cline{4-15} 
                               &                               &                         & \multicolumn{3}{c|}{Native}                                                                                                                                                                           & \multicolumn{3}{c|}{Frequency}                                                                                                                                                             & \multicolumn{3}{c|}{Spatial}                                                                                                                                                                & \multicolumn{3}{c}{Fairness-enhanced}                                                                                                                                 \\ \cline{4-15} 
                               &                               &                         & \begin{tabular}[c]{@{}c@{}}Xception\\ ~\cite{chollet2017xception}\end{tabular}                               & \begin{tabular}[c]{@{}c@{}}EfficientB4\\~\cite{tan2019efficientnet}\end{tabular}                              & \multicolumn{1}{c|}{\begin{tabular}[c]{@{}c@{}}ViT-B/16\\~\cite{dosovitskiy2020image}\end{tabular}} & \begin{tabular}[c]{@{}c@{}}F3Net\\~\cite{qian2020thinking}\end{tabular}  & \begin{tabular}[c]{@{}c@{}}SPSL\\~\cite{liu2021spatial}\end{tabular}                                    & \multicolumn{1}{c|}{\begin{tabular}[c]{@{}c@{}}SRM\\~\cite{luo2021generalizing}\end{tabular}} & \begin{tabular}[c]{@{}c@{}}UCF\\~\cite{yan2023ucf}\end{tabular}    & \begin{tabular}[c]{@{}c@{}}UnivFD\\~\cite{ojha2023towards}\end{tabular}                                 & \multicolumn{1}{c|}{\begin{tabular}[c]{@{}c@{}}CORE\\~\cite{ni2022core}\end{tabular}} & \begin{tabular}[c]{@{}c@{}}DAW-FDD\\~\cite{ju2024improving}\end{tabular} &\begin{tabular}[c]{@{}c@{}}DAG-FDD\\~\cite{ju2024improving}\end{tabular}                                & \begin{tabular}[c]{@{}c@{}}PG-FDD\\~\cite{lin2024preserving}\end{tabular}                                  \\ \hline \hline
\multirow{16}{*}{Fairness(\%)} & \multirow{4}{*}{Skin Tone}    & $F_{MEO}$                    & 9.815                                                     & 4.414                                                         & \multicolumn{1}{c|}{12.194}                                               & 10.801                                                 & 9.275                                                 & \multicolumn{1}{c|}{24.037}                                               & 12.299                                               & 2.568                                                   & \multicolumn{1}{c|}{15.734}                                                & 12.628                                               & 10.463                                               & 10.982                                                  \\
                               &                               & $F_{DP}$                     & 10.080                                                    & 10.413                                                        & \multicolumn{1}{c|}{9.475}                                                & 11.137                                                 & 10.644                                                & \multicolumn{1}{c|}{12.550}                                               & 10.632                                               & 8.322                                                   & \multicolumn{1}{c|}{11.251}                                                & 10.157                                               & 10.456                                               & 10.928                                                  \\
                               &                               & $F_{OAE}$                    & 0.122                                                     & 0.095                                                         & \multicolumn{1}{c|}{0.146}                                                & 0.154                                                  & 0.122                                                 & \multicolumn{1}{c|}{0.281}                                                & 0.150                                                & 0.043                                                   & \multicolumn{1}{c|}{0.180}                                                 & 0.154                                                & 0.135                                                & 0.143                                                   \\
                               &                               & $F_{EO}$                     & 1.472                                                     & 3.796                                                         & \multicolumn{1}{c|}{3.395}                                                & 2.088                                                  & 1.631                                                 & \multicolumn{1}{c|}{3.280}                                                & 1.533                                                & 2.898                                                   & \multicolumn{1}{c|}{2.384}                                                 & 1.571                                                & 1.323                                                & 1.188                                                   \\ \cline{2-15} 
                               & \multirow{4}{*}{Gender}       & $F_{DP}$                     & 5.576                                                     & 3.592                                                         & \multicolumn{1}{c|}{6.089}                                                & 5.985                                                  & 4.368                                                 & \multicolumn{1}{c|}{7.959}                                                & 4.400                                                & 3.438                                                   & \multicolumn{1}{c|}{5.234}                                                 & 5.960                                                & 4.797                                                & 5.390                                                   \\
                               &                               & $F_{MEO}$                    & 1.817                                                     & 0.822                                                         & \multicolumn{1}{c|}{1.853}                                                & 1.566                                                  & 1.227                                                 & \multicolumn{1}{c|}{2.658}                                                & 1.481                                                & 0.797                                                   & \multicolumn{1}{c|}{1.866}                                                 & 2.053                                                & 1.458                                                & 2.052                                                   \\
                               &                               & $F_{OAE}$                    & 1.559                                                     & 1.622                                                         & \multicolumn{1}{c|}{1.829}                                                & 1.966                                                  & 1.595                                                 & \multicolumn{1}{c|}{1.722}                                                & 1.338                                                & 1.507                                                   & \multicolumn{1}{c|}{1.369}                                                 & 1.473                                                & 1.530                                                & 1.303                                                   \\
                               &                               & $F_{EO}$                     & 0.060                                                     & 0.047                                                         & \multicolumn{1}{c|}{0.067}                                                & 0.069                                                  & 0.052                                                 & \multicolumn{1}{c|}{0.080}                                                & 0.049                                                & 0.045                                                   & \multicolumn{1}{c|}{0.055}                                                 & 0.062                                                & 0.055                                                & 0.055                                                   \\ \cline{2-15} 
                               & \multirow{4}{*}{Age}          & $F_{MEO}$                    & 32.781                                                    & 9.931                                                         & \multicolumn{1}{c|}{18.050}                                               & 26.665                                                 & 33.004                                                & \multicolumn{1}{c|}{54.967}                                               & 43.829                                               & 7.840                                                   & \multicolumn{1}{c|}{38.202}                                                & 41.707                                               & 34.285                                               & 33.582                                                  \\
                               &                               & $F_{DP}$                     & 12.161                                                    & 12.428                                                        & \multicolumn{1}{c|}{14.762}                                               & 10.954                                                 & 13.272                                                & \multicolumn{1}{c|}{16.006}                                               & 14.243                                               & 16.076                                                  & \multicolumn{1}{c|}{13.394}                                                & 13.661                                               & 11.630                                               & 12.955                                                  \\
                               &                               & $F_{OAE}$                    & 8.535                                                     & 4.312                                                         & \multicolumn{1}{c|}{6.949}                                                & 6.831                                                  & 8.210                                                 & \multicolumn{1}{c|}{14.795}                                               & 11.248                                               & 4.185                                                   & \multicolumn{1}{c|}{9.906}                                                 & 10.841                                               & 8.738                                                & 8.520                                                   \\
                               &                               & $F_{EO}$                     & 0.474                                                     & 0.235                                                         & \multicolumn{1}{c|}{0.348}                                                & 0.432                                                  & 0.470                                                 & \multicolumn{1}{c|}{0.738}                                                & 0.611                                                & 0.179                                                   & \multicolumn{1}{c|}{0.541}                                                 & 0.585                                                & 0.509                                                & 0.500                                                   \\ \cline{2-15} 
                               & \multirow{4}{*}{Intersection} & $F_{MEO}$                    & 13.451                                                    & 8.558                                                         & \multicolumn{1}{c|}{15.615}                                               & 13.559                                                 & 12.356                                                & \multicolumn{1}{c|}{30.133}                                               & 15.278                                               & 5.585                                                   & \multicolumn{1}{c|}{19.342}                                                & 17.006                                               & 12.784                                               & 13.197                                                  \\
                               &                               & $F_{DP}$                     & 13.795                                                    & 16.478                                                        & \multicolumn{1}{c|}{14.409}                                               & 16.096                                                 & 15.138                                                & \multicolumn{1}{c|}{15.893}                                               & 14.797                                               & 13.263                                                  & \multicolumn{1}{c|}{15.123}                                                & 14.586                                               & 14.345                                               & 15.166                                                  \\
                               &                               & $F_{OAE}$                    & 4.424                                                     & 7.462                                                         & \multicolumn{1}{c|}{5.221}                                                & 6.173                                                  & 4.728                                                 & \multicolumn{1}{c|}{5.030}                                                & 2.886                                                & 6.298                                                   & \multicolumn{1}{c|}{3.741}                                                 & 3.541                                                & 4.402                                                & 2.911                                                   \\
                               &                               & $F_{EO}$                     & 0.312                                                     & 0.236                                                         & \multicolumn{1}{c|}{0.416}                                                & 0.360                                                  & 0.309                                                 & \multicolumn{1}{c|}{0.550}                                                & 0.304                                                & 0.185                                                   & \multicolumn{1}{c|}{0.374}                                                 & 0.324                                                & 0.310                                                & 0.301                                                   \\ \hline
\multirow{5}{*}{Utility(\%)}   & \multirow{5}{*}{-}            & AUC                     & 0.979                                                     & 0.976                                                         & \multicolumn{1}{c|}{0.980}                                                & 0.974                                                  & 0.982                                                 & \multicolumn{1}{c|}{0.957}                                                & 0.979                                                & 0.979                                                   & \multicolumn{1}{c|}{0.981}                                                 & 0.976                                                & 0.978                                                & 0.982                                                   \\
                               &                               & ACC                     & 0.951                                                     & 0.940                                                         & \multicolumn{1}{c|}{0.933}                                                & 0.937                                                  & 0.951                                                 & \multicolumn{1}{c|}{0.938}                                                & 0.955                                                & 0.934                                                   & \multicolumn{1}{c|}{0.957}                                                 & 0.948                                                & 0.949                                                & 0.960                                                   \\
                               &                               & AP                      & 0.991                                                     & 0.991                                                         & \multicolumn{1}{c|}{0.993}                                                & 0.989                                                  & 0.993                                                 & \multicolumn{1}{c|}{0.983}                                                & 0.992                                                & 0.993                                                   & \multicolumn{1}{c|}{0.991}                                                 & 0.989                                                & 0.991                                                & 0.992                                                   \\
                               &                               & EER                     & 0.068                                                     & 0.078                                                         & \multicolumn{1}{c|}{0.081}                                                & 0.083                                                  & 0.066                                                 & \multicolumn{1}{c|}{0.117}                                                & 0.072                                                & 0.082                                                   & \multicolumn{1}{c|}{0.058}                                                 & 0.074                                                & 0.074                                                & 0.055                                                   \\
                               &                               & FPR                     & 0.165                                                     & 0.147                                                         & \multicolumn{1}{c|}{0.136}                                                & 0.186                                                  & 0.147                                                 & \multicolumn{1}{c|}{0.245}                                                & 0.157                                                & 0.180                                                   & \multicolumn{1}{c|}{0.151}                                                 & 0.184                                                & 0.169                                                & 0.130                                                   \\ \hline
\end{tabular}
}
\caption{\small \textit{Detailed fairness and utility evaluation results on 40\% training subset.}}
\label{tab:40_trainset}
\end{table*}

\begin{table*}[t]
\centering
\scalebox{0.675}{
\begin{tabular}{c|c|c|cccccccccccc}
\hline
\multirow{3}{*}{Measure}       & \multirow{3}{*}{Attribute}    & \multirow{3}{*}{Metric} & \multicolumn{12}{c}{Model Type}                                                           \\ \cline{4-15} 
                               &                               &                         & \multicolumn{3}{c|}{Native}                                                                                                                                                                           & \multicolumn{3}{c|}{Frequency}                                                                                                                                                             & \multicolumn{3}{c|}{Spatial}                                                                                                                                                                & \multicolumn{3}{c}{Fairness-enhanced}                                                                                                                                 \\ \cline{4-15} 
                               &                               &                         & \begin{tabular}[c]{@{}c@{}}Xception\\ ~\cite{chollet2017xception}\end{tabular}                               & \begin{tabular}[c]{@{}c@{}}EfficientB4\\~\cite{tan2019efficientnet}\end{tabular}                              & \multicolumn{1}{c|}{\begin{tabular}[c]{@{}c@{}}ViT-B/16\\~\cite{dosovitskiy2020image}\end{tabular}} & \begin{tabular}[c]{@{}c@{}}F3Net\\~\cite{qian2020thinking}\end{tabular}  & \begin{tabular}[c]{@{}c@{}}SPSL\\~\cite{liu2021spatial}\end{tabular}                                    & \multicolumn{1}{c|}{\begin{tabular}[c]{@{}c@{}}SRM\\~\cite{luo2021generalizing}\end{tabular}} & \begin{tabular}[c]{@{}c@{}}UCF\\~\cite{yan2023ucf}\end{tabular}    & \begin{tabular}[c]{@{}c@{}}UnivFD\\~\cite{ojha2023towards}\end{tabular}                                 & \multicolumn{1}{c|}{\begin{tabular}[c]{@{}c@{}}CORE\\~\cite{ni2022core}\end{tabular}} & \begin{tabular}[c]{@{}c@{}}DAW-FDD\\~\cite{ju2024improving}\end{tabular} &\begin{tabular}[c]{@{}c@{}}DAG-FDD\\~\cite{ju2024improving}\end{tabular}                                & \begin{tabular}[c]{@{}c@{}}PG-FDD\\~\cite{lin2024preserving}\end{tabular}                                  \\ \hline \hline
\multirow{16}{*}{Fairness(\%)} & \multirow{4}{*}{Skin Tone}    & $F_{MEO}$                    & 9.086                                                     & 14.704                                                        & \multicolumn{1}{c|}{4.388}                                                & 15.303                                                 & 6.813                                                 & \multicolumn{1}{c|}{14.516}                                               & 14.952                                               & 2.186                                                   & \multicolumn{1}{c|}{9.689}                                                 & 13.488                                               & 9.672                                                & 4.108                                                   \\
                               &                               & $F_{DP}$                     & 10.232                                                    & 11.784                                                        & \multicolumn{1}{c|}{7.714}                                                & 11.225                                                 & 9.979                                                 & \multicolumn{1}{c|}{14.909}                                               & 13.116                                               & 8.004                                                   & \multicolumn{1}{c|}{10.666}                                                & 10.844                                               & 10.054                                               & 8.575                                                   \\
                               &                               & $F_{OAE}$                    & 1.531                                                     & 2.017                                                         & \multicolumn{1}{c|}{2.572}                                                & 2.320                                                  & 1.247                                                 & \multicolumn{1}{c|}{1.733}                                                & 1.562                                                & 2.777                                                   & \multicolumn{1}{c|}{1.208}                                                 & 1.664                                                & 1.259                                                & 1.383                                                   \\
                               &                               & $F_{EO}$                     & 0.124                                                     & 0.194                                                         & \multicolumn{1}{c|}{0.084}                                                & 0.177                                                  & 0.100                                                 & \multicolumn{1}{c|}{0.234}                                                & 0.208                                                & 0.043                                                   & \multicolumn{1}{c|}{0.131}                                                 & 0.163                                                & 0.125                                                & 0.055                                                   \\ \cline{2-15} 
                               & \multirow{4}{*}{Gender}       & $F_{MEO}$                    & 4.418                                                     & 6.743                                                         & \multicolumn{1}{c|}{8.445}                                                & 5.545                                                  & 5.242                                                 & \multicolumn{1}{c|}{7.331}                                                & 5.713                                                & 4.182                                                   & \multicolumn{1}{c|}{4.395}                                                 & 5.579                                                & 4.978                                                & 2.430                                                   \\
                               &                               & $F_{DP}$                     & 1.243                                                     & 2.063                                                         & \multicolumn{1}{c|}{2.697}                                                & 2.096                                                  & 1.736                                                 & \multicolumn{1}{c|}{3.656}                                                & 1.845                                                & 1.142                                                   & \multicolumn{1}{c|}{1.622}                                                 & 2.153                                                & 1.600                                                & 1.061                                                   \\
                               &                               & $F_{OAE}$                    & 1.567                                                     & 1.846                                                         & \multicolumn{1}{c|}{2.052}                                                & 1.318                                                  & 1.499                                                 & \multicolumn{1}{c|}{0.630}                                                & 1.651                                                & 1.489                                                   & \multicolumn{1}{c|}{1.217}                                                 & 1.267                                                & 1.470                                                & 0.858                                                   \\
                               &                               & $F_{EO}$                     & 0.053                                                     & 0.072                                                         & \multicolumn{1}{c|}{0.086}                                                & 0.057                                                  & 0.057                                                 & \multicolumn{1}{c|}{0.087}                                                & 0.062                                                & 0.050                                                   & \multicolumn{1}{c|}{0.048}                                                 & 0.056                                                & 0.055                                                & 0.029                                                   \\ \cline{2-15} 
                               & \multirow{4}{*}{Age}          & $F_{MEO}$                    & 35.231                                                    & 27.998                                                        & \multicolumn{1}{c|}{17.573}                                               & 42.366                                                 & 35.428                                                & \multicolumn{1}{c|}{37.043}                                               & 34.243                                               & 5.520                                                   & \multicolumn{1}{c|}{28.666}                                                & 40.326                                               & 38.409                                               & 24.355                                                  \\
                               &                               & $F_{DP}$                     & 12.874                                                    & 12.663                                                        & \multicolumn{1}{c|}{11.691}                                               & 15.070                                                 & 12.924                                                & \multicolumn{1}{c|}{16.570}                                               & 13.579                                               & 15.256                                                  & \multicolumn{1}{c|}{11.264}                                                & 13.929                                               & 13.112                                               & 10.449                                                  \\
                               &                               & $F_{OAE}$                    & 8.954                                                     & 7.379                                                         & \multicolumn{1}{c|}{7.004}                                                & 10.921                                                 & 8.915                                                 & \multicolumn{1}{c|}{8.078}                                                & 8.232                                                & 3.900                                                   & \multicolumn{1}{c|}{7.519}                                                 & 10.379                                               & 9.909                                                & 6.870                                                   \\
                               &                               & $F_{EO}$                     & 0.514                                                     & 0.411                                                         & \multicolumn{1}{c|}{0.320}                                                & 0.585                                                  & 0.519                                                 & \multicolumn{1}{c|}{0.528}                                                & 0.520                                                & 0.134                                                   & \multicolumn{1}{c|}{0.438}                                                 & 0.570                                                & 0.558                                                & 0.362                                                   \\ \cline{2-15} 
                               & \multirow{4}{*}{Intersection} & $F_{MEO}$                    & 11.554                                                    & 18.923                                                        & \multicolumn{1}{c|}{10.063}                                               & 18.907                                                 & 10.175                                                & \multicolumn{1}{c|}{20.404}                                               & 18.818                                               & 5.414                                                   & \multicolumn{1}{c|}{12.995}                                                & 15.944                                               & 12.552                                               & 5.425                                                   \\
                               &                               & $F_{DP}$                     & 14.625                                                    & 15.884                                                        & \multicolumn{1}{c|}{9.908}                                                & 15.240                                                 & 14.093                                                & \multicolumn{1}{c|}{18.967}                                               & 18.087                                               & 12.890                                                  & \multicolumn{1}{c|}{15.331}                                                & 14.584                                               & 13.949                                               & 13.379                                                  \\
                               &                               & $F_{OAE}$                    & 4.755                                                     & 4.997                                                         & \multicolumn{1}{c|}{5.459}                                                & 3.617                                                  & 4.210                                                 & \multicolumn{1}{c|}{2.965}                                                & 4.832                                                & 6.152                                                   & \multicolumn{1}{c|}{3.747}                                                 & 3.106                                                & 4.102                                                & 3.306                                                   \\
                               &                               & $F_{EO}$                     & 0.279                                                     & 0.405                                                         & \multicolumn{1}{c|}{0.311}                                                & 0.362                                                  & 0.251                                                 & \multicolumn{1}{c|}{0.477}                                                & 0.431                                                & 0.200                                                   & \multicolumn{1}{c|}{0.276}                                                 & 0.331                                                & 0.282                                                & 0.159                                                   \\ \hline
\multirow{5}{*}{Utility(\%)}   & \multirow{5}{*}{-}            & AUC                     & 0.976                                                     & 0.980                                                         & \multicolumn{1}{c|}{0.986}                                                & 0.981                                                  & 0.983                                                 & \multicolumn{1}{c|}{0.976}                                                & 0.981                                                & 0.981                                                   & \multicolumn{1}{c|}{0.982}                                                 & 0.977                                                & 0.982                                                & 0.982                                                   \\
                               &                               & ACC                     & 0.948                                                     & 0.945                                                         & \multicolumn{1}{c|}{0.943}                                                & 0.961                                                  & 0.952                                                 & \multicolumn{1}{c|}{0.927}                                                & 0.951                                                & 0.935                                                   & \multicolumn{1}{c|}{0.956}                                                 & 0.960                                                & 0.950                                                & 0.960                                                   \\
                               &                               & AP                      & 0.989                                                     & 0.992                                                         & \multicolumn{1}{c|}{0.996}                                                & 0.991                                                  & 0.993                                                 & \multicolumn{1}{c|}{0.991}                                                & 0.992                                                & 0.994                                                   & \multicolumn{1}{c|}{0.992}                                                 & 0.987                                                & 0.993                                                & 0.991                                                   \\
                               &                               & EER                     & 0.074                                                     & 0.071                                                         & \multicolumn{1}{c|}{0.065}                                                & 0.058                                                  & 0.067                                                 & \multicolumn{1}{c|}{0.092}                                                & 0.068                                                & 0.078                                                   & \multicolumn{1}{c|}{0.062}                                                 & 0.060                                                & 0.069                                                & 0.058                                                   \\
                               &                               & FPR                     & 0.166                                                     & 0.183                                                         & \multicolumn{1}{c|}{0.177}                                                & 0.137                                                  & 0.154                                                 & \multicolumn{1}{c|}{0.137}                                                & 0.137                                                & 0.198                                                   & \multicolumn{1}{c|}{0.145}                                                 & 0.143                                                & 0.172                                                & 0.127                                                   \\ \hline
\end{tabular}
}
\caption{\small \textit{Detailed fairness and utility evaluation results on 60\% training subset.}}
\label{tab:60_trainset}
\end{table*}

\begin{table*}[t]
\centering
\scalebox{0.675}{
\begin{tabular}{c|c|c|cccccccccccc}
\hline
\multirow{3}{*}{Measure}       & \multirow{3}{*}{Attribute}    & \multirow{3}{*}{Metric} & \multicolumn{12}{c}{Model Type}                                                                                    \\ \cline{4-15} 
                               &                               &                         & \multicolumn{3}{c|}{Native}                                                                                                                                                                           & \multicolumn{3}{c|}{Frequency}                                                                                                                                                             & \multicolumn{3}{c|}{Spatial}                                                                                                                                                                & \multicolumn{3}{c}{Fairness-enhanced}                                                                                                                                 \\ \cline{4-15} 
                               &                               &                         & \begin{tabular}[c]{@{}c@{}}Xception\\ ~\cite{chollet2017xception}\end{tabular}                               & \begin{tabular}[c]{@{}c@{}}EfficientB4\\~\cite{tan2019efficientnet}\end{tabular}                              & \multicolumn{1}{c|}{\begin{tabular}[c]{@{}c@{}}ViT-B/16\\~\cite{dosovitskiy2020image}\end{tabular}} & \begin{tabular}[c]{@{}c@{}}F3Net\\~\cite{qian2020thinking}\end{tabular}  & \begin{tabular}[c]{@{}c@{}}SPSL\\~\cite{liu2021spatial}\end{tabular}                                    & \multicolumn{1}{c|}{\begin{tabular}[c]{@{}c@{}}SRM\\~\cite{luo2021generalizing}\end{tabular}} & \begin{tabular}[c]{@{}c@{}}UCF\\~\cite{yan2023ucf}\end{tabular}    & \begin{tabular}[c]{@{}c@{}}UnivFD\\~\cite{ojha2023towards}\end{tabular}                                 & \multicolumn{1}{c|}{\begin{tabular}[c]{@{}c@{}}CORE\\~\cite{ni2022core}\end{tabular}} & \begin{tabular}[c]{@{}c@{}}DAW-FDD\\~\cite{ju2024improving}\end{tabular} &\begin{tabular}[c]{@{}c@{}}DAG-FDD\\~\cite{ju2024improving}\end{tabular}                                & \begin{tabular}[c]{@{}c@{}}PG-FDD\\~\cite{lin2024preserving}\end{tabular}                                  \\ \hline \hline
\multirow{16}{*}{Fairness(\%)} & \multirow{4}{*}{Skin Tone}    & $F_{MEO}$                    & 15.463                                                    & 6.826                                                         & \multicolumn{1}{c|}{7.442}                                                & 13.642                                                 & 4.221                                                 & \multicolumn{1}{c|}{9.425}                                                & 13.574                                               & 2.368                                                   & \multicolumn{1}{c|}{13.487}                                                & 10.127                                               & 8.763                                                & 7.613                                                   \\
                               &                               & $F_{DP}$                     & 11.994                                                    & 10.440                                                        & \multicolumn{1}{c|}{8.378}                                                & 11.211                                                 & 8.998                                                 & \multicolumn{1}{c|}{6.875}                                                & 11.085                                               & 8.231                                                   & \multicolumn{1}{c|}{11.506}                                                & 10.182                                               & 8.920                                                & 11.914                                                  \\
                               &                               & $F_{OAE}$                    & 1.998                                                     & 2.390                                                         & \multicolumn{1}{c|}{1.540}                                                & 1.713                                                  & 2.141                                                 & \multicolumn{1}{c|}{7.511}                                                & 1.661                                                & 2.777                                                   & \multicolumn{1}{c|}{1.538}                                                 & 1.375                                                & 1.484                                                & 1.822                                                    \\
                               &                               & $F_{EO}$                     & 0.192                                                     & 0.107                                                         & \multicolumn{1}{c|}{0.095}                                                & 0.171                                                  & 0.064                                                 & \multicolumn{1}{c|}{0.187}                                                & 0.168                                                & 0.036                                                   & \multicolumn{1}{c|}{0.170}                                                 & 0.119                                                & 0.114                                                & 0.129                                                   \\ \cline{2-15} 
                               & \multirow{4}{*}{Gender}       & $F_{MEO}$                    & 4.209                                                     & 3.639                                                         & \multicolumn{1}{c|}{9.461}                                                & 5.189                                                  & 4.116                                                 & \multicolumn{1}{c|}{2.328}                                                & 5.402                                                & 4.084                                                   & \multicolumn{1}{c|}{5.058}                                                 & 4.112                                                & 4.143                                                & 4.035                                                   \\
                               &                               & $F_{DP}$                     & 1.171                                                     & 1.043                                                         & \multicolumn{1}{c|}{3.025}                                                & 1.749                                                  & 1.191                                                 & \multicolumn{1}{c|}{2.803}                                                & 1.778                                                & 1.082                                                   & \multicolumn{1}{c|}{1.899}                                                 & 1.537                                                & 1.560                                                & 1.401                                                   \\
                               &                               & $F_{OAE}$                    & 1.579                                                     & 1.395                                                         & \multicolumn{1}{c|}{2.163}                                                & 1.499                                                  & 1.507                                                 & \multicolumn{1}{c|}{1.248}                                                & 1.560                                                & 1.519                                                   & \multicolumn{1}{c|}{1.277}                                                 & 1.218                                                & 1.158                                                & 1.353                                                   \\
                               &                               & $F_{EO}$                     & 0.051                                                     & 0.045                                                         & \multicolumn{1}{c|}{0.095}                                                & 0.057                                                  & 0.050                                                 & \multicolumn{1}{c|}{0.037}                                                & 0.059                                                & 0.050                                                   & \multicolumn{1}{c|}{0.053}                                                 & 0.045                                                & 0.045                                                & 0.046                                                   \\ \cline{2-15} 
                               & \multirow{4}{*}{Age}          & $F_{MEO}$                    & 33.930                                                    & 16.272                                                        & \multicolumn{1}{c|}{10.222}                                               & 45.076                                                 & 20.048                                                & \multicolumn{1}{c|}{11.857}                                               & 45.508                                               & 5.788                                                   & \multicolumn{1}{c|}{37.055}                                                & 30.409                                               & 29.707                                               & 20.058                                                  \\
                               &                               & $F_{DP}$                     & 15.167                                                    & 11.254                                                        & \multicolumn{1}{c|}{11.643}                                               & 15.938                                                 & 11.379                                                & \multicolumn{1}{c|}{8.360}                                                & 15.357                                               & 15.620                                                  & \multicolumn{1}{c|}{13.565}                                                & 11.925                                               & 9.880                                                & 10.487                                                  \\
                               &                               & $F_{OAE}$                    & 8.520                                                     & 4.738                                                         & \multicolumn{1}{c|}{5.662}                                                & 11.409                                                 & 5.123                                                 & \multicolumn{1}{c|}{10.447}                                               & 11.546                                               & 3.777                                                   & \multicolumn{1}{c|}{9.454}                                                 & 7.784                                                & 8.513                                                & 4.450                                                   \\
                               &                               & $F_{EO}$                     & 0.488                                                     & 0.286                                                         & \multicolumn{1}{c|}{0.231}                                                & 0.623                                                  & 0.322                                                 & \multicolumn{1}{c|}{0.228}                                                & 0.625                                                & 0.136                                                   & \multicolumn{1}{c|}{0.526}                                                 & 0.441                                                & 0.459                                                & 0.357                                                   \\ \cline{2-15} 
                               & \multirow{4}{*}{Intersection} & $F_{MEO}$                    & 19.488                                                    & 11.396                                                        & \multicolumn{1}{c|}{16.807}                                               & 15.829                                                 & 6.631                                                 & \multicolumn{1}{c|}{12.599}                                               & 16.926                                               & 5.116                                                   & \multicolumn{1}{c|}{16.594}                                                & 13.597                                               & 11.226                                               & 10.523                                                  \\
                               &                               & $F_{DP}$                     & 17.093                                                    & 15.841                                                        & \multicolumn{1}{c|}{11.946}                                               & 15.708                                                 & 13.376                                                & \multicolumn{1}{c|}{11.010}                                               & 15.607                                               & 13.121                                                  & \multicolumn{1}{c|}{15.834}                                                & 14.758                                               & 12.474                                               & 16.906                                                  \\
                               &                               & $F_{OAE}$                    & 3.583                                                     & 5.700                                                         & \multicolumn{1}{c|}{5.469}                                                & 3.139                                                  & 5.452                                                 & \multicolumn{1}{c|}{10.020}                                               & 2.878                                                & 6.209                                                   & \multicolumn{1}{c|}{2.852}                                                 & 2.313                                                & 2.755                                                & 4.348                                                   \\
                               &                               & $F_{EO}$                     & 0.392                                                     & 0.262                                                         & \multicolumn{1}{c|}{0.340}                                                & 0.349                                                  & 0.225                                                 & \multicolumn{1}{c|}{0.420}                                                & 0.342                                                & 0.188                                                   & \multicolumn{1}{c|}{0.345}                                                 & 0.254                                                & 0.226                                                & 0.294                                                   \\ \hline
\multirow{5}{*}{Utility(\%)}   & \multirow{5}{*}{-}            & AUC                     & 0.985                                                     & 0.985                                                         & \multicolumn{1}{c|}{0.987}                                                & 0.981                                                  & 0.984                                                 & \multicolumn{1}{c|}{0.979}                                                & 0.982                                                & 0.981                                                   & \multicolumn{1}{c|}{0.984}                                                 & 0.981                                                & 0.987                                                & 0.988                                                   \\
                               &                               & ACC                     & 0.950                                                     & 0.949                                                         & \multicolumn{1}{c|}{0.940}                                                & 0.958                                                  & 0.950                                                 & \multicolumn{1}{c|}{0.816}                                                & 0.956                                                & 0.936                                                   & \multicolumn{1}{c|}{0.959}                                                 & 0.963                                                & 0.960                                                & 0.953                                                   \\
                               &                               & AP                      & 0.994                                                     & 0.994                                                         & \multicolumn{1}{c|}{0.996}                                                & 0.992                                                  & 0.994                                                 & \multicolumn{1}{c|}{0.992}                                                & 0.993                                                & 0.994                                                   & \multicolumn{1}{c|}{0.993}                                                 & 0.989                                                & 0.995                                                & 0.996                                                   \\
                               &                               & EER                     & 0.065                                                     & 0.064                                                         & \multicolumn{1}{c|}{0.062}                                                & 0.069                                                  & 0.064                                                 & \multicolumn{1}{c|}{0.066}                                                & 0.068                                                & 0.078                                                   & \multicolumn{1}{c|}{0.060}                                                 & 0.053                                                & 0.057                                                & 0.058                                                   \\
                               &                               & FPR                     & 0.145                                                     & 0.162                                                         & \multicolumn{1}{c|}{0.206}                                                & 0.134                                                  & 0.148                                                 & \multicolumn{1}{c|}{0.039}                                                & 0.141                                                & 0.189                                                   & \multicolumn{1}{c|}{0.138}                                                 & 0.118                                                & 0.144                                                & 0.099                                                   \\ \hline
\end{tabular}
}
\caption{\small \textit{Detailed fairness and utility evaluation results on 80\% training subset.}}
\label{tab:80_trainset}
\end{table*}

\begin{table*}[t]
\centering
\scalebox{0.675}{
\begin{tabular}{c|c|c|cccccccccccc}
\hline
\multirow{3}{*}{Measure}       & \multirow{3}{*}{Attribute}    & \multirow{3}{*}{Metric} & \multicolumn{12}{c}{Model Type}                                                                                                                                                                                                                                                                                                                                                                                                                                                                                                                                                                                                                                                                                                                                          \\ \cline{4-15} 
                               &                               &                         & \multicolumn{3}{c|}{Native}                                                                                                                                                                           & \multicolumn{3}{c|}{Frequency}                                                                                                                                                             & \multicolumn{3}{c|}{Spatial}                                                                                                                                                                & \multicolumn{3}{c}{Fairness-enhanced}                                                                                                                                 \\ \cline{4-15} 
                               &                               &                         & \begin{tabular}[c]{@{}c@{}}Xception\\ ~\cite{chollet2017xception}\end{tabular}                               & \begin{tabular}[c]{@{}c@{}}EfficientB4\\~\cite{tan2019efficientnet}\end{tabular}                              & \multicolumn{1}{c|}{\begin{tabular}[c]{@{}c@{}}ViT-B/16\\~\cite{dosovitskiy2020image}\end{tabular}} & \begin{tabular}[c]{@{}c@{}}F3Net\\~\cite{qian2020thinking}\end{tabular}  & \begin{tabular}[c]{@{}c@{}}SPSL\\~\cite{liu2021spatial}\end{tabular}                                    & \multicolumn{1}{c|}{\begin{tabular}[c]{@{}c@{}}SRM\\~\cite{luo2021generalizing}\end{tabular}} & \begin{tabular}[c]{@{}c@{}}UCF\\~\cite{yan2023ucf}\end{tabular}    & \begin{tabular}[c]{@{}c@{}}UnivFD\\~\cite{ojha2023towards}\end{tabular}                                 & \multicolumn{1}{c|}{\begin{tabular}[c]{@{}c@{}}CORE\\~\cite{ni2022core}\end{tabular}} & \begin{tabular}[c]{@{}c@{}}DAW-FDD\\~\cite{ju2024improving}\end{tabular} &\begin{tabular}[c]{@{}c@{}}DAG-FDD\\~\cite{ju2024improving}\end{tabular}                                & \begin{tabular}[c]{@{}c@{}}PG-FDD\\~\cite{lin2024preserving}\end{tabular}                                  \\ \hline \hline
                               
\multirow{16}{*}{Fairness(\%)} & \multirow{4}{*}{Skin Tone}    & $F_{MEO}$                    & 9.750                                                     & 11.908                                                        & \multicolumn{1}{c|}{5.095}                                                & 12.896                                                 & 5.859                                                 & \multicolumn{1}{c|}{18.083}                                               & 10.363                                               & 5.926                                                   & \multicolumn{1}{c|}{10.955}                                                & 9.570                                                & 8.980                                                & 3.889                                                   \\
                               &                               & $F_{DP}$                     & 16.824                                                    & 14.637                                                        & \multicolumn{1}{c|}{12.547}                                               & 17.621                                                 & 14.479                                                & \multicolumn{1}{c|}{19.676}                                               & 15.803                                               & 8.744                                                   & \multicolumn{1}{c|}{16.109}                                                & 16.063                                               & 15.700                                               & 12.899                                                  \\
                               &                               & $F_{OAE}$                    & 2.401                                                     & 2.973                                                         & \multicolumn{1}{c|}{1.714}                                                & 3.630                                                  & 0.905                                                 & \multicolumn{1}{c|}{5.752}                                                & 3.249                                                & 5.302                                                   & \multicolumn{1}{c|}{3.524}                                                 & 2.641                                                & 2.684                                                & 1.164                                                   \\
                               &                               & $F_{EO}$                     & 0.164                                                     & 0.155                                                         & \multicolumn{1}{c|}{0.119}                                                & 0.183                                                  & 0.113                                                 & \multicolumn{1}{c|}{0.248}                                                & 0.147                                                & 0.083                                                   & \multicolumn{1}{c|}{0.149}                                                 & 0.142                                                & 0.132                                                & 0.069                                                   \\ \cline{2-15} 
                               & \multirow{4}{*}{Gender}       & $F_{MEO}$                    & 3.535                                                     & 0.952                                                         & \multicolumn{1}{c|}{4.275}                                                & 4.788                                                  & 4.439                                                 & \multicolumn{1}{c|}{4.201}                                                & 4.655                                                & 6.898                                                   & \multicolumn{1}{c|}{3.981}                                                 & 3.848                                                & 3.674                                                & 3.792                                                   \\
                               &                               &  $F_{DP}$                     & 2.241                                                     & 0.715                                                         & \multicolumn{1}{c|}{3.129}                                                & 3.184                                                  & 3.060                                                 & \multicolumn{1}{c|}{2.245}                                                & 3.426                                                & 4.328                                                   & \multicolumn{1}{c|}{2.850}                                                 & 2.562                                                & 2.514                                                & 2.790                                                   \\
                               &                               & $F_{OAE}$                    & 2.482                                                     & 0.833                                                         & \multicolumn{1}{c|}{2.263}                                                & 2.690                                                  & 2.496                                                 & \multicolumn{1}{c|}{2.920}                                                & 2.332                                                & 3.028                                                   & \multicolumn{1}{c|}{2.236}                                                 & 2.451                                                & 2.360                                                & 2.166                                                   \\
                               &                               & $F_{EO}$                     & 0.051                                                     & 0.011                                                         & \multicolumn{1}{c|}{0.047}                                                & 0.057                                                  & 0.052                                                 & \multicolumn{1}{c|}{0.062}                                                & 0.049                                                & 0.070                                                   & \multicolumn{1}{c|}{0.048}                                                 & 0.051                                                & 0.049                                                & 0.045                                                   \\ \cline{2-15} 
                               & \multirow{4}{*}{Age}          & $F_{MEO}$                    & 23.282                                                    & 14.195                                                        & \multicolumn{1}{c|}{13.627}                                               & 32.679                                                 & 26.796                                                & \multicolumn{1}{c|}{52.939}                                               & 30.332                                               & 8.251                                                   & \multicolumn{1}{c|}{32.612}                                                & 23.934                                               & 23.834                                               & 15.408                                                  \\
                               &                               & $F_{DP}$                     & 17.521                                                    & 18.304                                                        & \multicolumn{1}{c|}{16.426}                                               & 20.437                                                 & 18.489                                                & \multicolumn{1}{c|}{28.303}                                               & 18.892                                               & 12.343                                                  & \multicolumn{1}{c|}{19.913}                                                & 17.103                                               & 18.221                                               & 15.255                                                  \\
                               &                               & $F_{OAE}$                    & 11.063                                                    & 3.474                                                         & \multicolumn{1}{c|}{9.185}                                                & 16.411                                                 & 13.102                                                & \multicolumn{1}{c|}{26.805}                                               & 15.234                                               & 6.602                                                   & \multicolumn{1}{c|}{16.794}                                                & 11.909                                               & 11.523                                               & 7.893                                                   \\
                               &                               & $F_{EO}$                     & 0.381                                                     & 0.203                                                         & \multicolumn{1}{c|}{0.303}                                                & 0.490                                                  & 0.436                                                 & \multicolumn{1}{c|}{0.769}                                                & 0.472                                                & 0.162                                                   & \multicolumn{1}{c|}{0.469}                                                 & 0.378                                                & 0.370                                                & 0.273                                                   \\ \cline{2-15} 
                               & \multirow{4}{*}{Intersection} & $F_{MEO}$                    & 11.921                                                    & 17.564                                                        & \multicolumn{1}{c|}{7.727}                                                & 17.622                                                 & 10.201                                                & \multicolumn{1}{c|}{21.679}                                               & 15.031                                               & 12.980                                                  & \multicolumn{1}{c|}{12.775}                                                & 13.010                                               & 11.104                                               & 7.173                                                   \\
                               &                               & $F_{DP}$                     & 23.484                                                    & 21.311                                                        & \multicolumn{1}{c|}{17.501}                                               & 23.519                                                 & 21.115                                                & \multicolumn{1}{c|}{24.119}                                               & 22.167                                               & 13.434                                                  & \multicolumn{1}{c|}{21.787}                                                & 22.542                                               & 22.120                                               & 18.247                                                  \\
                               &                               & $F_{OAE}$                    & 4.000                                                     & 4.650                                                         & \multicolumn{1}{c|}{3.969}                                                & 5.957                                                  & 4.306                                                 & \multicolumn{1}{c|}{9.974}                                                & 5.003                                                & 11.530                                                  & \multicolumn{1}{c|}{5.750}                                                 & 3.903                                                & 4.600                                                & 4.426                                                   \\
                               &                               & $F_{EO}$                     & 0.344                                                     & 0.335                                                         & \multicolumn{1}{c|}{0.304}                                                & 0.378                                                  & 0.250                                                 & \multicolumn{1}{c|}{0.515}                                                & 0.324                                                & 0.270                                                   & \multicolumn{1}{c|}{0.325}                                                 & 0.290                                                & 0.303                                                & 0.196                                                   \\ \hline
\multirow{5}{*}{Utility(\%)}   & \multirow{5}{*}{-}            & AUC                     & 0.984                                                     & 0.984                                                         & \multicolumn{1}{c|}{0.984}                                                & 0.982                                                  & 0.982                                                 & \multicolumn{1}{c|}{0.935}                                                & 0.981                                                & 0.979                                                   & \multicolumn{1}{c|}{0.986}                                                 & 0.983                                                & 0.987                                                & 0.989                                                   \\
                               &                               & ACC                     & 0.937                                                     & 0.892                                                         & \multicolumn{1}{c|}{0.922}                                                & 0.928                                                  & 0.929                                                 & \multicolumn{1}{c|}{0.900}                                                & 0.930                                                & 0.830                                                   & \multicolumn{1}{c|}{0.933}                                                 & 0.937                                                & 0.944                                                & 0.940                                                   \\
                               &                               & AP                      & 0.980                                                     & 0.986                                                         & \multicolumn{1}{c|}{0.986}                                                & 0.977                                                  & 0.978                                                 & \multicolumn{1}{c|}{0.910}                                                & 0.977                                                & 0.980                                                   & \multicolumn{1}{c|}{0.982}                                                 & 0.980                                                & 0.984                                                & 0.987                                                   \\
                               &                               & EER                     & 0.062                                                     & 0.064                                                         & \multicolumn{1}{c|}{0.067}                                                & 0.061                                                  & 0.066                                                 & \multicolumn{1}{c|}{0.129}                                                & 0.065                                                & 0.082                                                   & \multicolumn{1}{c|}{0.056}                                                 & 0.061                                                & 0.053                                                & 0.052                                                   \\
                               &                               & FPR                     & 0.087                                                     & 0.005                                                         & \multicolumn{1}{c|}{0.111}                                                & 0.122                                                  & 0.111                                                 & \multicolumn{1}{c|}{0.164}                                                & 0.116                                                & 0.337                                                   & \multicolumn{1}{c|}{0.115}                                                 & 0.095                                                & 0.083                                                & 0.095                                                   \\ \hline
\end{tabular}
}
\caption{\small \textit{Detailed fairness and utility evaluation results on a training subset with the ratio of real vs fake is 1:1.}}
\label{tab:ratio1_1}
\end{table*}

\begin{table*}[t]
\centering
\scalebox{0.675}{
\begin{tabular}{c|c|c|cccccccccccc}
\hline
\multirow{3}{*}{Measure}       & \multirow{3}{*}{Attribute}    & \multirow{3}{*}{Metric} & \multicolumn{12}{c}{Model Type}                                                                                                                                                                                                               \\ \cline{4-15} 
                               &                               &                         & \multicolumn{3}{c|}{Native}                                                                                                                                                                           & \multicolumn{3}{c|}{Frequency}                                                                                                                                                             & \multicolumn{3}{c|}{Spatial}                                                                                                                                                                & \multicolumn{3}{c}{Fairness-enhanced}                                                                                                                                 \\ \cline{4-15} 
                               &                               &                         &\begin{tabular}[c]{@{}c@{}}Xception\\ ~\cite{chollet2017xception}\end{tabular}                               & \begin{tabular}[c]{@{}c@{}}EfficientB4\\~\cite{tan2019efficientnet}\end{tabular}                              & \multicolumn{1}{c|}{\begin{tabular}[c]{@{}c@{}}ViT-B/16\\~\cite{dosovitskiy2020image}\end{tabular}} & \begin{tabular}[c]{@{}c@{}}F3Net\\~\cite{qian2020thinking}\end{tabular}  & \begin{tabular}[c]{@{}c@{}}SPSL\\~\cite{liu2021spatial}\end{tabular}                                    & \multicolumn{1}{c|}{\begin{tabular}[c]{@{}c@{}}SRM\\~\cite{luo2021generalizing}\end{tabular}} & \begin{tabular}[c]{@{}c@{}}UCF\\~\cite{yan2023ucf}\end{tabular}    & \begin{tabular}[c]{@{}c@{}}UnivFD\\~\cite{ojha2023towards}\end{tabular}                                 & \multicolumn{1}{c|}{\begin{tabular}[c]{@{}c@{}}CORE\\~\cite{ni2022core}\end{tabular}} & \begin{tabular}[c]{@{}c@{}}DAW-FDD\\~\cite{ju2024improving}\end{tabular} &\begin{tabular}[c]{@{}c@{}}DAG-FDD\\~\cite{ju2024improving}\end{tabular}                                & \begin{tabular}[c]{@{}c@{}}PG-FDD\\~\cite{lin2024preserving}\end{tabular}                                  \\ \hline \hline
\multirow{16}{*}{Fairness(\%)} & \multirow{4}{*}{Skin Tone}    & $F_{MEO}$                    & 11.678                                                    & 10.565                                                        & \multicolumn{1}{c|}{8.595}                                                & 12.629                                                 & 11.790                                                & \multicolumn{1}{c|}{17.068}                                               & 9.661                                                & 5.615                                                   & \multicolumn{1}{c|}{11.138}                                                & 11.726                                               & 8.680                                                & 6.435                                                   \\
                               &                               & $F_{DP}$                     & 14.133                                                    & 12.859                                                        & \multicolumn{1}{c|}{10.579}                                               & 15.157                                                 & 13.983                                                & \multicolumn{1}{c|}{16.962}                                               & 14.081                                               & 8.438                                                   & \multicolumn{1}{c|}{14.161}                                                & 14.724                                               & 13.388                                               & 13.256                                                  \\
                               &                               & $F_{OAE}$                    & 4.539                                                     & 3.671                                                         & \multicolumn{1}{c|}{4.407}                                                & 4.104                                                  & 4.894                                                 & \multicolumn{1}{c|}{5.036}                                                & 3.931                                                & 5.461                                                   & \multicolumn{1}{c|}{4.389}                                                 & 3.379                                                & 3.006                                                & 2.232                                                   \\
                               &                               & $F_{EO}$                     & 0.128                                                     & 0.151                                                         & \multicolumn{1}{c|}{0.102}                                                & 0.148                                                  & 0.127                                                 & \multicolumn{1}{c|}{0.200}                                                & 0.107                                                & 0.081                                                   & \multicolumn{1}{c|}{0.121}                                                 & 0.141                                                & 0.103                                                & 0.081                                                   \\ \cline{2-15} 
                               & \multirow{4}{*}{Gender}       & $F_{MEO}$                    & 6.942                                                     & 2.295                                                         & \multicolumn{1}{c|}{10.586}                                               & 7.818                                                  & 7.327                                                 & \multicolumn{1}{c|}{8.990}                                                & 6.054                                                & 6.525                                                   & \multicolumn{1}{c|}{7.518}                                                 & 6.259                                                & 5.934                                                & 4.944                                                   \\
                               &                               & $F_{DP}$                     & 4.378                                                     & 0.094                                                         & \multicolumn{1}{c|}{6.203}                                                & 4.881                                                  & 4.632                                                 & \multicolumn{1}{c|}{5.572}                                                & 4.086                                                & 4.136                                                   & \multicolumn{1}{c|}{4.867}                                                 & 4.093                                                & 3.917                                                & 3.565                                                   \\
                               &                               & $F_{OAE}$                    & 3.225                                                     & 1.508                                                         & \multicolumn{1}{c|}{4.799}                                                & 3.669                                                  & 3.303                                                 & \multicolumn{1}{c|}{4.060}                                                & 2.906                                                & 2.842                                                   & \multicolumn{1}{c|}{3.445}                                                 & 2.890                                                & 2.795                                                & 2.407                                                   \\
                               &                               & $F_{EO}$                     & 0.072                                                     & 0.023                                                         & \multicolumn{1}{c|}{0.106}                                                & 0.080                                                  & 0.074                                                 & \multicolumn{1}{c|}{0.091}                                                & 0.063                                                & 0.067                                                   & \multicolumn{1}{c|}{0.075}                                                 & 0.065                                                & 0.062                                                & 0.052                                                   \\ \cline{2-15} 
                               & \multirow{4}{*}{Age}          & $F_{MEO}$                    & 36.384                                                    & 13.175                                                        & \multicolumn{1}{c|}{25.574}                                               & 35.508                                                 & 30.942                                                & \multicolumn{1}{c|}{36.806}                                               & 32.134                                               & 6.629                                                   & \multicolumn{1}{c|}{34.717}                                                & 32.860                                               & 29.474                                               & 28.923                                                  \\
                               &                               & $F_{DP}$                     & 18.815                                                    & 19.006                                                        & \multicolumn{1}{c|}{16.522}                                               & 18.393                                                 & 17.007                                                & \multicolumn{1}{c|}{18.947}                                               & 17.086                                               & 12.524                                                  & \multicolumn{1}{c|}{18.454}                                                & 16.798                                               & 15.004                                               & 19.634                                                  \\
                               &                               & $F_{OAE}$                    & 19.144                                                    & 2.128                                                         & \multicolumn{1}{c|}{14.899}                                               & 18.668                                                 & 16.331                                                & \multicolumn{1}{c|}{19.294}                                               & 16.714                                               & 6.598                                                   & \multicolumn{1}{c|}{18.249}                                                & 17.426                                               & 15.605                                               & 15.099                                                  \\
                               &                               & $F_{EO}$                     & 0.524                                                     & 0.193                                                         & \multicolumn{1}{c|}{0.373}                                                & 0.525                                                  & 0.442                                                 & \multicolumn{1}{c|}{0.553}                                                & 0.490                                                & 0.147                                                   & \multicolumn{1}{c|}{0.507}                                                 & 0.482                                                & 0.447                                                & 0.420                                                   \\ \cline{2-15} 
                               & \multirow{4}{*}{Intersection} & $F_{MEO}$                    & 16.037                                                    & 18.196                                                        & \multicolumn{1}{c|}{19.081}                                               & 16.394                                                 & 19.895                                                & \multicolumn{1}{c|}{21.535}                                               & 13.589                                               & 12.135                                                  & \multicolumn{1}{c|}{15.921}                                                & 14.772                                               & 12.424                                               & 10.340                                                  \\
                               &                               & $F_{DP}$                     & 16.749                                                    & 17.705                                                        & \multicolumn{1}{c|}{16.207}                                               & 17.525                                                 & 18.313                                                & \multicolumn{1}{c|}{19.312}                                               & 17.029                                               & 12.639                                                  & \multicolumn{1}{c|}{17.144}                                                & 16.813                                               & 15.850                                               & 16.846                                                  \\
                               &                               & $F_{OAE}$                    & 7.914                                                     & 4.565                                                         & \multicolumn{1}{c|}{12.301}                                               & 6.936                                                  & 8.704                                                 & \multicolumn{1}{c|}{9.372}                                                & 5.877                                                & 11.469                                                  & \multicolumn{1}{c|}{7.025}                                                 & 5.523                                                & 6.909                                                & 5.109                                                   \\
                               &                               & $F_{EO}$                     & 0.381                                                     & 0.346                                                         & \multicolumn{1}{c|}{0.400}                                                & 0.399                                                  & 0.394                                                 & \multicolumn{1}{c|}{0.467}                                                & 0.317                                                & 0.249                                                   & \multicolumn{1}{c|}{0.364}                                                 & 0.336                                                & 0.313                                                & 0.243                                                   \\ \hline
\multirow{5}{*}{Utility(\%)}   & \multirow{5}{*}{-}            & AUC                     & 0.958                                                     & 0.967                                                         & \multicolumn{1}{c|}{0.975}                                                & 0.966                                                  & 0.964                                                 & \multicolumn{1}{c|}{0.951}                                                & 0.976                                                & 0.978                                                   & \multicolumn{1}{c|}{0.969}                                                 & 0.962                                                & 0.967                                                & 0.983                                                   \\
                               &                               & ACC                     & 0.864                                                     & 0.864                                                         & \multicolumn{1}{c|}{0.823}                                                & 0.876                                                  & 0.855                                                 & \multicolumn{1}{c|}{0.862}                                                & 0.909                                                & 0.829                                                   & \multicolumn{1}{c|}{0.888}                                                 & 0.874                                                & 0.882                                                & 0.925                                                   \\
                               &                               & AP                      & 0.934                                                     & 0.972                                                         & \multicolumn{1}{c|}{0.976}                                                & 0.948                                                  & 0.946                                                 & \multicolumn{1}{c|}{0.938}                                                & 0.964                                                & 0.979                                                   & \multicolumn{1}{c|}{0.952}                                                 & 0.942                                                & 0.951                                                & 0.977                                                   \\
                               &                               & EER                     & 0.087                                                     & 0.098                                                         & \multicolumn{1}{c|}{0.091}                                                & 0.082                                                  & 0.081                                                 & \multicolumn{1}{c|}{0.116}                                                & 0.064                                                & 0.085                                                   & \multicolumn{1}{c|}{0.073}                                                 & 0.084                                                & 0.079                                                & 0.056                                                   \\
                               &                               & FPR                     & 0.267                                                     & 0.012                                                         & \multicolumn{1}{c|}{0.349}                                                & 0.242                                                  & 0.285                                                 & \multicolumn{1}{c|}{0.272}                                                & 0.172                                                & 0.339                                                   & \multicolumn{1}{c|}{0.221}                                                 & 0.246                                                & 0.227                                                & 0.142                                                   \\ \hline
\end{tabular}
}
\caption{\small \textit{Detailed fairness and utility evaluation results on a training subset with the ratio of real vs fake is 1:10.}}
\label{tab:ratio1_10}
\end{table*}

\begin{table*}[t]
\centering
\scalebox{0.675}{
\begin{tabular}{c|c|c|cccccccccccc}
\hline
\multirow{3}{*}{Measure}       & \multirow{3}{*}{Attribute}    & \multirow{3}{*}{Metric} & \multicolumn{12}{c}{Model Type}                                                                                                                                                                                                                                                                                                                                                                                                                                                                                                                                                                                                                                                                                                                                          \\ \cline{4-15} 
                               &                               &                         & \multicolumn{3}{c|}{Native}                                                                                                                                                                           & \multicolumn{3}{c|}{Frequency}                                                                                                                                                             & \multicolumn{3}{c|}{Spatial}                                                                                                                                                                & \multicolumn{3}{c}{Fairness-enhanced}                                                                                                                                 \\ \cline{4-15} 
                               &                               &                         & \begin{tabular}[c]{@{}c@{}}Xception\\ ~\cite{chollet2017xception}\end{tabular}                               & \begin{tabular}[c]{@{}c@{}}EfficientB4\\~\cite{tan2019efficientnet}\end{tabular}                              & \multicolumn{1}{c|}{\begin{tabular}[c]{@{}c@{}}ViT-B/16\\~\cite{dosovitskiy2020image}\end{tabular}} & \begin{tabular}[c]{@{}c@{}}F3Net\\~\cite{qian2020thinking}\end{tabular}  & \begin{tabular}[c]{@{}c@{}}SPSL\\~\cite{liu2021spatial}\end{tabular}                                    & \multicolumn{1}{c|}{\begin{tabular}[c]{@{}c@{}}SRM\\~\cite{luo2021generalizing}\end{tabular}} & \begin{tabular}[c]{@{}c@{}}UCF\\~\cite{yan2023ucf}\end{tabular}    & \begin{tabular}[c]{@{}c@{}}UnivFD\\~\cite{ojha2023towards}\end{tabular}                                 & \multicolumn{1}{c|}{\begin{tabular}[c]{@{}c@{}}CORE\\~\cite{ni2022core}\end{tabular}} & \begin{tabular}[c]{@{}c@{}}DAW-FDD\\~\cite{ju2024improving}\end{tabular} &\begin{tabular}[c]{@{}c@{}}DAG-FDD\\~\cite{ju2024improving}\end{tabular}                                & \begin{tabular}[c]{@{}c@{}}PG-FDD\\~\cite{lin2024preserving}\end{tabular}                                  \\ \hline \hline
\multirow{16}{*}{Fairness(\%)} & \multirow{4}{*}{Skin Tone}    & $F_{MEO}$                    & 6.557                                                     & 11.357                                                        & \multicolumn{1}{c|}{7.743}                                                & 7.364                                                  & 5.850                                                 & \multicolumn{1}{c|}{18.290}                                               & 5.814                                                & 12.493                                                  & \multicolumn{1}{c|}{6.397}                                                 & 5.083                                                & 6.515                                                & 5.122                                                   \\
                               &                               & $F_{DP}$                     & 13.087                                                    & 13.112                                                        & \multicolumn{1}{c|}{12.348}                                               & 15.751                                                 & 11.728                                                & \multicolumn{1}{c|}{22.766}                                               & 14.881                                               & 13.771                                                  & \multicolumn{1}{c|}{14.364}                                                & 12.265                                               & 15.536                                               & 12.718                                                  \\
                               &                               & $F_{OAE}$                    & 1.779                                                     & 2.939                                                         & \multicolumn{1}{c|}{1.589}                                                & 1.352                                                  & 2.402                                                 & \multicolumn{1}{c|}{3.823}                                                & 1.495                                                & 2.386                                                   & \multicolumn{1}{c|}{0.996}                                                 & 1.401                                                & 1.471                                                & 1.742                                                   \\
                               &                               & $F_{EO}$                     & 0.108                                                     & 0.140                                                         & \multicolumn{1}{c|}{0.093}                                                & 0.153                                                  & 0.090                                                 & \multicolumn{1}{c|}{0.331}                                                & 0.135                                                & 0.155                                                   & \multicolumn{1}{c|}{0.125}                                                 & 0.084                                                & 0.143                                                & 0.096                                                   \\ \cline{2-15} 
                               & \multirow{4}{*}{Gender}       & $F_{MEO}$                    & 1.808                                                     & 1.803                                                         & \multicolumn{1}{c|}{2.106}                                                & 1.659                                                  & 2.316                                                 & \multicolumn{1}{c|}{3.035}                                                & 3.035                                                & 1.997                                                   & \multicolumn{1}{c|}{3.499}                                                 & 1.920                                                & 1.812                                                & 3.016                                                   \\
                               &                               & $F_{DP}$                     & 1.769                                                     & 0.136                                                         & \multicolumn{1}{c|}{0.369}                                                & 1.321                                                  & 0.470                                                 & \multicolumn{1}{c|}{2.871}                                                & 2.503                                                & 2.008                                                   & \multicolumn{1}{c|}{2.119}                                                 & 1.097                                                & 1.405                                                & 2.633                                                   \\
                               &                               & $F_{OAE}$                    & 1.385                                                     & 1.269                                                         & \multicolumn{1}{c|}{1.615}                                                & 1.661                                                  & 1.777                                                 & \multicolumn{1}{c|}{1.270}                                                & 1.817                                                & 0.579                                                   & \multicolumn{1}{c|}{2.662}                                                 & 1.928                                                & 1.734                                                & 1.677                                                   \\
                               &                               & $F_{EO}$                     & 0.025                                                     & 0.018                                                         & \multicolumn{1}{c|}{0.027}                                                & 0.032                                                  & 0.032                                                 & \multicolumn{1}{c|}{0.038}                                                & 0.035                                                & 0.020                                                   & \multicolumn{1}{c|}{0.052}                                                 & 0.036                                                & 0.033                                                & 0.033                                                   \\ \cline{2-15} 
                               & \multirow{4}{*}{Age}          & $F_{MEO}$                    & 8.571                                                     & 11.954                                                        & \multicolumn{1}{c|}{9.832}                                                & 8.680                                                  & 9.523                                                 & \multicolumn{1}{c|}{36.509}                                               & 11.812                                               & 10.258                                                  & \multicolumn{1}{c|}{13.180}                                                & 8.539                                                & 9.553                                                & 11.109                                                  \\
                               &                               & $F_{DP}$                     & 17.740                                                    & 18.338                                                        & \multicolumn{1}{c|}{18.662}                                               & 17.788                                                 & 18.910                                                & \multicolumn{1}{c|}{27.851}                                               & 16.998                                               & 16.446                                                  & \multicolumn{1}{c|}{16.719}                                                & 18.069                                               & 18.034                                               & 16.696                                                  \\
                               &                               & $F_{OAE}$                    & 3.191                                                     & 0.656                                                         & \multicolumn{1}{c|}{1.405}                                                & 4.291                                                  & 2.010                                                 & \multicolumn{1}{c|}{12.139}                                               & 5.174                                                & 2.737                                                   & \multicolumn{1}{c|}{5.723}                                                 & 3.924                                                & 4.283                                                & 4.859                                                   \\
                               &                               & $F_{EO}$                     & 0.196                                                     & 0.157                                                         & \multicolumn{1}{c|}{0.141}                                                & 0.228                                                  & 0.174                                                 & \multicolumn{1}{c|}{0.685}                                                & 0.281                                                & 0.173                                                   & \multicolumn{1}{c|}{0.288}                                                 & 0.209                                                & 0.240                                                & 0.246                                                   \\ \cline{2-15} 
                               & \multirow{4}{*}{Intersection} & $F_{MEO}$                    & 11.785                                                    & 18.146                                                        & \multicolumn{1}{c|}{14.840}                                               & 14.116                                                 & 12.940                                                & \multicolumn{1}{c|}{23.975}                                               & 11.313                                               & 16.240                                                  & \multicolumn{1}{c|}{12.729}                                                & 10.948                                               & 13.399                                               & 9.441                                                   \\
                               &                               & $F_{DP}$                     & 20.192                                                    & 17.674                                                        & \multicolumn{1}{c|}{19.309}                                               & 23.135                                                 & 18.594                                                & \multicolumn{1}{c|}{29.625}                                               & 22.394                                               & 19.583                                                  & \multicolumn{1}{c|}{21.935}                                                & 18.674                                               & 23.584                                               & 17.906                                                  \\
                               &                               & $F_{OAE}$                    & 3.419                                                     & 4.233                                                         & \multicolumn{1}{c|}{3.638}                                                & 4.263                                                  & 4.020                                                 & \multicolumn{1}{c|}{5.166}                                                & 3.479                                                & 3.133                                                   & \multicolumn{1}{c|}{4.945}                                                 & 3.388                                                & 3.885                                                & 3.667                                                   \\
                               &                               & $F_{EO}$                     & 0.251                                                     & 0.312                                                         & \multicolumn{1}{c|}{0.256}                                                & 0.329                                                  & 0.232                                                 & \multicolumn{1}{c|}{0.678}                                                & 0.305                                                & 0.307                                                   & \multicolumn{1}{c|}{0.285}                                                 & 0.225                                                & 0.314                                                & 0.232                                                   \\ \hline
\multirow{5}{*}{Utility(\%)}   & \multirow{5}{*}{-}            & AUC                     & 0.978                                                     & 0.973                                                         & \multicolumn{1}{c|}{0.982}                                                & 0.979                                                  & 0.982                                                 & \multicolumn{1}{c|}{0.933}                                                & 0.978                                                & 0.975                                                   & \multicolumn{1}{c|}{0.979}                                                 & 0.980                                                & 0.982                                                & 0.983                                                   \\
                               &                               & ACC                     & 0.920                                                     & 0.862                                                         & \multicolumn{1}{c|}{0.895}                                                & 0.928                                                  & 0.916                                                 & \multicolumn{1}{c|}{0.832}                                                & 0.921                                                & 0.849                                                   & \multicolumn{1}{c|}{0.921}                                                 & 0.920                                                & 0.930                                                & 0.933                                                   \\
                               &                               & AP                      & 0.978                                                     & 0.977                                                         & \multicolumn{1}{c|}{0.984}                                                & 0.978                                                  & 0.984                                                 & \multicolumn{1}{c|}{0.915}                                                & 0.979                                                & 0.978                                                   & \multicolumn{1}{c|}{0.978}                                                 & 0.979                                                & 0.981                                                & 0.984                                                   \\
                               &                               & EER                     & 0.070                                                     & 0.088                                                         & \multicolumn{1}{c|}{0.075}                                                & 0.066                                                  & 0.064                                                 & \multicolumn{1}{c|}{0.141}                                                & 0.076                                                & 0.087                                                   & \multicolumn{1}{c|}{0.074}                                                 & 0.070                                                & 0.065                                                & 0.066                                                   \\
                               &                               & FPR                     & 0.034                                                     & 0.008                                                         & \multicolumn{1}{c|}{0.009}                                                & 0.042                                                  & 0.018                                                 & \multicolumn{1}{c|}{0.116}                                                & 0.054                                                & 0.004                                                   & \multicolumn{1}{c|}{0.055}                                                 & 0.037                                                & 0.040                                                & 0.051                                                   \\ \hline
\end{tabular}
}
\caption{\small \textit{Detailed fairness and utility evaluation results on a training subset with the ratio of real vs fake is 10:1.}}
\label{tab:ratio10_1}
\end{table*}

\subsection{Performance on Different Age Subgroups}
We conduct an analysis of all detectors on age subgroups.
\textbf{1)} As shown in Fig.~\ref{fig:age_fpr}, 
facial images with an age range of 0-14 (Child) are more often misclassified as fake,  likely due to the underrepresentation of children in our dataset (see Fig. \ref{fig:dataset_stas} (b)). This suggests detectors tend to show higher error rates for minority groups and show higher accuracy for the majority (Adult).
\textbf{2)} Among those detectors, EfficientB4, UnivFD, and PG-FDD demonstrate a smaller FPR gap between age subgroups, indicating these models may be less susceptible to age bias.

\begin{figure*}[t]
    \centering
    \includegraphics[width=1\textwidth]{Figs/age_fpr.pdf}
    \vspace{-4mm}
    \caption{\small \textit{FPR(\%) of each age subgroup. The subgroup with the highest FPR score is highlighted in red, while the subgroup with the lowest FPR score is shown in green.}}
    \label{fig:age_fpr}
\end{figure*}

\subsection{Details of Post-Processing} \label{appendix:post_processing_details}
\begin{figure*}[t]
    \centering
    \includegraphics[width=0.8\textwidth]{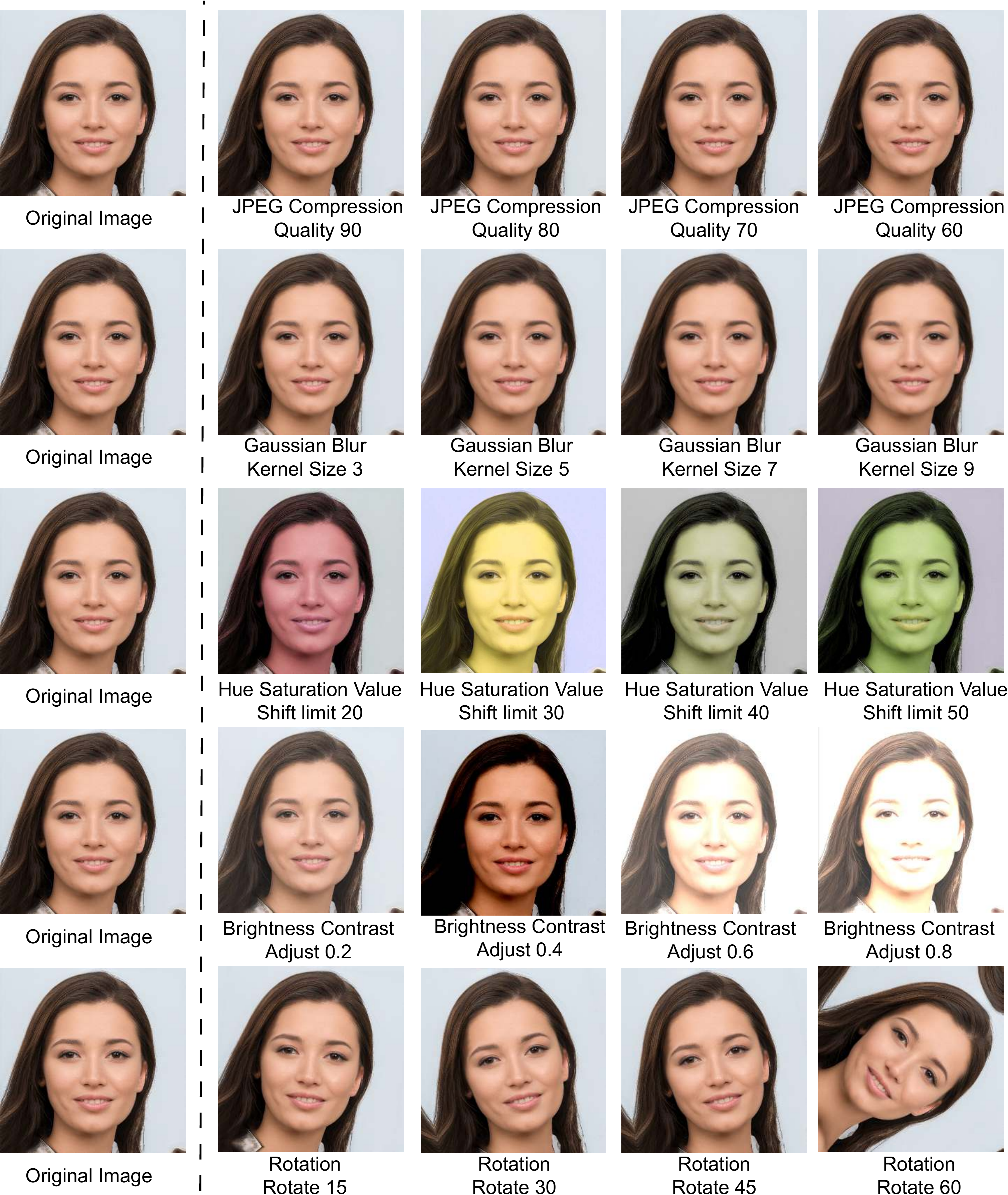}
    \vspace{-2mm}
    \caption{\textit{Visualization of the image after different post-processing.}}
    \label{fig:images_applied_post_processing}
    \vspace{-4mm}
\end{figure*}

In Section~\ref{sec:fairness_benchmark} we have applied 6 post-processing methods to evaluate detectors' robustness. Fig.~\ref{fig:images_applied_post_processing} visualizes the image after being applied different post-processing methods.
We describe each post-processing method as follows:

\textbf{JPEG Compression}: Image compression introduces compression artifacts and reduces the image quality, simulating real-world scenarios where images may be of lower quality or have compression artifacts. In Fig.~\ref{fig:robustness} we apply image compression with quality 80 to each image in the test set.

\textbf{Gaussian Blur}: This post-processing reduces image detail and noise by smoothing it through averaging pixel values with a Gaussian kernel. In Fig.~\ref{fig:robustness} we apply gaussian blur with kernel size 7 to each image in the test set.

\textbf{Hue Saturation Value}: Alters the hue, saturation, and value of the image within specified limits. This post-processing technique is used to simulate variations in color and lighting conditions. Adjusting the hue changes the overall color tone, saturation controls the intensity of colors, and value adjusts the brightness. The results in Fig.~\ref{fig:robustness} are after we adjust hue, saturation, and value with shifting limits 30.

\textbf{Random Brightness and Contrast}: This post-processing method adjusts the brightness and contrast of the image within specified limits. By applying random brightness and contrast variations, it introduces changes in the illumination and contrast levels of the images. This evaluates detector's robustness to different illumination conditions. The results in Fig.~\ref{fig:robustness} are after we adjust brightness and contrast with shifting limits 0.4.

\textbf{Random Crop}: Resizes the image to a specified size and then randomly crops a portion of it to the target dimensions. This post-processing method is used to evaluate the detector's robustness to variations in the spatial content of the image. The results in Fig.~\ref{fig:robustness} are after we randomly crop the image with target dimension of $244 \times 244$.

\textbf{Rotation}: Rotates the image within a specified angle limit. This post-processing method is used to evaluate the detector's robustness to changes in the orientation of objects within the image. The results in Fig.~\ref{fig:robustness} are after we randomly rotate the image within a range of -30 to 30 degrees. 

\begin{figure*}[t]
    \centering
    \includegraphics[width=1\textwidth]{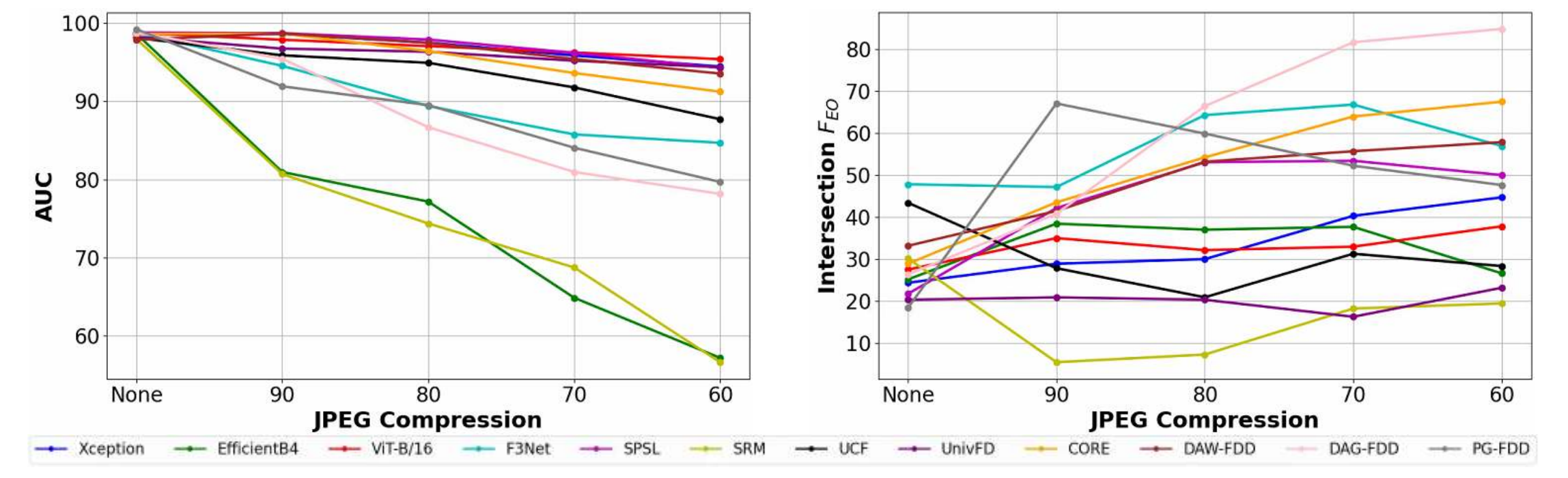}
    \vspace{-6mm}
    \caption{\textit{Robustness analysis in terms of utility and fairness under varying degrees of JPEG compression.}}
    \label{fig:JPEG_compression}
    \vspace{1mm}
\end{figure*}
\begin{figure*}[ht]
    \centering
    \includegraphics[width=1\textwidth]{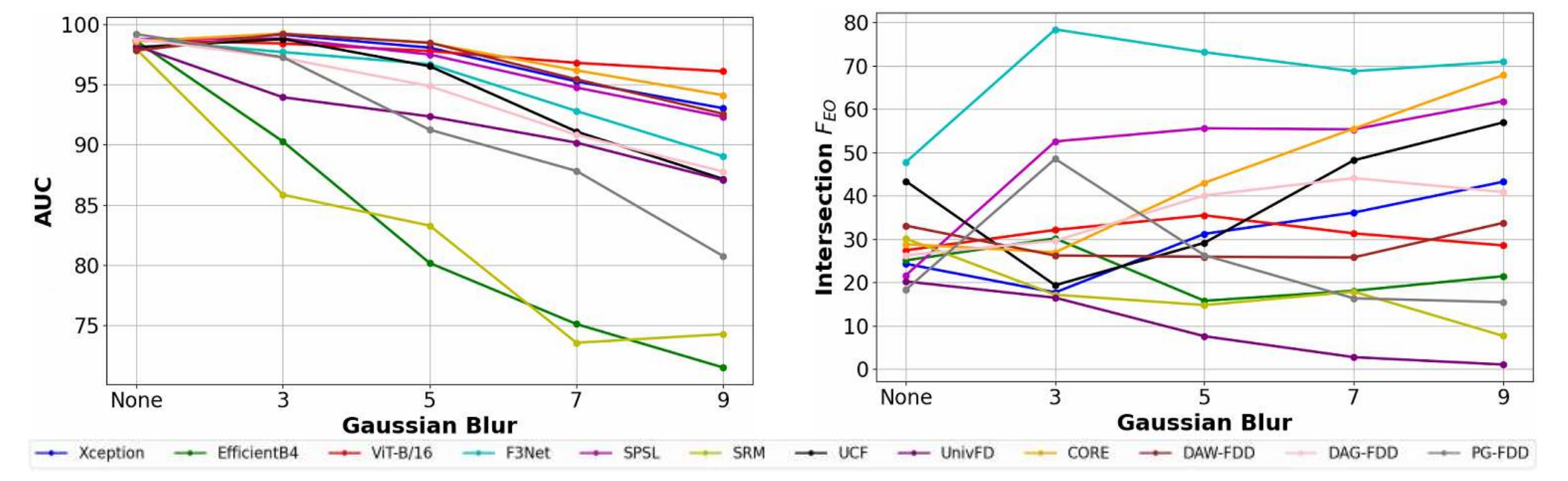}
    \vspace{-6mm}
    \caption{\textit{Robustness analysis in terms of utility and fairness under varying kernel sizes of Gaussian Blur.}}
    \label{fig:Gaussian_blur}
    \vspace{1mm}
\end{figure*}
\begin{figure*}[ht]
    \centering
    \includegraphics[width=1\textwidth]{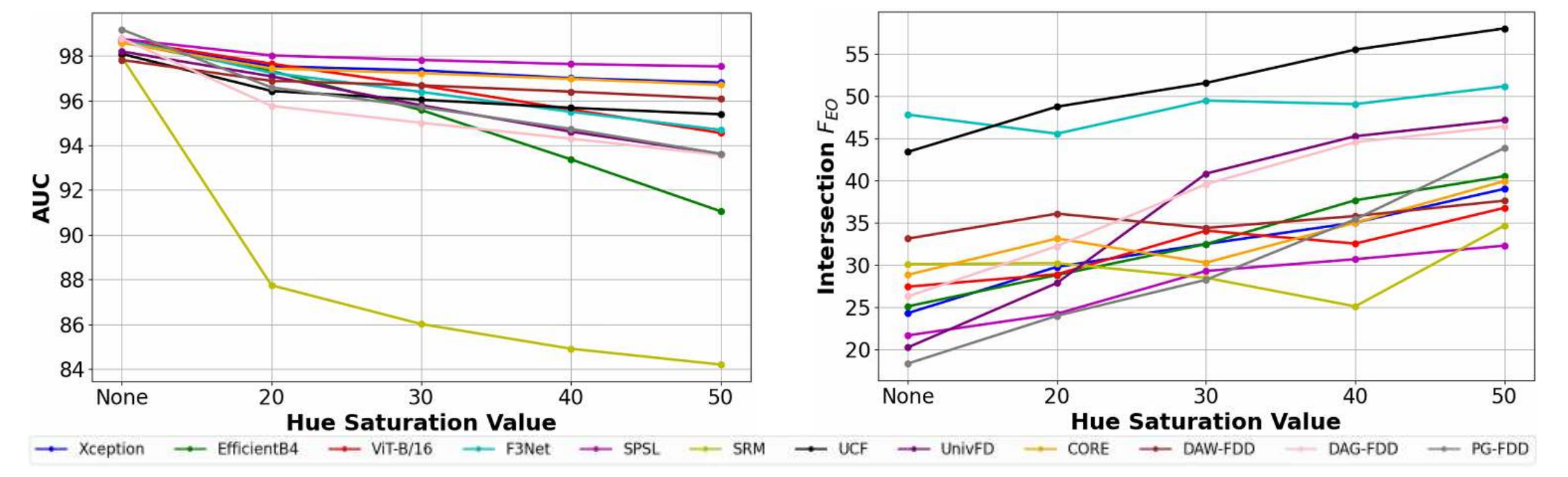}
    \vspace{-6mm}
    \caption{\textit{Robustness analysis in terms of utility and fairness under varying degrees of Hue Saturation Value.}}
    \label{fig:Hue_saturation}
    \vspace{1mm}
\end{figure*}
\begin{figure*}[ht]
    \centering
    \includegraphics[width=1\textwidth]{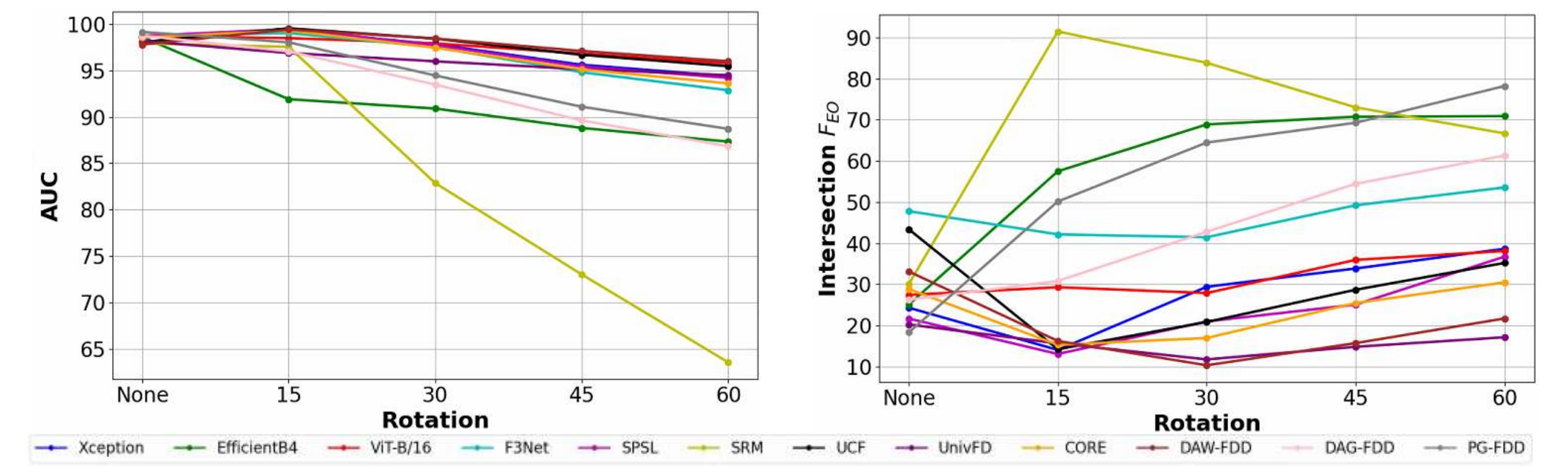}
    \vspace{-6mm}
    \caption{\textit{Robustness analysis in terms of utility and fairness under varying degrees of Rotations.}}
    \label{fig:Rotation}
    \vspace{1mm}
\end{figure*}
\begin{figure*}[ht]
    \centering
    \includegraphics[width=1\textwidth]{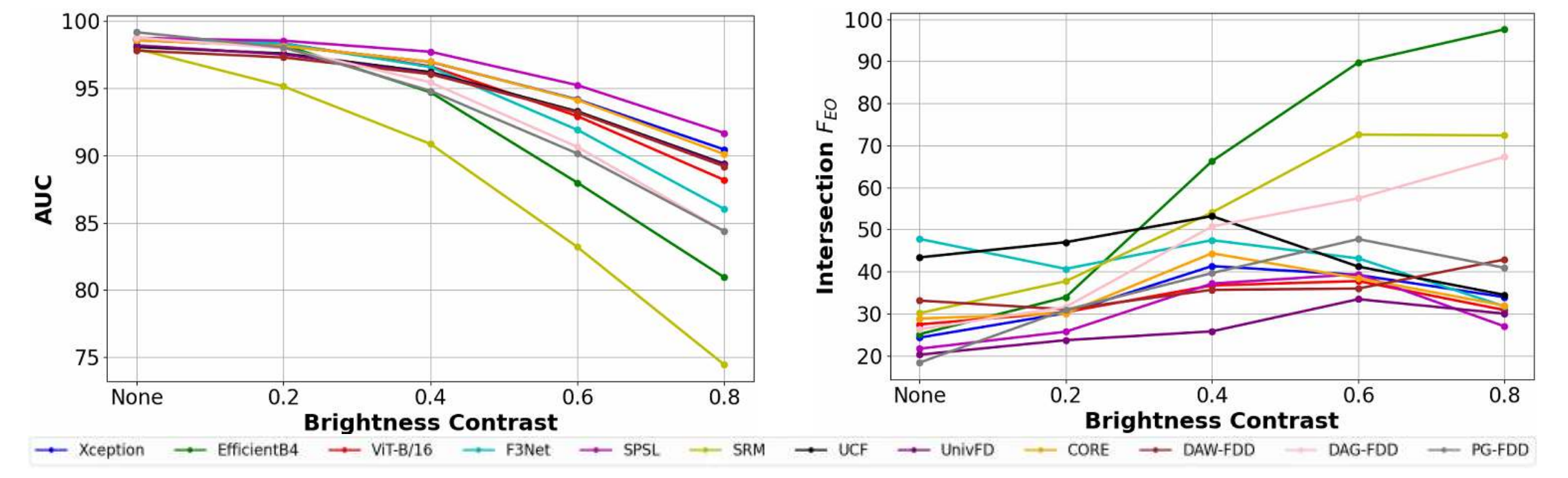}
    \vspace{-6mm}
    \caption{\textit{Robustness analysis in terms of utility and fairness under varying degrees of Brightness Contrast.}}
    \label{fig:Brightness_contrast}
    \vspace{1mm}
\end{figure*}

\subsection{Additional Fairness Robustness Evaluation Results} \label{appendix:additional_robustness_results}
Fig.~\ref{fig:JPEG_compression} to Fig.~\ref{fig:Brightness_contrast} demonstrate detectors' robustness analysis in more detail as a function of different degrees of post-processing. Overall, ViT-B/16~\cite{dosovitskiy2020image} and UnivFD~\cite{ojha2023towards} show stronger robustness to various post-processing methods compared to other detection methods. Fairness-enhanced detectors do not have robustness against post-processing; this would be a direction for future studies to work on.
Figure~\ref{fig:JPEG_compression} presents a detailed robustness analysis in terms of utility and fairness under varying degrees of JPEG compression. The utility of all detectors decreases as image quality is reduced. Among the detectors, ViT-B/16~\cite{dosovitskiy2020image} exhibits the highest utility robustness, ViT-B/16~\cite{dosovitskiy2020image} and UnivFD~\cite{ojha2023towards} both demonstrate the strongest fairness robustness.
When considering Gaussian blur, ViT-B/16 again stands out as the most robust detector in terms of utility, whereas DAW-FDD~\cite{ju2024improving} and UnivFD~\cite{ojha2023towards} show the great robustness in terms of fairness. Against Hue Saturation Value adjustments, SPSL~\cite{liu2021spatial} shows the strongest utility robustness, while the fairness of DAW-FDD~\cite{ju2024improving} fluctuates less with different Hue Saturation Value adjustments.
ViT-B/16 demonstrates superior robustness in both utility and fairness when facing rotations. For brightness contrast variations, SPSL~\cite{liu2021spatial} is the most robust detector in terms of utility, while UnivFD once again shows superior robustness in terms of fairness. Last, we can get the same conclusion from Fig.~\ref{fig:JPEG_compression} to Fig.~\ref{fig:Brightness_contrast} as in the main manuscript, that post-processing clearly impairs detectors' utility but does not necessarily make detectors more biased.

\subsection{Full Results of Effect of Increasing the Size of Train Set} 
\label{appendix:full_results_of_train_set_size}
In this section, we provide the full evaluation results tested under different sizes of train set, as shown from Table~\ref{tab:20_trainset} to Table~\ref{tab:80_trainset}. Intersection $F_{EO}$ and AUC align with the results in Fig.~\ref{fig:datasize_realfake_ratio} of the submitted manuscript.

\subsection{Full Results of Effect of the Ratio of Real and Fake} 
\label{appendix:full_results_of_real_fake_ratio}
In this section, we provide the full evaluation results tested under the train set with different ratios of real and fake, as shown from Table~\ref{tab:ratio1_1} to Table~\ref{tab:ratio10_1}. Intersection $F_{EO}$ and AUC align with the results in Fig.~\ref{fig:datasize_realfake_ratio} of the submitted manuscript.

\subsection{Comparison Results with Foundation Model} \label{appendix:clip_mlp_results}
In the Discussion (Section~\ref{subsec:Discussion}) of the main manuscript, we highlighted the potential of integrating foundation models (\eg, CLIP) into detector design as a strategy for mitigating bias. To explore this, we conducted a preliminary experiment by designing a detector using a frozen CLIP model combined with a trainable 3-layer MLP. This model was trained and tested on the AI-Face dataset. For comparison, we selected one representative detector from each model type: EfficientB4~\cite{tan2019efficientnet}, SPSL~\cite{liu2021spatial}, UnivFD~\cite{ojha2023towards}, and PG-FDD~\cite{lin2024preserving}. These four detectors' results are consistent with those reported in Table~\ref{tab:intra_domain_overall}.
As shown in Table~\ref{tab:clip_mlp_fairness}, the CLIP+MLP detector demonstrates a clear advantage in both fairness and utility metrics, suggesting that foundation models hold significant promise for bias mitigation. For instance, its $F_{EO}$ score is 3.11\% lower than the second-best method, PG-FDD, for the Skin Tone group, and 14.046\% lower for the Intersection group. 

\begin{table*}[t]
    \centering
    \scalebox{0.57}{
\begin{tabular}{c|cccccccccccccccc|ccccc}
\hline
                         & \multicolumn{16}{c|}{Fairness}                                                                                                                                                                                                                                                                                                                                                                                                                                                                                                                                                                                                                                                                                                                & \multicolumn{5}{c}{Utiliy}                                                                                                                                                                                    \\ \cline{2-22} 
                         & \multicolumn{4}{c|}{Skin Tone}                                                                                                                                                         & \multicolumn{4}{c|}{Gender}                                                                                                                                                            & \multicolumn{4}{c|}{Age}                                                                                                                                                                & \multicolumn{4}{c|}{Intersection}                                                                                                                                 & \multicolumn{5}{c}{-}                                                                                                                                                                                         \\ \cline{2-22} 
\multirow{-3}{*}{Method} & $F_{MEO}$                                   & $F_{DP}$                                    & $F_{OAE}$                                   & \multicolumn{1}{c|}{$F_{EO}$}                                    & $F_{MEO}$                                   & $F_{DP}$                                    & $F_{OAE}$                                   & \multicolumn{1}{c|}{$F_{EO}$}                                    & $F_{MEO}$                                   & $F_{DP}$                                     & $F_{OAE}$                                   & \multicolumn{1}{c|}{$F_{EO}$}                                    & $F_{MEO}$                                   & $F_{DP}$                                    & $F_{OAE}$                                   & $F_{EO}$                                    & AUC                                     & ACC                                     & AP                                      & EER                                    & FPR                                    \\ \hline
EfficientB4              & 5.385                                  & 1.725                                  & 1.487                                  & \multicolumn{1}{c|}{5.863}                                  & 8.300                                  & \cellcolor[HTML]{E4E0E1}\textbf{6.184} & 4.377                                  & \multicolumn{1}{c|}{11.062}                                 & 6.796                                  & 11.849                                  & 2.856                                  & \multicolumn{1}{c|}{10.300}                                 & 17.586                                 & \cellcolor[HTML]{E4E0E1}\textbf{8.607} & 8.461                                  & 25.114                                 & 98.611                                  & 94.203                                  & 99.542                                  & 6.689                                  & 20.066                                 \\
SPSL                     & 4.411                                  & 1.827                                  & 1.037                                  & \multicolumn{1}{c|}{4.534}                                  & 8.055                                  & 9.379                                  & 1.135                                  & \multicolumn{1}{c|}{9.789}                                  & 27.614                                 & 11.232                                  & 7.270                                  & \multicolumn{1}{c|}{40.943}                                 & 10.379                                 & 13.259                                 & 2.464                                  & 21.679                                 & 98.747                                  & 96.346                                  & 99.356                                  & 4.371                                  & 13.661                                 \\
UnivFD                   & 4.503                                  & 1.19                                   & 1.622                                  & \multicolumn{1}{c|}{5.408}                                  & 2.577                                  & 8.556                                  & 2.748                                  & \multicolumn{1}{c|}{5.536}                                  & 5.436                                  & 15.249                                  & 3.793                                  & \multicolumn{1}{c|}{14.148}                                 & 6.119                                  & 14.026                                 & 6.287                                  & 20.255                                 & 98.192                                  & 93.651                                  & 99.400                                  & 7.633                                  & 18.550                                 \\
PG-FDD                   & 3.190                                  & 1.252                                  & 1.071                                  & \multicolumn{1}{c|}{3.702}                                  & 6.465                                  & 9.746                                  & 0.882                                  & \multicolumn{1}{c|}{9.115}                                  & 14.804                                 & \cellcolor[HTML]{E4E0E1}\textbf{10.467} & 5.009                                  & \multicolumn{1}{c|}{29.585}                                 & 9.578                                  & 14.697                                 & 3.062                                  & 18.348                                 & 99.172                                  & 96.174                                  & 99.694                                  & 4.961                                  & 10.971                                 \\ \hline \hline
CLIP+MLP                 & \cellcolor[HTML]{E4E0E1}\textbf{0.419} & \cellcolor[HTML]{E4E0E1}\textbf{0.938} & \cellcolor[HTML]{E4E0E1}\textbf{0.227} & \multicolumn{1}{c|}{\cellcolor[HTML]{E4E0E1}\textbf{0.591}} & \cellcolor[HTML]{E4E0E1}\textbf{0.506} & 8.658                                  & \cellcolor[HTML]{E4E0E1}\textbf{0.334} & \multicolumn{1}{c|}{\cellcolor[HTML]{E4E0E1}\textbf{1.021}} & \cellcolor[HTML]{E4E0E1}\textbf{0.765} & 14.473                                  & \cellcolor[HTML]{E4E0E1}\textbf{0.395} & \multicolumn{1}{c|}{\cellcolor[HTML]{E4E0E1}\textbf{1.802}} & \cellcolor[HTML]{E4E0E1}\textbf{1.973} & 13.992                                 & \cellcolor[HTML]{E4E0E1}\textbf{1.000} & \cellcolor[HTML]{E4E0E1}\textbf{4.302} & \cellcolor[HTML]{E4E0E1}\textbf{99.973} & \cellcolor[HTML]{E4E0E1}\textbf{99.290} & \cellcolor[HTML]{E4E0E1}\textbf{99.991} & \cellcolor[HTML]{E4E0E1}\textbf{0.793} & \cellcolor[HTML]{E4E0E1}\textbf{1.171} \\ \hline
\end{tabular}
}
\vspace{-2mm}
\caption{\small \textit{Fairness and utility performance of CLIP+MLP compared to representative detectors on the AI-Face dataset, highlighting the potential of foundation models for bias mitigation.}}
\vspace{-6mm}
\label{tab:clip_mlp_fairness}
\end{table*}

\newpage
\clearpage
\section{Datasheet for AI-Face}\label{appendix:ai_face_datasheet}
In this section, we present a DataSheet~\cite{gebru2021datasheets} for AI-Face.
\subsection{Motivation For Dataset Creation}
\begin{compactitem}
    \item \textbf{Why is the dataset created?} For researchers to evaluate the fairness of AI face detection models or to train fairer models. Please see Section~\ref{sec:Background and Motivation} `Background and Motivation' in the submitted manuscript.
    \item \textbf{Has the dataset been used already?} Yes. Our fairness benchmark is based on this dataset.
    \item \textbf{What (other) tasks could the dataset be used for?} Could be used as training data for generative methods attribution task.
\end{compactitem}

\subsection{Data Composition}
\begin{compactitem}
    \item \textbf{What are the instances?} The instances that we consider in this work are real face images and AI-generated face images from public datasets.
    \item \textbf{How many instances are there?} We include 1,646,545 face images from public datasets. Please see Table~\ref{tab:dataset_info_detail} for details.
    \item \textbf{What data does each instance consist of?} Each instance consists of an image. 
    \item \textbf{Is there a label or target associated with each instance?} Each image is associated with gender annotation, age annotation, skin tone annotation, intersectional attribute (gender and skin tone) annotation, and target label (fake or real).
    \item \textbf{Is any information missing from individual instances?} No.
    \item \textbf{Are relationships between individual instances made explicit?} Not applicable -- we do not study the relationship between each image.
    \item \textbf{Does the dataset contain all possible instances or is it a sample?} Contains all instances our curation pipeline collected. Since the current dataset does not cover all available images online, there is a high probability more instances can be collected in the future.
    \item \textbf{Are there recommended data splits (\eg, training, development/validation, testing)?} For detector development and training, the dataset can be split as 6:2:2. 
    \item \textbf{Are there any errors, sources of noise, or redundancies in the dataset? If so, please provide a description.} Yes. Despite our extensive efforts to mitigate the bias that may introduced by the automated annotator and reduce demographic label noise, there may still be mislabeled instances. Given the dataset's size of over 1 million images and most are generated face images, it is impractical for humans to manually check and correct each image individually.
    \item \textbf{Is the dataset self-contained, or does it link to or otherwise rely on external resources (e.g., websites, tweets, other datasets)?} The dataset is self-contained.
\end{compactitem}

\subsection{Collection Process}
\begin{compactitem}
    \item \textbf{What mechanisms or procedures were used to collect the data?} We build our AI-Face dataset by collecting and integrating public AI-generated face images sourced from academic publications, GitHub repositories, and commercial tools. Please see `Data Collection' in Section~\ref{subsec:data_collection}
    \item \textbf{How was the data associated with each instance acquired? Was the data directly observable (\eg, raw text, movie ratings), reported by subjects (\eg, survey responses), or indirectly inferred/derived from other data?} The data can be acquired after our verification of user submitted and signed EULA.
    \item \textbf{If the dataset is a sample from a larger set, what was the sampling strategy (\eg, deterministic, probabilistic with specific sampling probabilities)?} Not applicable. We did not sample data from a larger set. But we use RetinaFace~\cite{deng2020retinaface} for detecting and cropping faces to ensure each image only contains one face. 
    \item \textbf{Over what timeframe was the data collected? Does this timeframe match the creation timeframe of the data associated with the instances (\eg, recent crawl of old news articles)? If not, please describe the timeframe in which the data associated with the instances was created.} The data was collected from February 2024 to April 2024, even though the data were originally released before this time. Please refer to the cited papers in Table~\ref{tab:dataset_info_detail} for specific original data released time.
\end{compactitem}

\subsection{Data Processing}
\begin{compactitem}
    \item \textbf{Was any preprocessing/cleaning/labeling of the data done (\eg, discretization or bucketing, tokenization, part-of-speech tagging, SIFT feature extraction, removal of instances, processing of missing values)?} Yes. We discussed in `Data Collection' in Section~\ref{subsec:data_collection}.
    \item \textbf{Was the `raw' data saved in addition to the preprocessed/cleaned/labeled data (\eg, to support unanticipated future uses)? If so, please provide a link or other access point to the `raw' data.} The `raw' data can be acquired through the original data publisher. Please see the cited papers in Table~\ref{tab:dataset_info_detail}.
    \item \textbf{Is the software used to preprocess/clean/label the instances available? If so, please provide a link or other access point.} Yes. We use RetinaFace~\cite{deng2020retinaface} for detecting and cropping faces to ensure each image only contains one face. Demographic annotations are given by our annotator, see `Annotation Generation' in Section~\ref{subsec:annotation_generation}. Our annotator code will not be released considering the ethical guidelines.
    \item \textbf{Does this dataset collection/processing procedure achieve the motivation for creating the dataset stated in the first section of this datasheet? If not, what are the limitations?} Yes. The dataset does allow for the study of our goal, as it covers comprehensive generation methods, demographic annotations for evaluating current detectors and training fairer detectors.
\end{compactitem}

\subsection{Dataset Distribution}
\begin{compactitem}
    \item \textbf{How will the dataset be distributed?} We distribute all the data as well as CSV files that formatted all annotations of images under the CC BY-NC-ND 4.0 license and strictly for research purposes.
    \item \textbf{When will the dataset be released/first distributed? What license (if any) is it distributed under?} The dataset will be released following the paper's acceptance, and it will be under the permissible CC BY-NC-ND 4.0 license for research-based use only.  Users can access our dataset by submitting an EULA.
    \item \textbf{Are there any copyrights on the data?} We believe our use is `fair use' since all data in our dataset is collected from public datasets.
    \item \textbf{Are there any fees or access restrictions?} No. 
\end{compactitem}

\subsection{Dataset Maintenance}
\begin{compactitem}
    \item \textbf{Who is supporting/hosting/maintaining the dataset?} The first author of this paper.
    \item \textbf{Will the dataset be updated? If so, how often and by whom?} We do not plan to update it at this time.
    \item \textbf{Is there a repository to link to any/all papers/systems that use this dataset?}  Our fairness benchmark uses this dataset, a brief instruction of how to use this dataset and the code of fairness benchmark is on {\small \url{https://github.com/Purdue-M2/AI-Face-FairnessBench}}. 
    \item \textbf{If others want to extend/augment/build on this dataset, is there a mechanism for them to do so?} Not at this time.
\end{compactitem}

\subsection{Legal and Ethical Considerations}
\begin{compactitem}
    \item \textbf{Were any ethical review processes conducted (\eg, by an institutional review board)?} No official processes were done since all data in our dataset were collected from the existing public datasets.
    \item \textbf{Does the dataset contain data that might be considered confidential?} No. We only use data from public datasets.
    \item \textbf{Does the dataset contain data that, if viewed directly, might be offensive, insulting, threatening, or might otherwise cause anxiety? If so, please describe why} No. It is a face image dataset, we have not seen any instance of offensive or abusive content.
    \item \textbf{Does the dataset relate to people?} Yes. It is a face image dataset containing real face images and AI-generated face images.
    \item \textbf{Does the dataset identify any subpopulations (\eg, by age, gender)?} Yes, through demographic annotations.
    \item \textbf{Is it possible to identify individuals (\ie, one or more natural persons), either directly or indirectly (\ie, in combination with other data) from the dataset?} Yes. It is a face image dataset. The age, gender, and skin tone can be identified through the face image, also through the demographic annotation we provide. All of the images that we use are from publicly available data.
\end{compactitem}

\subsection{Author Statement and Confirmation of Data License}
The authors of this work declare that the dataset described and provided has been collected, processed, and made available with full adherence to all applicable ethical guidelines and regulations. We accept full responsibility for any violations of rights or ethical guidelines that may arise from the use of this dataset. We also confirm that the dataset is released under the CC BY-NC-ND 4.0 license, permitting sharing and downloading of the work in any medium, provided the original author is credited, and it is used non-commercially with no derivative works created.